\documentclass{article}

\usepackage[eandd,final,nonatbib]{neurips_2026}
\usepackage[numbers,compress]{natbib}

\usepackage[utf8]{inputenc}
\usepackage[T1]{fontenc}
\usepackage{hyperref}
\usepackage{url}
\usepackage{booktabs}
\usepackage{multirow}
\usepackage{amsfonts}
\usepackage{amsmath}
\usepackage{amssymb}
\usepackage{nicefrac}
\usepackage{microtype}
\usepackage{xcolor}
\usepackage{graphicx}
\usepackage{tikz}
\usepackage{enumitem}
\usepackage{xspace}
\usepackage{soul}

\usepackage{algorithm}
\usepackage{algpseudocode}
\usepackage{listings}
\usepackage{tcolorbox}
\tcbuselibrary{listings, skins}
\usetikzlibrary{fit, positioning, arrows.meta, backgrounds}

\usepackage{subcaption}
\usepackage{tikz, pgfplots}
\usepackage{pgfplotstable}
\usetikzlibrary{patterns}
\usetikzlibrary{positioning}

\usepackage[table]{xcolor}

\newcommand{\ours}{LCS-Bench\xspace}

\usepackage{pifont}
\newcommand{\cmark}{\ding{51}}
\newcommand{\xmark}{\ding{55}}

\lstdefinelanguage{lean}{
  keywords={def,theorem,lemma,inductive,structure,where,by,fun,let,have,show,from,
            match,with,if,then,else,return,do,namespace,end,open,import,abbrev,
            instance,class,extends,deriving,Type,Prop,Sort,sorry},
  sensitive=true,
  morecomment=[l]{--},
  morecomment=[s]{/-}{-/},
  morestring=[b]",
}

\definecolor{codebox-frame}{RGB}{101,126,176}
\definecolor{codebox-bg}   {RGB}{219,224,235}

\definecolor{mode-zeroshot}{RGB}{101,126,176}
\definecolor{mode-thinking}{RGB}{176,145,101}
\definecolor{mode-agentic} {RGB}{176,101,110}
\tcbset{
  codebox/.style={
    fonttitle=\bfseries\small,
    colback=codebox-bg, colframe=codebox-frame,
    top=2pt, bottom=2pt, left=3pt, right=3pt,
    boxsep=0pt, toptitle=4pt, bottomtitle=4pt,
  }
}

\definecolor{gh-keyword}{HTML}{CF222E}
\definecolor{gh-comment}{HTML}{6E7781}
\definecolor{gh-string} {HTML}{0A3069}
\definecolor{gh-bg}     {HTML}{F6F8FA}

\newcommand{\lcsrepoloc}{85K\xspace}
\newcommand{\lcsnumtextbookitems}{327\xspace}
\newcommand{\lcsgit}{\url{https://github.com/SnowBallMacaron/lcs-bench}}
\newcommand{\lcskaggle}{\url{https://www.kaggle.com/datasets/macaroonotz/lcs-bench}}

\title{Theory-Scale Auto-Formalization of \\ Logic for Computer Science}

\author{%
  Yuming Feng \\
  Johns Hopkins University \\
  \texttt{yfeng97@cs.jhu.edu} \\
  \And
  Frederick Pu \\
  University of Toronto \\
  \texttt{frederick.pu@mail.utoronto.ca} \\
  \And
  One An \\
  University of Pennsylvania \\
  \texttt{onean@sas.upenn.edu} \\
  \And
  Osbert Bastani \\
  University of Pennsylvania \\
  \texttt{obastani@seas.upenn.edu} \\
  \And
  Li Zhang \\
  Drexel University \\
  \texttt{Harry.Zhang@drexel.edu} \\
  \And
  Jiani Huang \\
  University of Pennsylvania \\
  \texttt{jianih@seas.upenn.edu} \\
  \And
  Xujie Si$^\dagger$ \\
  University of Toronto \\
  \texttt{six@cs.toronto.edu} \\
  \And
  Ziyang Li$^\dagger$ \\
  Johns Hopkins University \\
  \texttt{ziyang@cs.jhu.edu}
}

\begin{document}

\maketitle

{\renewcommand{\thefootnote}{\fnsymbol{footnote}}\footnotetext[2]{Equal advising.}}

\begin{abstract}
  Auto-formalization is critical for scalable formal verification, but existing progress largely focuses on isolated statements, while \emph{theory-scale} auto-formalization, which coherently translates hundreds of interdependent definitions, lemmas, and theorems, remains open due to challenges in consistency, faithfulness, scalability, and correctness.
In this paper, we introduce \emph{LCS-Bench}, a stand-alone, theory-scale benchmark based on \emph{Logic for Computer Science}.
LCS-Bench is built through a novel semi-automated agentic pipeline that leverages concept graphs, formal signature planning, issue tracking, sorry-filling with counter-example search, complemented by faithfulness review from human experts.
The resulting artifact covers \lcsnumtextbookitems textbook items, over {4,076} Lean declarations, and more than \lcsrepoloc lines of Lean code.
The dataset supports broad evaluation through a data engine that automatically derives five tracks of evaluation benchmarks, measuring different aspects of auto-formalization and theorem-proving capabilities.
We also introduce a novel evaluation protocol featuring definitional equivalence checkers, enabling more fine-grained and faithful assessment.
Through extensive evaluation on 17 models, we demonstrate that
(1) LCS-Bench is of high quality, consistent, and faithful;
(2) the benchmark is challenging, with state-of-the-art models achieving only 20.3\% on auto-formalization tasks; and
(3) our analysis reveals key findings regarding theory-scale auto-formalization and suggests promising directions for future work.

\end{abstract}

\section{Introduction}
\label{sec:introduction}

Auto-formalization, the translation of informal mathematics and logic into machine-verifiable statements and proofs, has become a central problem at the intersection of AI and formal methods~\cite{wu2022autoformalization,yang2024formal,li2024survey,weng2025autoformalization}.
Despite recent progress, \emph{theory-scale} auto-formalization remains largely unsolved: real mathematical theories require hundreds of interdependent definitions, lemmas, and theorems to be formalized as a coherent whole.
This setting raises fundamental challenges in maintaining global consistency, preserving faithfulness to the source text, and producing artifacts that are not only syntactically valid but also formally provable.
Progress is further limited by the lack of tailored benchmarks and evaluation protocols, as many existing evaluations rely heavily on mature libraries such as Mathlib~\cite{mathlib2020}, obscuring the difficulty of building theories from first principles.

Existing auto-formalization benchmarks fall short of this setting in scope, coverage, and self-containment.
Most focus on isolated statements rather than full developments with heterogeneous artifacts, such as  definitions, lemmas, theorems, proofs, algorithms, and exercises~\cite{zheng2022minif2f,azerbayev2023proofnet,liu2023fimo,jiang2025fate,liu2025combibench,xu2025leancat}.
Others either do not require formally verified proofs, as in PutnamBench~\cite{tsoukalas2024putnambench}, or are built on substantial pre-existing infrastructure, as in LeanDojo~\cite{yang2023leandojo} and miniCTX~\cite{hu2024minictx}.
As a result, current evaluations do not adequately capture theory-scale auto-formalization.
We therefore seek a benchmark for Lean that is theory-scale, stand-alone, and heterogeneous, covering a broad spectrum of formalization tasks within a single coherent development.

\begin{figure}[t]
    \centering
    \includegraphics[width=\textwidth]{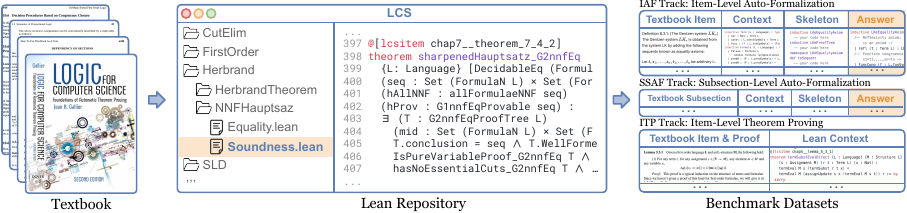}
    \vspace{-10px}
    \caption{
        An overview of LCS-Bench.
        Tracks of datasets are generated from the Lean repository derived from the textbook, covering a wide range of auto-formalization and theorem-proving tasks.
    }
    \label{fig:banner}
    \vspace{-10px}
\end{figure}

Figure~\ref{fig:banner} illustrates the overall design of \ours, where we turn the textbook into a Lean repository and then derive evaluation tracks from it.
To realize this goal, we start constructing our benchmark from \textit{Logic for Computer Science: Foundations of Automatic Theorem Proving, Second Edition} \cite{gallier2015logic}.
We choose this text because it is self-contained, comprehensive, and centered on core topics in logic for computer science, spanning propositional logic, first-order logic, theorem proving, and logic programming including Prolog.
Beyond theorems and proofs, it contains rich structural content such as domain-specific languages, deeply embedded proof trees, algorithms, and meta-theory, which are largely absent from prior auto-formalization benchmarks. LCS-Bench therefore broadens auto-formalization from isolated mathematical statements to heterogeneous, theory-scale developments common in computer science.

From this source, we build a large-scale Lean repository through a semi-automated, expert-in-the-loop formalization pipeline.
The pipeline combines concept graph construction, formal signature planning, automated sorry-filling with counter-example search, and issue tracking, with expert intervention resolving unfaithfulness and formalization gaps.
The resulting repository formalizes essentially the full technical content of the textbook, covering eight chapters, yielding 341 textbook items, {4,076} Lean declarations, and 85K lines of Lean code, including 28K lines of proof; of which \lcsnumtextbookitems items enter the benchmark.
We aim to prove every theorem and lemma, and achieve this for 86\% of them;
the remaining cases are primarily mechanically demanding meta-theorems, such as termination and completeness results, whose formalizations are validated by human experts.

Operating on the codebase, we build a data engine that automatically derives five evaluation tracks: three auto-formalization (AF) tracks, Item-level AF (IAF), Subsection-level AF (SSAF), and IAF with Distractors (IAF-D), and two theorem-proving (TP) tracks, Declaration-level TP (DTP) and Item-level TP (ITP), totaling {1,271} benchmark instances.
Each instance provides sufficient context for theory-scale reasoning and is designed to test coherence across related declarations rather than isolated statement translation.
For evaluation, we introduce a novel \emph{definitional equivalence checker}, which supports fine-grained assessment of type definitions and declarations beyond logical equivalence alone.
The resulting dataset is challenging: even item-level instances require formalizing multiple heterogeneous declarations within a rich theory-scale context.
In summary, our contributions are:
\begin{enumerate}[leftmargin=*, topsep=2pt, itemsep=1pt, parsep=0pt]
\item We introduce and responsibly release \textbf{LCS-Bench}, a comprehensive, stand-alone, and challenging benchmark for theory-scale auto-formalization based on \emph{Logic for Computer Science}~\cite{gallier2015logic}.
\item
We develop a semi-automated agentic pipeline for benchmark construction, integrating concept graph construction, formal signature planning, and issue tracking with a validation-and-repair loop that adversarially validates declarations, repairs their upstream sources, and replays dependent proofs, complemented by human expert review.
\item We implement a data engine that generates five evaluation tracks comprising 1,271 data points, together with a definitional equivalence checker for fine-grained and faithful assessment.
\item By evaluating 17 models in zero-shot, thinking, and agentic modes, we demonstrate the quality and difficulty of LCS-Bench and uncover key findings for theory-scale auto-formalization.
\end{enumerate}

\section{Illustrative Examples}
\label{sec:motivating-example}

\begin{figure}
    \centering
    \includegraphics[width=\linewidth]{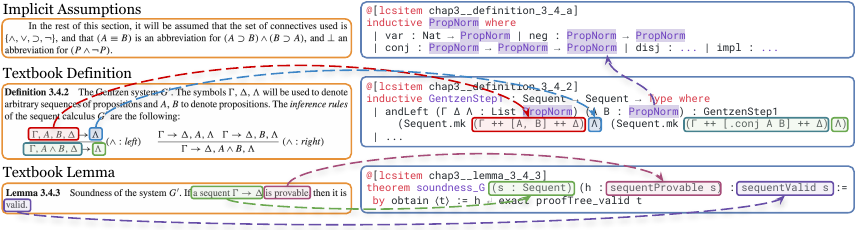}
    \vspace{-10px}
    \caption{
        Illustrative examples in the textbook about Gentzen system $G'$ (Left) and its formalization in \ours (Right).
        We show one staged implicit assumption in the textbook (``In the rest of this section, it will be assumed\dots''), one graphical illustration for Gentzen system definition (horizontal line-based inference rules), and one soundness theorem about the Gentzen system.
    }
    \label{fig:motivating}
    \vspace{-10px}
\end{figure}

We start by presenting a few illustrative examples in \ours (Figure~\ref{fig:motivating}).
The examples are drawn from Chap.~3 of the textbook, which formalizes propositional logic and the Gentzen sequent calculus $G'$.
Definition~3.4.2 introduces the eight inference rules of $G'$, which are presented in the textbook entirely as a typeset diagram of inference rules, and Lemma~3.4.3 states the soundness of $G'$.

Each numbered textbook item is linked to its Lean formalization via the \texttt{@[lcsitem]} attribute, providing a stable identifier used throughout the pipeline and benchmark.
One textbook item typically corresponds to multiple Lean declarations.
Definition~3.4.2 maps to two inductive types (\texttt{GentzenStep1}, \texttt{GentzenStep2}), and Lemma~3.4.3 maps to a single theorem \texttt{soundness\_G}.
Several features make this example representative of the challenges in \ours:
\begin{enumerate}[leftmargin=*, label=\arabic*), topsep=2pt, itemsep=1pt, parsep=0pt]
  \item \textit{Theory-scale context.} Lemma~3.4.3 transitively depends on 79 Lean declarations across 11 textbook items, including propositional semantics, sequents, axioms, and proof trees.
  \item \textit{Graphical definitions.} Definition~3.4.2 exists in the textbook \emph{only} as a diagram of horizontal-line inference rules, which must be translated into inductive types forming a deeply embedded DSL.
  \item \textit{Implicit assumptions.} Formalizing sequents requires a separate type \texttt{PropNorm} (full name \texttt{PropFormulaNormalized}, shortened here for brevity) than the formal definition of propositional formulas, yet this restriction appears only as a prose convention with no labeled definition.
  \item \textit{Heterogeneous items.} Definition~3.4.2 is an inductive type (a DSL); Lemma~3.4.3 is a theorem---two fundamentally different Lean constructs arising from the same chapter.
  \item \textit{Formally proved.} \texttt{soundness\_G} is fully proved, reducing to case analyses over all eight inference rules. \ours contains proofs of individual declarations reaching up to 1,600 lines.
\end{enumerate}

These examples yield tasks for two tracks.
In the \emph{auto-formalization} track, a model receives the textbook text for Definition~3.4.2 (including the graphical rules) together with all dependency context, and must produce the two inductive types that faithfully embed the rule system.
In the \emph{theorem-proving} track, a model receives the statement of \texttt{soundness\_G} and the full Lean context, and must fill in the proof, which requires reasoning about valuations, sequent validity, and all eight rule cases.

Notably, these examples are drawn from Chap.~3, the first technical chapter after mathematical background.
The textbook subsequently covers proof search algorithms, first-order logic, Herbrand's theorem, resolution, and SLD resolution for Prolog, each introducing new DSLs, meta-theorems, and algorithmic content of increasing complexity.

\section{Methodology}
\label{sec:methodology}

\begin{figure}[t]
    \centering
    \includegraphics[width=\linewidth]{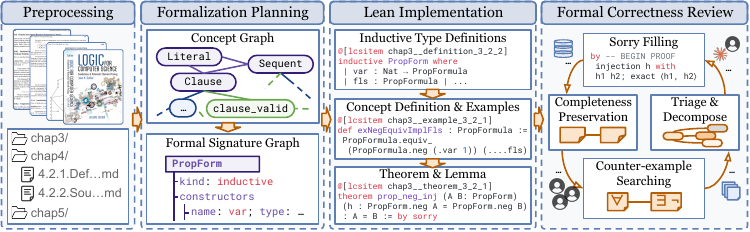}
    \vspace{-10px}
    \caption{
        Overview of the \ours formalization pipeline, which consists of four stages from preprocessing to planning, implementation, and formal correctness review.
    }
    \label{fig:pipeline}
    \vspace{-10px}
\end{figure}

We now describe the semi-automated agentic pipeline for converting the textbook into a validated, high-quality Lean library, which we call the \emph{\ours Lean artifact}.
The pipeline consists of four stages illustrated in Figure~\ref{fig:pipeline}: pre-processing, formalization planning, Lean implementation, and formal correctness review.
The primary models used in the pipeline are Claude Opus 4.6~\cite{claudeopus4} and GPT 5.3~\cite{gpt5}, accessed via Claude Code and Codex CLI respectively, used in conjunction with each other in agentic mode.
Further details are provided in Appendix~\ref{sec:appendix-pipeline}.

\noindent\textbf{Pre-Processing.}
To ensure each formalization target is explainable and precisely traceable, we preprocess the textbook into a structured, item-level representation.
We convert the textbook from PDF to Markdown with Marker~\cite{marker2024}, one file per subsection, and manually redraw eligible graphics, such as proof trees, inference rules, and directed acyclic graphs, as ASCII diagrams embedded in the Markdown.
Each textbook item is wrapped in a typed XML tag, such as \texttt{<theorem>}, \texttt{<lemma>}, \texttt{<definition>}, \texttt{<example>}, or \texttt{<proof>}.
We also manually annotate implicit formal statements expressed in prose, including scope restrictions, standing assumptions, and informal lemmas.

\noindent\textbf{Formalization Planning.}
Before any Lean code is implemented, an agent analyzes the textbook's mathematical structure and generates a two-layer intermediate representation (IR) blueprint for faithful, self-contained formalization.
First, we generate a textbook-self-consistent \emph{concept graph}: nodes are keywords or sub-clauses from the text (e.g., ``propositional formula'', ``left parenthesis count'', ``disjoint antecedent and succedent''), and edges encode textbook-level relationships such as \textsc{is-a}, \textsc{has-a}, \textsc{relies-on}, and \textsc{defined-by}.
Second, we generate a \emph{formal signature graph} that maps each concept node to its Lean realization, including its \texttt{lean\_kind},
constructors, fields, companion definitions, and embedding level. The embedding level is \emph{deep} for syntactic DSLs encoded as inductive types, \emph{shallow} for semantic properties over deep syntax using Lean's \texttt{Prop} layer, and \emph{mixed} for meta-theorems connecting the two, such as soundness, completeness, and termination.
Critically, the embedding level must be assigned correctly so that deep syntactic constructs are available as first-class inductive types when the agent later states meta-theorems about them.

\noindent\textbf{Lean Implementation.}
Given the informal concept graph and the formal signature graph, an agentic implementer translates each module into Lean code, proceeding in three rounds that respect the dependency order established in the planning stage.
After each round, \texttt{lake build} is run and a review script verifies correctness and faithfulness to the signature graph.
\begin{itemize}[leftmargin=*, topsep=2pt, itemsep=1pt, parsep=0pt]
  \item \textit{Round 1: Structures and inductive types.}
  The agent first defines all \texttt{inductive} types and \texttt{structure}s, which form the backbone for later definitions and theorems.
  Each Lean declaration is tagged with \texttt{@[lcsitem]} encoding the stable item ID; the \texttt{lcsmain} flag marks the primary authored declaration, distinguishing it from companion definitions and elaboration byproducts.
  \item \textit{Round 2: Definitions, functions, and examples.}
  Given the type scaffolding, the agent defines functions, abbreviations, instances, and examples, including computational content (e.g., evaluators, normalizers, search procedures) and semantic predicates (e.g., validity, satisfiability, tautology).
  \item \textit{Round 3: Theorems and lemmas.}
  The agent writes theorem and lemma \emph{statements} drawn directly from the formal signature graph, leaving all proof bodies as \texttt{sorry} placeholders.
  The goal is theory-scale coherence: all statements type-check, all dependencies compile, and the overall structure faithfully mirrors the textbook.
  An implementation validator flags vacuous theorem statements, missing inductive cases, identity-function stubs, and dead \texttt{@[lcsitem]} declarations.
\end{itemize}

\noindent\textbf{Formal Correctness Review.}
This stage fills \texttt{sorry} placeholders and checks that proofs are both formally correct and faithful to the textbook.

Before dependent proofs are attempted, the pipeline adversarially validates each declaration by searching for counter-proofs or pathological inhabitants; a successful attack repairs the responsible definition or signature, after which dependent obligations are replayed and an opposite-direction check guards against silently strengthening the claim.
A high-level orchestrator tracks remaining issues, triages them by severity, decomposes stuck proofs into subgoals, and triggers a mandatory three-way review once a theorem is claimed resolved, comparing the final Lean declaration against the textbook, the formal signature graph, and the implementation history.

\begin{figure}[t]
    \centering
    \includegraphics[width=\linewidth]{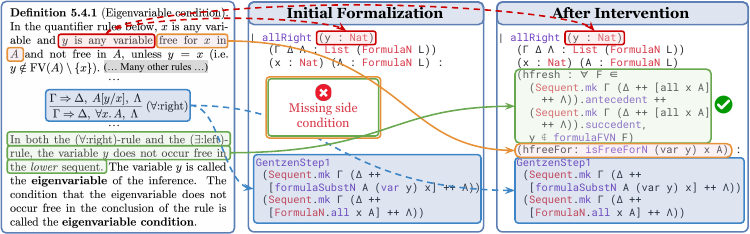}
    \vspace{-10pt}
    \caption{
        A concrete example of human expert intervention.
        The $(\forall\text{:right})$ constructor is initially formalized without the textbook's two side conditions (the eigenvariable condition and the free-for condition), yielding a rule that Lean accepts but that admits unsound inferences.
    }
    \label{fig:example-intervention}
    \vspace{-10pt}
\end{figure}

\noindent\textbf{Human Expert Intervention.}
The automated pipeline handles the bulk of formalization, but human expert intervention is essential for issues requiring global judgment or resolution of genuine ambiguity.
Human experts, all of whom are authors with graduate-level training in logic and programming languages and prior Lean~4 experience, invested approximately six person-months, contributing 190K tokens of feedback across 600+ tracked issues, alongside significant direct rewrite of roughly 30\% of major definitions, implicit assumptions, and core theorems.
To further alleviate code-smells from few driving models, the main naming and stylistic conventions are all set by human experts.
Other than these, the primary categories of failures demanding interventions are:
\begin{enumerate}[leftmargin=*, topsep=2pt, itemsep=1pt, parsep=0pt]
  \item \textit{Missing standing assumptions.}
Textbook assumptions may be introduced far from where they are used, with either broad or local scope.  Agents relying on immediate context often miss or dismiss them, producing theorems that become unprovable only in later proof attempts.
  \item \textit{Vacuous formalization.}
  The Lean statement may diverge from the textbook even when its type-level spec appears correct.
  A common failure is signature drift toward an overly loose statement, admitting unintended witnesses and making downstream lemmas trivially provable but unfaithful.
  \item \textit{Accumulated workarounds on a false foundation.}
  When an agent inherits a declaration whose statement is already wrong, subsequent agents patch around it rather than fixing it, trading faithfulness for the effort of a proper rewrite.
  Over time this compounds into rippling unfaithful statements built on top of a false claim, requiring a full revert of the dependent chain.
\end{enumerate}
Figure~\ref{fig:example-intervention} shows a false foundation that triggers accumulated workarounds.
Due to the length of Definition 5.4.1, one side condition (eigenvariable) appears at the very end, so the agent omitted it in the initial formalization, and earlier checks failed to catch the error.
During the proof of a subsequent theorem, the agent repeatedly failed to close the cases generated by this rule, unable to tell whether the proof was hard or the definition was wrong.
At this point, a human expert can step in and use dependency analysis to identify and correct the missing side condition.

In all cases, the human expert's role is to review, identify, and resolve the issue at the source.
Human experts often significantly rewrite the statement and channel it back to re-implement and re-prove downstream from stage 2 or 3, ensuring that the fixes are properly propagated.

\section{Benchmark and Datasets}
\label{sec:benchmark}

\ours organizes evaluation datasets into five tracks spanning two task families: \emph{auto-formalization} (AF) and \emph{theorem-proving} (TP).
Across all tracks, there are {1,271} data points in total.
Table~\ref{tab:af-tracks-main} and Table~\ref{tab:tp-tracks-main} summarize the track statistics including the lines-of-code (LoC) and number of Lean declarations (\#Decls).
Further details are provided in the Appendix~\ref{sec:appendix-dataset-details}.

\noindent\textbf{Auto-formalization tracks.}
Each AF data point provides a context of existing Lean code and textbook content and asks the model to produce a \texttt{submission.lean} containing the formalized declarations.
\emph{IAF} (Item-level Auto-Formalization) is the atomic unit: one textbook item per data point.
\emph{SSAF} (Subsection-level) scales up the output, requiring the model to jointly formalize all items in a subsection.
\emph{IAF-D} (IAF with Distractors) injects semantically related but irrelevant declarations into the context to test robustness to redundant information.

Each AF data point is a quadruplet $(Q, C, T_y, y)$: a textbook quote $Q$, a context $C$ of dependent Lean declarations, a starter template $T_y$ with declaration skeletons, and a reference formalization $y$.
Since a textbook item admits many valid formalizations, each data point is interface-constrained: $T_y$ fixes the target declaration names, kinds, namespaces, and constructor names, so the model implements a specified component of the surrounding theory rather than an independent encoding.
A model receives $(Q, C, T_y)$ and produces $\hat{y}$, a \texttt{Submission.lean} file.
For evaluation, we propose a novel \emph{DefEq Checker}, a Lean program that checks definitional equivalence between $y$ and $\hat{y}$.
No existing metric handles the heterogeneous declaration kinds present in our dataset (constants, definitions, theorems, inductives, and constructors).
Table~\ref{tab:defeq-features} summarizes the checker's coverage, with full details and checker algorithm in Appendix~\ref{sec:appendix-defeq-checker}.

\begin{figure*}[t]
    \centering
    \begin{minipage}{\textwidth}
        \begin{minipage}{0.48\linewidth}
\centering
\footnotesize
\setlength{\tabcolsep}{5.3pt}
\begin{tabular}{lrrrrr}
\toprule
\multirow{2}{*}{\textbf{Track}} & \multirow{2}{*}{\textbf{\#Points}} &
  \multicolumn{2}{c}{\textbf{Input LoC}} &
  \multicolumn{2}{c}{\textbf{\#Output Decls}} \\
\cmidrule(lr){3-4}\cmidrule(lr){5-6}
 & & Mean & Max & Mean & Max \\
\midrule
IAF   & 327 &  359.37 & 1559 &  6.49 &  57 \\
SSAF  & 116 &  428.05 & 1551 & 18.21 &  93 \\
IAF-D & 327 & 1383.30 & 4951 &  6.49 &  57  \\
\bottomrule
\end{tabular}
\vspace{-5px}
\captionof{table}{Dataset statistics for the three AF tracks.}
\label{tab:af-tracks-main}
\end{minipage}

        \hfill
        \begin{minipage}{0.465\linewidth}
\centering
\footnotesize
\setlength{\tabcolsep}{4.5pt}
\begin{tabular}{lrrrrr}
\toprule
\multirow{2}{*}{\textbf{Track}} & \multirow{2}{*}{\textbf{\#Points}} &
  \multicolumn{2}{c}{\textbf{Input LoC}} &
  \multicolumn{2}{c}{\textbf{Ref. Output LoC}} \\
\cmidrule(lr){3-4}\cmidrule(lr){5-6}
 & & Mean & Max & Mean & Max  \\
\midrule
  DTP & 420 & 383.26 & 1799 &  38.57 & 480 \\
  ITP &  81 & 300.96 &  845 & 138.85 & 608 \\
  -- & -- & -- & -- & -- & -- \\
\bottomrule
\end{tabular}
\vspace{-5px}
\captionof{table}{Dataset statistics for the two TP tracks.}
\label{tab:tp-tracks-main}
\end{minipage}

    \end{minipage}
\end{figure*}

\begin{figure}[t]
  \vspace{-5px}

  \pgfplotsset{
    iafbar/.style={
      bar width=4pt,
      height=3.8cm,
      ymin=0,
      ymajorgrids=true,
      major grid style={dotted, color=codebox-frame!20!black},
      ylabel style={font=\footnotesize, color=black},
      xticklabel style={font=\scriptsize},
      yticklabel style={font=\scriptsize},
      enlarge x limits=0.12,
    }
  }

\begin{subfigure}[b]{0.31\textwidth}
  \centering
  \begin{tikzpicture}
    \begin{axis}[
      iafbar,
      ybar stacked,
      bar width=7pt,
      width=1.2\linewidth,
      ylabel={\#Items},
      symbolic x coords={3,4,5,6,7,8,9,10},
      xtick=data,
      ymax=110,
      ytick={0,20,40,60,80,100},
      y label style={at={(axis description cs:0.18,.5)}},
      at={(0,0)}, anchor=south west,
      legend style={
        at={(1,1)}, anchor=north east,
        draw=codebox-frame!60, fill=white, fill opacity=0.9,
        text opacity=1,
        font=\tiny,
        inner sep=2pt,
        row sep=-3pt,
      },
      legend cell align=left,
      legend image code/.code={
        \draw[#1] (0cm,-0.08cm) rectangle (0.22cm,0.08cm);
      },
    ]
    \addplot[fill=codebox-frame!60, draw=codebox-frame] coordinates {
      (3,71)(4,25)(5,79)(6,41)(7,42)(8,29)(9,35)(10,5)
    };
    \addlegendentry{Formalized}
    \addplot[
      fill=codebox-frame!25, draw=codebox-frame,
      nodes near coords,
      point meta=explicit,
      every node near coord/.append style={font=\tiny, inner sep=1pt},
    ] coordinates {
    (3,0)[71] (4,0)[25] (5,10)[89] (6,0)[41]
    (7,12)[54] (8,0)[29] (9,0)[35] (10,27)[32]
    };
    \addlegendentry{Non-formalized}
    \end{axis}
  \end{tikzpicture}
  \caption{Count by chapter.}
  \label{fig:iaf-stats-chapter}
\end{subfigure}
\hfill
\begin{subfigure}[b]{0.31\textwidth}
  \centering
  \begin{tikzpicture}
    \begin{axis}[
      iafbar,
      ybar stacked,
      width=1.2\linewidth,
      height=3.8cm,
      symbolic x coords={Def,Thm,Lem,Ex},
      xtick=data,
      ymax=200,
      ytick={0,30,60,90,120,150,180},
      bar width=10pt,
      enlarge x limits=0.25,
      xticklabel shift=0pt,
      at={(0,0)}, anchor=south west,
      nodes near coords,
      every node near coord/.append style={font=\tiny, inner sep=1pt},
      legend style={
        at={(1,1)}, anchor=north east,
        draw=codebox-frame!60, fill=white, fill opacity=0.9,
        text opacity=1,
        font=\tiny,
        inner sep=2pt,
        row sep=-3pt,
        legend columns=2,
      },
      legend cell align=left,
      legend image code/.code={
        \draw[#1] (0cm,-0.08cm) rectangle (0.22cm,0.08cm);
      },
    ]
      \addplot[fill=codebox-frame!60, draw=codebox-frame, nodes near coords={}]
        coordinates {(Def,103)(Thm,46)(Lem,99)(Ex,79)};
      \addlegendentry{Formalized}
      \addplot[fill=codebox-frame!25, draw=codebox-frame] coordinates {
        (Def,26)[129] (Thm,5)[51] (Lem,10)[109] (Ex,8)[87]
      };
      \addlegendentry{Non-formalized}
    \end{axis}
  \end{tikzpicture}
  \caption{Count by item type.}
  \label{fig:iaf-stats-type}
\end{subfigure}
\hfill
\begin{subfigure}[b]{0.33\textwidth}
  \centering
  \begin{tikzpicture}
    \begin{axis}[
        width=1.15\linewidth,
        height=3.8cm,
        xlabel={Input context LoC},
        ylabel={Ref. output LoC},
        xlabel style={font=\scriptsize, color=black},
        ylabel style={font=\scriptsize, color=black},
        xticklabel style={font=\tiny},
        yticklabel style={font=\tiny},
        xticklabel shift=-4pt,
        yticklabel shift=-4pt,
        xmode=log, ymode=log,
        log basis x=10, log basis y=10,
        xmin=10, ymin=10,
        minor x tick num=8,
        minor y tick num=8,
        grid=both,
        major grid style={dotted, color=codebox-frame!20!black},
        minor grid style={dotted, color=codebox-frame!10!black},
        tick align=outside,
        enlargelimits=0.08,
        y label style={at={(axis description cs:0.15,.5)}},
        x label style={at={(axis description cs:0.5,0.12)}},
        scatter/use mapped color={
            draw=codebox-frame,
            fill=codebox-frame!60,
        },
    ]
    \addplot[
        scatter,
        only marks,
        mark=*,
        mark options={opacity=0.55},
        visualization depends on={\thisrow{count} \as \count},
        scatter/@pre marker code/.append style={
            /tikz/mark size=\count*1.2
        },
    ]
    table[
        x=context_bucket,
        y=gt_bucket,
    ] {data/iaf_loc_bucketed.dat};
    \end{axis}
  \end{tikzpicture}
  \vspace{-19px}
  \caption{Distribution of output/input LoC.}
  \label{fig:iaf-stats-tokens}
\end{subfigure}
\hfill

\caption{
  IAF track statistics. We show the number of textbook items by chapter, item type, and by input or output sizes (LoC).
  The bars are stacked to show the portion of formalized items (darker) vs. non-formalized items from the textbook (faded).
}
\label{fig:iaf-stats}
\vspace{-10px}
\end{figure}
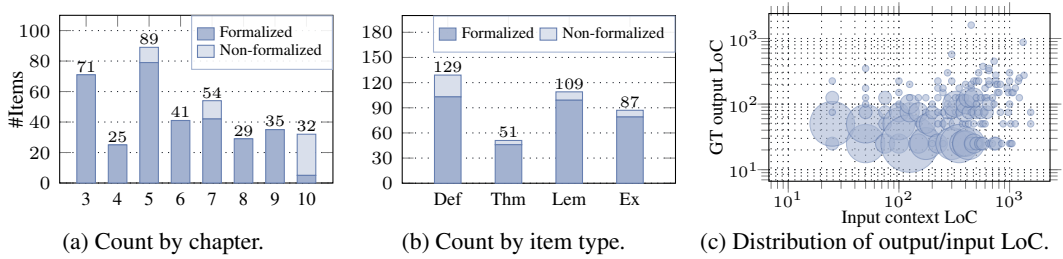

Figure~\ref{fig:iaf-stats} provides detailed per-chapter, per-type, and input/output LoC-distribution breakdowns for the IAF track.
We focus on IAF because it is the most fine-grained AF track and thus the most representative lens into the distribution; the other AF tracks are scaled-up composites built from the same atomic items.
Notably, Figure~\ref{fig:iaf-stats-chapter} shows that \ours covers a vast majority of the textbook: 86\% of all textbook items across chapters 3--10 are formalized and included in the benchmark.

\noindent\textbf{Theorem-proving tracks.}
Each TP data point provides Lean source with \texttt{sorry} placeholders and asks the model to fill them within an interactive budget.
\emph{DTP} (Declaration-level Theorem Proving) targets a single \texttt{sorry}-bearing declaration.
\emph{ITP} (Item-level Theorem Proving) scopes to a full textbook theorem or lemma and requires all its \texttt{sorry}s to be filled jointly, with helper declarations left unconstrained.
As shown in Table~\ref{tab:tp-tracks-main}, of the 145 formalized theorems and lemmas, 81 have fully machine-verified proofs and constitute the TP tracks' data-points.
The size of the reference proofs is illustrated in Figure~\ref{fig:itp-loc-dist} as a measurement of the TP tracks' difficulty distribution.

Each TP data point is also a tuple $(Q, C, T_y, y)$: a textbook item $Q$, a context $C$, a target $T_y$ containing the declaration(s) to prove with their theorem bodies replaced by \texttt{sorry}, and a reference formalization $y$ with complete proofs.
For DTP, $T_y$ is a single \texttt{sorry}-stubbed declaration, often a decomposed helper not directly corresponding to $Q$, so the model receives only $(C, T_y)$.
For ITP, $T_y$ spans every declaration of the textbook item with all theorem bodies \texttt{sorry}-stubbed (other declarations kept intact), and the model receives $(Q, C, T_y)$.
In both cases, the model produces $\hat{y}$, accepted when all expected declarations are present, no \texttt{sorry} remains, and no new \texttt{axiom} is introduced.

\begin{figure}[t]
\centering
\begin{minipage}{0.56\linewidth}
    \centering
    \footnotesize
    \setlength{\tabcolsep}{5pt}
    \begin{tabular}{lll}
        \toprule
        \textbf{Kind} & \textbf{Compared} & \textbf{Criterion} \\
        \midrule
        Definition  & Body expr.         & \textsc{isDefEq} or \textsc{grind} \\
        Theorem     & Statement type     & \textsc{isDefEq} or \textsc{grind} \\
        Inductive   & Constructor names  & Same set (order-independent) \\
        Constructor & Constructor type   & \textsc{isDefEq} only \\
        \bottomrule
    \end{tabular}
    \captionof{table}{Equivalence criteria used by the DefEq Checker.}
    \label{tab:defeq-features}
\end{minipage}%
\hfill
\begin{minipage}{0.40\linewidth}
    \centering
    \begin{tikzpicture}
  \pgfplotsset{
    itpbar/.style={
      ybar,
      bar width=8pt,
      height=3.2cm,
      ymin=0,
      ymajorgrids=true,
      major grid style={dotted, color=codebox-frame!20!black},
      ylabel style={font=\footnotesize, color=black},
      xticklabel style={font=\scriptsize},
      yticklabel style={font=\scriptsize},
      enlarge x limits=0.20,
    }
  }
  \begin{axis}[
    itpbar,
    width=1.13\linewidth,
    ylabel={\#Items},
    y label style={at={(axis description cs:0.15,.5)}},
    symbolic x coords={{<25},{25--49},{50--99},{100--199},{200--399},{$\geq$400}},
    xtick=data,
    x tick label style={font=\scriptsize, rotate=22, anchor=north east},
    xticklabel shift=-6pt,
    ymax=24,
    ytick={0,10,20,30},
    nodes near coords,
    every node near coord/.append style={font=\tiny, inner sep=1pt},
  ]
  \addplot[fill=codebox-frame!60, draw=codebox-frame] coordinates {
      ({<25},4) ({25--49},20) ({50--99},18)
      ({100--199},20) ({200--399},13) ({$\geq$400},6)
  };
  \end{axis}
\end{tikzpicture}
\vspace{-16px}
    \vspace{-6px}
    \captionof{figure}{\#ITP items by Ref. output LoC.}
    \label{fig:itp-loc-dist}
\end{minipage}
\vspace{-10px}

\end{figure}

\noindent\textbf{Data Engine.}
We develop a data engine that performs systematic dependency analysis and Lean code modification to automatically generate all evaluation tracks from the \ours artifact.
This process is AI-free: whatever is stabilized in the artifact is faithfully used to derive the datasets.
For AF tracks, the engine fetches all dependencies, strips proofs, and composes them into a compact context $C$ while generating $T_y$ and $y$. IAF-D is produced by additionally injecting distractor declarations.
For TP tracks, the engine removes proof bodies to produce $C$ and generates a data point only when the full dependency tree is proved, so partially-proved declarations in the current artifact are excluded.

\section{Evaluation}
\label{sec:evaluation}

\subsection{Evaluation Setup}
\label{sec:evaluation-setup}

\noindent\textbf{Models.}
We evaluate 17 models spanning both proprietary and open-sourced families.
The proprietary models are Claude Opus 4.6 (CO46), Claude Sonnet 4.6 (CS46), Claude Haiku 4.5 (CH45)~\cite{claudeopus4}, GPT-5.4 (GPT54)~\cite{gpt5}, and Gemini 3.1 Pro (GP31)~\cite{gemini3}.
The open-sourced models are DeepSeek R1~\cite{deepseekr1} (DSR1), DeepSeek V3.2~\cite{deepseekv3} (DS32), Kimi K2.5~\cite{kimik2} (KK25), Kimi K2.7~\cite{kimik2} (KK27), Devstral 2-123B~\cite{devstral} (D2), Llama 4~\cite{llama4} (L4), Qwen 3 Coder~\cite{qwen3} (Q3C), Qwen V3.2~\cite{qwen25coder} (Q32), MiniMax-M2.5~\cite{minimaxm1} (MM25), GLM-5~\cite{chatglm4} (GLM5), GLM-5.2~\cite{chatglm4} (GLM52), and GPT-OSS-120B~\cite{gptoss} (GPTOSS).

\noindent\textbf{Modes.}
Depending on each model's capabilities, we apply it in up to three modes: (i) \emph{zero-shot}, direct output without thinking; (ii) \emph{thinking}, with extended reasoning enabled; and (iii) \emph{agentic}, running within an agent loop under a fixed budget on the number of turns.
The agentic framework is a lightweight SWE-agent-style scaffold based on mini-swe-agent~\cite{yang2024swe}, with three custom tools exposed to the model: file read/write, \texttt{lake build} (for compilation checking, optionally also verifying the absence of \texttt{sorry}), and a \texttt{submit} action that finalizes the model's output.
We allow up to 20 turns for IAF and IAF-D, 64 for SSAF, and 20 for DTP and ITP.

\noindent\textbf{Track coverage.}
As \ours is primarily an auto-formalization benchmark, we focus our evaluation on the IAF track, running all 17 models across all three applicable modes; results for the remaining tracks are reported alongside.
For IAF-D, we adopt the same setup as IAF.
For SSAF, DTP, and ITP, the excessive context length and per-item difficulty render zero-shot and single-turn thinking modes ineffective, so we report agentic-mode results only.

\begin{figure}[t]
\centering
\resizebox{\linewidth}{!}{
\begin{tikzpicture}
\begin{axis}[
  width=\linewidth, height=4.0cm,
  ymin=0, ymax=30, ytick distance=5,
  ymajorgrids, grid style={dashed, gray!30},
  ylabel={Pass rate (\%)},
  ylabel style={font=\small, at={(axis description cs:0.05,.5)}, anchor=south},
  enlarge x limits=0.015,
  symbolic x coords={
    zs01,zs02,zs03,zs04,zs05,zs06,zs07,zs08,zs09,zs10,zs11,
    gap1,
    th01,th02,th03,th04,th05,th06,th07,th08,th09,th10,th11,th12,
    gap2,
    ag01,ag02,ag03,ag04,ag05,ag06,ag07,ag08,ag09,ag10,ag11
  },
  xtick={zs01,zs02,zs03,zs04,zs05,zs06,zs07,zs08,zs09,zs10,zs11,
         th01,th02,th03,th04,th05,th06,th07,th08,th09,th10,th11,th12,
         ag01,ag02,ag03,ag04,ag05,ag06,ag07,ag08,ag09,ag10,ag11},
  xticklabels={{CO46},{GPT54},{CS46},{GLM5},{CH45},{GLM52},{DS32},{KK25},{D2},{L4},{Q3C},
               {GP31},{CS46},{CO46},{GLM52},{GPT54},{KK27},{GPTOSS},{GLM5},{KK25},{MM25},{Q32},{DSR1},
               {GP31},{KK27},{GLM52},{CO46},{CS46},{GPT54},{GLM5},{KK25},{MM25},{Q32},{GPTOSS}},
  x tick label style={font=\scriptsize, rotate=45, anchor=north east},
  y tick label style={font=\scriptsize},
  every node near coord/.append style={
    font=\fontsize{4.5}{5}\selectfont, anchor=south, yshift=0.5pt, text=black,
    /pgf/number format/fixed, /pgf/number format/fixed zerofill, /pgf/number format/precision=1,
    /pgf/number format/assume math mode=true,
  },
  nodes near coords,
  nodes near coords align={vertical},
  error bars/y dir=both, error bars/y explicit,
  error bars/error bar style={line width=0.4pt, black!70},
  clip=false,
  legend style={
    at={(0,1)}, anchor=north west,
    draw=black!30, fill=white, fill opacity=0.9, text opacity=1,
    font=\scriptsize, inner sep=2pt, row sep=-1pt,
    legend columns=3, column sep=1ex,
  },
  legend cell align=left,
  legend image code/.code={
    \draw[#1] (0cm,-0.08cm) rectangle (0.22cm,0.08cm);
  },
]
  \addplot+[ybar, bar width=6pt, fill=mode-zeroshot, fill opacity=0.6, draw=mode-zeroshot!70!black, mark=none,
           error bars/.cd, y explicit
           ]
    coordinates {
      (zs01,16.8) +- (0,0.47)
      (zs02,15.4) +- (0,0.41)
      (zs03,15.0) +- (0,0.31)
      (zs04,13.9) +- (0,0.62)
      (zs05,11.1) +- (0,0.27)
      (zs06,10.3) +- (0,0.10)
      (zs07, 9.5) +- (0,0.61)
      (zs08, 9.0) +- (0,0.67)
      (zs09, 8.4) +- (0,0.37)
      (zs10, 7.6) +- (0,0.18)
      (zs11, 3.1) +- (0,0.31)
    };
  \addlegendentry{Zero-shot}
  \addplot+[ybar, bar width=6pt, fill=mode-thinking, fill opacity=0.6, draw=mode-thinking!70!black, mark=none,
           error bars/.cd, y explicit]
    coordinates {
      (th01,19.7) +- (0,0.44)
      (th02,17.9) +- (0,0.82)
      (th03,17.6) +- (0,0.27)
      (th04,17.6) +- (0,0.37)
      (th05,16.3) +- (0,0.57)
      (th06,16.1) +- (0,0.27)
      (th07,15.5) +- (0,0.44)
      (th08,14.1) +- (0,0.35)
      (th09,10.7) +- (0,0.35)
      (th10, 8.1) +- (0,0.20)
      (th11, 6.8) +- (0,0.10)
      (th12, 4.2) +- (0,0.10)
    };
  \addlegendentry{Thinking}
  \addplot+[ybar, bar width=6pt, fill=mode-agentic, fill opacity=0.6, draw=mode-agentic!70!black, mark=none,
           error bars/.cd, y explicit]
    coordinates {
      (ag01,20.3) +- (0,0.27)
      (ag02,20.3) +- (0,0.54)
      (ag03,20.2) +- (0,0.53)
      (ag04,20.1) +- (0,0.67)
      (ag05,18.0) +- (0,0.53)
      (ag06,16.1) +- (0,0.37)
      (ag07,15.5) +- (0,0.20)
      (ag08,14.3) +- (0,1.25)
      (ag09,12.1) +- (0,0.10)
      (ag10, 5.5) +- (0,0.00)
      (ag11, 2.1) +- (0,0.31)
    };
  \addlegendentry{Agentic}
\end{axis}
  \end{tikzpicture}
  }
  \vspace{-20px}
  \caption{
    IAF pass rate (\%) per model, averaged over three runs, grouped by evaluation mode (zero-shot, thinking, agentic).
    Error bars show $\pm 1$ standard error of the mean (SEM) across timestamps.
  }
  \label{fig:iaf-passrate}
  \vspace{-10px}
\end{figure}
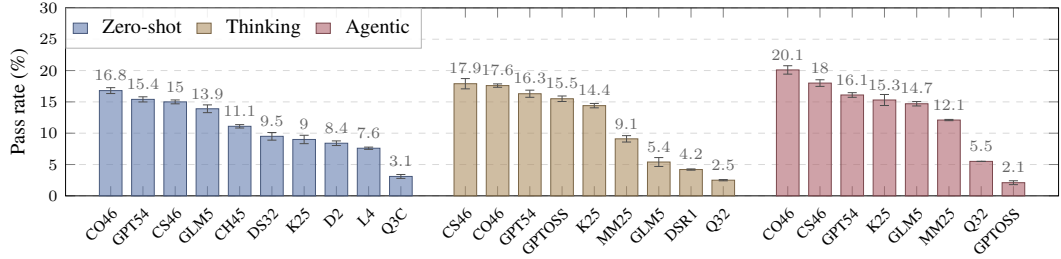

\subsection{Research Questions and Findings}
\label{sec:evaluation-rqs}

\noindent\textbf{RQ1: Can current LLMs auto-formalize \textit{Logic for Computer Science}?}
Current models fall well short of saturation on every track.
As shown in Figure~\ref{fig:iaf-passrate}, the best IAF pass rate is 20.3\%, achieved by Gemini 3.1 Pro and Kimi K2.7 in agentic mode; the weakest model scores only 2.1\%.
Because we repeat each IAF evaluation three times, the standard errors are consistently small (error bars in Figure~\ref{fig:iaf-passrate}), confirming that the rankings faithfully reflect model capability.
Pass@3 rates remain below 26\% for every model, as detailed in the Appendix~\ref{sec:appendix-passk}.
Enabling extended thinking or an agentic loop provides a modest but consistent lift over zero-shot, yet the absolute numbers remain low.

We further note that circular evaluation is not a big concern: in agentic mode, three models not used in construction outperform Claude Opus 4.6 and GPT-5.4, and only 7 of the 101 items solved by Claude or GPT models are solved by no other family, as detailed in the Appendix~\ref{sec:appendix-fairness}.

\begin{figure}[t]
\centering
\begin{minipage}{0.60\linewidth}

\pgfplotsset{
  rq2panel/.style={
    height=4.2cm,
    ymin=0, ymajorgrids, grid style={dashed, gray!30},
    ylabel style={font=\footnotesize, anchor=south},
    y tick label style={font=\scriptsize},
    x tick label style={font=\scriptsize},
    legend style={
      draw=black!30, fill=white, fill opacity=0.9, text opacity=1,
      font=\scriptsize, inner sep=3pt, row sep=-1pt,
      legend columns=1, column sep=1ex,
    },
    legend cell align=left,
  }
}

\begin{subfigure}[t]{0.48\linewidth}
  \centering
  \begin{tikzpicture}
    \begin{axis}[
      rq2panel,
      ylabel style={at={(axis description cs:0.25,.5)}, anchor=south},
      width=1.2\linewidth,
      enlarge x limits=0.08,
      xtick={0,1,2,3,4,5,6,7},
      xticklabels={3,4,5,6,7,8,9,10},
      legend style={at={(1,1)}, anchor=north east},
      ylabel={Pass rate (\%)},
    ]
      \addplot+[mark=*, color=mode-zeroshot, mark options={fill=mode-zeroshot, scale=0.7},
                line width=0.9pt]
        coordinates {(0,25.95)(1,8.12)(2,14.23)(3,4.14)(4,0.44)(5,0.63)(6,5.11)(7,2.43)};
      \addlegendentry{Zero-shot}
      \addplot+[mark=square*, color=mode-thinking, mark options={fill=mode-thinking, scale=0.7},
                line width=0.9pt]
        coordinates {(0,31.14)(1,12.44)(2,17.86)(3,4.81)(4,1.72)(5,1.44)(6,6.03)(7,6.11)};
      \addlegendentry{Thinking}
      \addplot+[mark=triangle*, color=mode-agentic, mark options={fill=mode-agentic, scale=0.7},
                line width=0.9pt]
        coordinates {(0,32.74)(1,13.09)(2,20.25)(3,6.21)(4,2.24)(5,1.78)(6,5.37)(7,10.30)};
      \addlegendentry{Agentic}
    \end{axis}
  \end{tikzpicture}
  \vspace{-17px}
  \caption{Pass rate by chapter.}
  \label{fig:rq2-chapter-main}
\end{subfigure}
\hfill
\begin{subfigure}[t]{0.48\linewidth}
  \centering
  \begin{tikzpicture}
    \begin{axis}[
      rq2panel,
      ymax=33,
      width=1.25\linewidth,
      ybar=0pt, bar width=4pt,
      symbolic x coords={Def,Ex,Lem,Thm},
      xtick=data,
      enlarge x limits=0.2,
      legend style={
        at={(1,1)}, anchor=north east,
        legend columns=2,
      },
      legend image code/.code={
        \draw[#1] (0cm,-0.08cm) rectangle (0.22cm,0.08cm);
      },
      xticklabel shift=-5pt,
    ]
      \addplot+[ybar, fill=mode-zeroshot, fill opacity=0.6,
                draw=mode-zeroshot!70!black, mark=none]
        coordinates {(Def,14.89)(Ex,4.84)(Lem,13.62)(Thm,6.59)};
      \addlegendentry{Zero-shot}
      \addplot+[ybar, fill=mode-thinking, fill opacity=0.6,
                draw=mode-thinking!70!black, mark=none]
        coordinates {(Def,16.99)(Ex,6.29)(Lem,18.15)(Thm,9.60)};
      \addlegendentry{Thinking}
      \addplot+[ybar, fill=mode-agentic, fill opacity=0.6,
                draw=mode-agentic!70!black, mark=none]
        coordinates {(Def,18.06)(Ex,7.94)(Lem,19.80)(Thm,9.62)};
      \addlegendentry{Agentic}
    \end{axis}
  \end{tikzpicture}
  \vspace{-17px}
  \caption{Pass rate by item kind.}
  \label{fig:rq2-kind-main}
\end{subfigure}
\vspace{-5px}
\caption{IAF pass-rate breakdowns by chapter and item kind across three inference modes.}
\label{fig:rq2-main}
\end{minipage}
\hfill
\begin{minipage}{0.37\linewidth}
\centering
\begin{tikzpicture}
\begin{axis}[
  width=1.12\linewidth, height=4.2cm,
  ymin=0, ymax=130, ymajorgrids, grid style={dashed, gray!30},
  ylabel={Proof rate (\%)},
  xlabel style={font=\footnotesize},
  ylabel style={font=\footnotesize, at={(axis description cs:0.18,.5)}, anchor=south},
  x tick label style={font=\scriptsize},
  y tick label style={font=\scriptsize},
  xtick={1,5,10,15,20},
  ytick={0,20,40,60,80,100},
  enlarge x limits=0.04,
  clip=false,
  legend style={
    at={(0.0,1)}, anchor=north west,
    draw=black!30, fill=white, fill opacity=0.9, text opacity=1,
    font=\scriptsize, inner sep=2pt, row sep=-3pt,
    legend columns=2,
  },
  legend cell align=left,
]
  \addplot+[color=mode-agentic, mark=*, mark options={fill=mode-agentic, scale=0.7}, mark repeat=5, mark phase=4, line width=0.9pt]
    coordinates {(1,0.0)(2,13.6)(3,29.5)(4,34.3)(5,45.0)(6,48.1)(7,53.8)(8,54.8)(9,57.4)(10,59.0)(11,62.9)(12,64.0)(13,66.9)(14,68.3)(15,70.0)(16,71.4)(17,72.4)(18,73.8)(19,74.0)(20,74.5)};
  \addlegendentry{DTP / CO46}
  \addplot+[color=mode-zeroshot, mark=*, mark options={fill=mode-zeroshot, scale=0.7}, mark repeat=5, mark phase=4, line
  width=0.9pt]
    coordinates {(1,0.0)(2,7.4)(3,7.4)(4,7.4)(5,8.6)(6,12.3)(7,13.6)(8,16.0)(9,18.5)(10,22.2)(11,22.2)(12,24.7)(13,27.2)(14,27.2)(15,29.6)(16,29.6)(17,29.6)(18,30.9)(19,32.1)(20,33.3)};
  \addlegendentry{ITP / CO46}
  \addplot+[color=mode-agentic, mark=square*, mark options={fill=mode-agentic, scale=0.7}, mark repeat=5, mark phase=4, line width=0.9pt, densely dashed]
    coordinates {(1,0.0)(2,24.5)(3,25.5)(4,36.2)(5,38.6)(6,43.6)(7,44.8)(8,47.6)(9,48.1)(10,51.0)(11,52.4)(12,55.5)(13,56.4)(14,58.3)(15,59.3)(16,60.0)(17,60.2)(18,60.5)(19,61.0)(20,62.1)};
  \addlegendentry{DTP / CS46}
  \addplot+[color=mode-zeroshot, mark=square*, mark options={fill=mode-zeroshot, scale=0.7}, mark repeat=5, mark phase=4, line width=0.9pt, densely dashed]
    coordinates {(1,0.0)(2,1.2)(3,1.2)(4,2.5)(5,3.7)(6,6.2)(7,8.6)(8,9.9)(9,9.9)(10,9.9)(11,9.9)(12,9.9)(13,9.9)(14,11.1)(15,11.1)(16,12.3)(17,12.3)(18,13.6)(19,14.8)(20,14.8)};
  \addlegendentry{ITP / CS46}
  \addplot+[color=mode-agentic, mark=triangle*, mark options={fill=mode-agentic, scale=0.7}, mark repeat=5, mark phase=4, line width=0.9pt, dotted]
    coordinates {(1,0.0)(2,8.1)(3,18.8)(4,22.9)(5,27.9)(6,30.5)(7,32.1)(8,34.8)(9,36.4)(10,39.5)(11,41.0)(12,41.9)(13,42.6)(14,43.6)(15,44.8)(16,45.5)(17,46.2)(18,46.9)(19,47.6)(20,48.1)};
  \addlegendentry{DTP / KK25}
  \addplot+[color=mode-zeroshot, mark=triangle*, mark options={fill=mode-zeroshot, scale=0.7}, mark repeat=5, mark phase=4, line width=0.9pt, dotted]
    coordinates {(1,0.0)(2,1.2)(3,1.2)(4,3.7)(5,3.7)(6,4.9)(7,4.9)(8,4.9)(9,6.2)(10,6.2)(11,6.2)(12,6.2)(13,6.2)(14,6.2)(15,7.4)(16,7.4)(17,7.4)(18,7.4)(19,7.4)(20,7.4)};
  \addlegendentry{ITP / KK25}
\end{axis}
\end{tikzpicture}
\vspace{-18px}
\captionof{figure}{Cumulative proof rate vs.\ \# turns for DTP and ITP, across three models: CO46, CS46, and KK25.}
\label{fig:tp-passrate-curve}
\vspace{-5px}
\end{minipage}
\end{figure}

\noindent\textbf{RQ2: How does performance vary with textbook difficulty?}
Figure~\ref{fig:rq2-main} decomposes IAF pass rates by chapter and by item kind.
Chapter-wise (Figure~\ref{fig:rq2-chapter-main}), models perform best on Chapters~3 and~5, the foundational chapters on propositional and first-order logic.
Performance drops sharply to near 0\% on Chapters~7--8, which cover Gentzen's Sharpened Hauptsatz, Herbrand's theorem, and resolution, demanding much deeper abstract reasoning.
By item kind (Figure~\ref{fig:rq2-kind-main}), definitions are one of the easier categories, while examples and theorems lag behind, consistent with the intuition that theorems require more nuanced reasoning to construct a logically valid statement.

\noindent\textbf{RQ3: Can LLMs do theorem proving for LCS?}
As shown in Figure~\ref{fig:tp-passrate-curve}, Claude Opus 4.6 reaches a 74.5\% proof rate on DTP within 20 agentic turns.
While a strong result, it is partly attributable to DTP's design: each goal is a bite-sized sub-problem decomposed from larger proofs, and the proof steps are largely mechanical.
ITP is harder, demanding long-context reasoning over full proof states, yet Opus 4.6 still reaches 33.3\% at $T{=}20$ with no clear plateau, suggesting further gains are likely.

\begin{figure}[t]
\begin{minipage}[t]{0.49\linewidth}
  \centering
\resizebox{\linewidth}{!}{
\begin{tikzpicture}
\begin{axis}[
  ybar stacked,
  width=1.1\linewidth, height=4.0cm,
  ymin=0, ymax=100, ytick distance=20,
  ymajorgrids, grid style={dashed, gray!30},
  ylabel={Pass / Compile rate (\%)},
  ylabel style={font=\scriptsize, at={(axis description cs:0.13,.5)}, anchor=south},
  enlarge x limits=0.05,
  symbolic x coords={
    zs01,zs02,zs03,zs04,zs05,zs06,zs07,zs08,zs09,
    gap1,
    th01,th02,th03,th04,th05,th06,th07,
    gap2,
    ag01,ag02,ag03,ag04,ag05,ag06
  },
  xtick={zs01,zs02,zs03,zs04,zs05,zs06,zs07,zs08,zs09,
         th01,th02,th03,th04,th05,th06,th07,
         ag01,ag02,ag03,ag04,ag05,ag06},
  xticklabels={CS46,CO46,CH45,KK25,DS32,D2,GLM5,L4,Q3C,
               KK25,CO46,CS46,MM25,Q32,DSR1,GLM5,
               CO46,GLM5,KK25,CS46,MM25,Q32},
  x tick label style={font=\tiny, rotate=45, anchor=north east},
  y tick label style={font=\scriptsize},
  bar width=6pt,
  clip=false,
  legend style={
    at={(0,1)}, anchor=north west,
    draw=black!30, fill=white, fill opacity=0.9, text opacity=1,
    font=\scriptsize, inner sep=3pt, row sep=-1pt,
    legend columns=2, column sep=1ex,
  },
  legend cell align=left,
  legend image code/.code={
    \draw[#1] (0cm,-0.08cm) rectangle (0.22cm,0.08cm);
  },
]

  \addplot+[fill=mode-zeroshot, fill opacity=0.6, draw=mode-zeroshot!70!black]
    coordinates {
      (zs01,5.2)(zs02,2.6)(zs03,2.6)(zs04,2.6)(zs05,1.7)(zs06,1.7)(zs07,0.9)(zs08,0.9)(zs09,0.0)
      (gap1,0)
      (th01,0)(th02,0)(th03,0)(th04,0)(th05,0)(th06,0)(th07,0)
      (gap2,0)
      (ag01,0)(ag02,0)(ag03,0)(ag04,0)(ag05,0)(ag06,0)
    };
  \addlegendentry{Zero-shot}
  \addplot+[fill=mode-zeroshot, fill opacity=0.2, draw=mode-zeroshot!70!black, forget plot]
    coordinates {
      (zs01,35.3)(zs02,47.4)(zs03,17.2)(zs04,11.2)(zs05,16.4)(zs06,6.1)(zs07,9.4)(zs08,12.0)(zs09,0.0)
      (gap1,0)
      (th01,0)(th02,0)(th03,0)(th04,0)(th05,0)(th06,0)(th07,0)
      (gap2,0)
      (ag01,0)(ag02,0)(ag03,0)(ag04,0)(ag05,0)(ag06,0)
    };

  \addplot+[fill=mode-thinking, fill opacity=0.6, draw=mode-thinking!70!black]
    coordinates {
      (zs01,0)(zs02,0)(zs03,0)(zs04,0)(zs05,0)(zs06,0)(zs07,0)(zs08,0)(zs09,0)
      (gap1,0)
      (th01,6.9)(th02,4.3)(th03,2.6)(th04,1.7)(th05,1.7)(th06,0.9)(th07,0.9)
      (gap2,0)
      (ag01,0)(ag02,0)(ag03,0)(ag04,0)(ag05,0)(ag06,0)
    };
  \addlegendentry{Thinking}
  \addplot+[fill=mode-thinking, fill opacity=0.2, draw=mode-thinking!70!black, forget plot]
    coordinates {
      (zs01,0)(zs02,0)(zs03,0)(zs04,0)(zs05,0)(zs06,0)(zs07,0)(zs08,0)(zs09,0)
      (gap1,0)
      (th01,25.0)(th02,52.6)(th03,39.6)(th04,3.5)(th05,4.3)(th06,3.4)(th07,8.6)
      (gap2,0)
      (ag01,0)(ag02,0)(ag03,0)(ag04,0)(ag05,0)(ag06,0)
    };

  \addplot+[fill=mode-agentic, fill opacity=0.6, draw=mode-agentic!70!black]
    coordinates {
      (zs01,0)(zs02,0)(zs03,0)(zs04,0)(zs05,0)(zs06,0)(zs07,0)(zs08,0)(zs09,0)
      (gap1,0)
      (th01,0)(th02,0)(th03,0)(th04,0)(th05,0)(th06,0)(th07,0)
      (gap2,0)
      (ag01,5.2)(ag02,4.3)(ag03,3.4)(ag04,3.4)(ag05,2.6)(ag06,0.0)
    };
  \addlegendentry{Agentic}
  \addplot+[fill=mode-agentic, fill opacity=0.2, draw=mode-agentic!70!black, forget plot]
    coordinates {
      (zs01,0)(zs02,0)(zs03,0)(zs04,0)(zs05,0)(zs06,0)(zs07,0)(zs08,0)(zs09,0)
      (gap1,0)
      (th01,0)(th02,0)(th03,0)(th04,0)(th05,0)(th06,0)(th07,0)
      (gap2,0)
      (ag01,91.4)(ag02,92.3)(ag03,88.8)(ag04,93.2)(ag05,82.7)(ag06,0.9)
    };
\end{axis}
\end{tikzpicture}
}
\vspace{-20px}
\captionof{figure}{SSAF results per model-mode. Each stacked bar shows the pass rate (bottom, solid) and the additional compile-only rate (top, faded).}
\label{fig:ssaf-passrate}
\vspace{-10px}
\end{minipage}
\hfill
\begin{minipage}[t]{0.49\linewidth}
  \centering
\resizebox{\linewidth}{!}{
\begin{tikzpicture}
\begin{axis}[
  ybar=0pt,
  width=1.1\linewidth, height=4.0cm,
  ymin=0, ymax=22, ytick distance=5,
  ymajorgrids, grid style={dashed, gray!30},
  ylabel={Pass rate (\%)},
  ylabel style={font=\scriptsize, at={(axis description cs:0.13,.5)}, anchor=south},
  enlarge x limits=0.13,
  symbolic x coords={
    zs01,zs02,zs03,zs04,zs05,zs06,zs07,zs08,zs09,
  },
  xtick={zs01,zs02,zs03,zs04,zs05,zs06,zs07,zs08,zs09},
  xticklabels={CO46,CS46,GLM5,CH45,DS32,KK25,D2,L4,Q3C},
  x tick label style={font=\tiny, rotate=45, anchor=north east},
  y tick label style={font=\scriptsize},
  xticklabel shift=-4pt,
  bar width=5pt,
  clip=false,
  legend style={
    at={(1,1)}, anchor=north east,
    draw=black!30, fill=white, fill opacity=0.9, text opacity=1,
    font=\scriptsize, inner sep=3pt, row sep=-1pt,
    legend columns=2, column sep=1ex,
  },
  legend cell align=left,
  legend image code/.code={
    \draw[#1] (0cm,-0.08cm) rectangle (0.22cm,0.08cm);
  },
]
  \addplot+[fill=codebox-frame, fill opacity=0.6, draw=codebox-frame!70!black]
    coordinates {
      (zs01,16.8)(zs02,15.0)(zs03,13.9)(zs04,11.1)(zs05, 9.5)(zs06, 9.0)(zs07, 8.4)(zs08, 7.6)(zs09, 3.1)
    };
  \addlegendentry{IAF}
  \addplot+[fill=mode-agentic, fill opacity=0.6, draw=mode-agentic!70!black]
    coordinates {
      (zs01,14.1)(zs02,12.8)(zs03,10.7)(zs04, 9.8)(zs05, 7.3)(zs06, 7.0)(zs07, 7.6)(zs08, 6.4)(zs09, 3.4)
    };
  \addlegendentry{IAF-D}

  \node[font=\tiny, text=red!70!black]  at (axis cs:zs01,16.8) [above, yshift=1pt] {$-2.7$};
  \node[font=\tiny, text=red!70!black]  at (axis cs:zs02,15.0) [above, yshift=1pt] {$-2.2$};
  \node[font=\tiny, text=red!70!black]  at (axis cs:zs03,13.9) [above, yshift=1pt] {$-3.2$};
  \node[font=\tiny, text=red!70!black]  at (axis cs:zs04,11.1) [above, yshift=1pt] {$-1.3$};
  \node[font=\tiny, text=red!70!black]  at (axis cs:zs05, 9.5) [above, yshift=1pt] {$-2.2$};
  \node[font=\tiny, text=red!70!black]  at (axis cs:zs06, 9.0) [above, yshift=1pt] {$-2.0$};
  \node[font=\tiny, text=red!70!black]  at (axis cs:zs07, 8.4) [above, yshift=1pt] {$-0.8$};
  \node[font=\tiny, text=red!70!black]  at (axis cs:zs08, 7.6) [above, yshift=1pt] {$-1.2$};
  \node[font=\tiny, text=blue!70!black] at (axis cs:zs09, 3.4) [above, yshift=1pt] {$+0.3$};
\end{axis}
\end{tikzpicture}
}
\vspace{-20px}
\captionof{figure}{
  IAF vs.\ IAF-D pass rates per model under zero-shot mode.
  Injecting distractors consistently degrades pass rate.
}
\label{fig:iafd-comparison}
\vspace{-10px}
\end{minipage}
\end{figure}

\noindent\textbf{RQ4: How does performance react to theory-scale setups?}
Scaling up the formalization context exposes a sharp capability gap.
Figure~\ref{fig:ssaf-passrate} reports results on SSAF, where each item is a full textbook sub-section averaging 2.82 items.
Pass rates collapse to single digits for all models.
While adopting an agentic framework raises the \emph{compilation} rate, it does not translate into a pass-rate gain, suggesting the agentic loop helps surface syntax errors but cannot fix semantic mismatches.
On IAF-D, adding distractors consistently degrades pass rates across all models (Figure~\ref{fig:iafd-comparison}), even though the intrinsic difficulty of the target item is unchanged.
This reveals that current models struggle to identify the critical information within a longer context, which is a prerequisite for theory-scale auto-formalization where relevant definitions are always embedded in the surrounding material.

\begin{figure}[t]
\centering
\begin{subfigure}[t]{0.511\linewidth}
  \includegraphics[width=\linewidth]{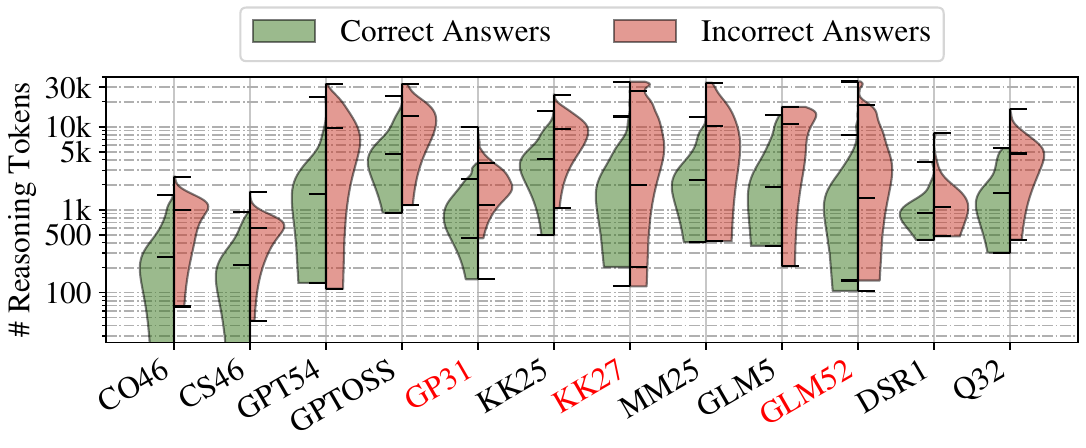}
  \vspace{-20px}
  \caption{Thinking: reasoning tokens per attempt.}
  \label{fig:iaf-ttc-thinking}
\end{subfigure}\hfill
\begin{subfigure}[t]{0.464\linewidth}
  \includegraphics[width=\linewidth]{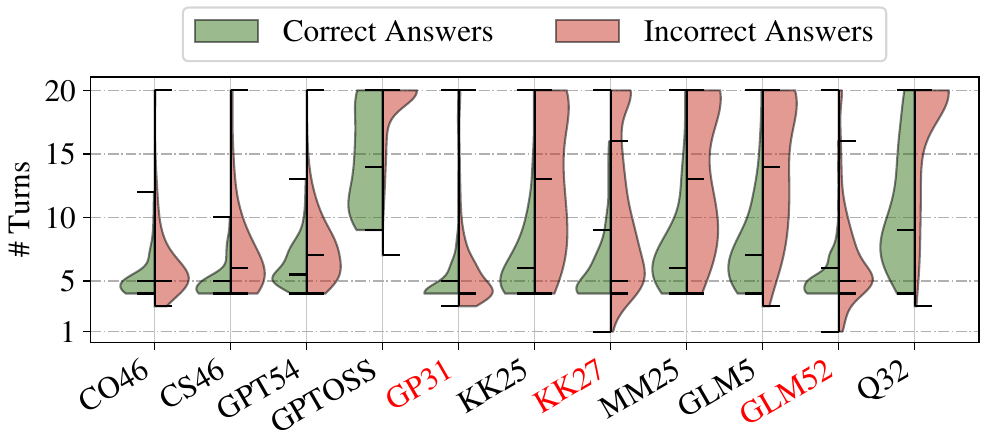}
  \vspace{-20px}
  \caption{Agentic: turns per attempt.}
  \label{fig:iaf-ttc-agentic}
\end{subfigure}
\vspace{-5px}
\caption{Violin plots showing the relationship between test-time compute vs. correctness on IAF.}
\label{fig:iaf-ttc}
\vspace{-5px}
\end{figure}

\noindent\textbf{RQ5: Does test-time compute drive better formalization?}
Figure~\ref{fig:iaf-ttc-thinking} shows that, in thinking mode, incorrect outputs are associated with substantially more reasoning tokens than correct ones.
Models that fail tend to ``overthink'' rather than converging.
High-performing models (Gemini 3.1 Pro, Claude Opus 4.6, Sonnet 4.6, GPT-5.4) consistently use fewer reasoning tokens than lower-performing ones, yet produce better results.
The same pattern holds in agentic mode (Figure~\ref{fig:iaf-ttc-agentic}): top models succeed or fail decisively in fewer turns, while weaker models burn more interactions.
Together, these observations suggest that scaling test-time compute alone is not a viable path to auto-formalization and that progress demands deeper innate capability.

\begin{figure}[t]
  \begin{minipage}{0.32\linewidth}
    \definecolor{err-pass}  {RGB}{56,118,29}
    \definecolor{err-nmexpr}{RGB}{180, 55, 40}
    \definecolor{err-cmiss} {RGB}{216,190,148}
    \definecolor{err-nmkind}{RGB}{180,130, 78}
    \definecolor{err-comp}  {RGB}{160,160,160}
    \definecolor{err-tmo}   {RGB}{220,220,220}

    \centering
    \begin{tikzpicture}
      \begin{axis}[
        xbar stacked,
        xmin=0, xmax=100,
        xtick=\empty,
        xmajorgrids, grid style={dashed, gray!30},
        y tick label style={font=\tiny},
        tick align=outside,
        width=\linewidth, height=3.2cm,
        bar width=6pt,
        enlarge y limits=0.18,
        symbolic y coords={zs-CO46, zs-GLM5, ag-MM25, ag-GLM5, ag-CO46},
        ytick=data,
        yticklabels={{CO46 (ZS)},{GLM5 (ZS)},{MM25 (AG)},{GLM5 (AG)},{CO46 (AG)}},
        legend style={
          at={(1,-0.03)}, anchor=north east,
          draw=black!30, fill=white, fill opacity=0.9, text opacity=1,
          font=\scriptsize, inner sep=1pt, row sep=-2pt,
          legend columns=3, column sep=0.4ex,
        },
        legend cell align=left,
        legend image code/.code={
          \draw[#1] (0cm,-0.06cm) rectangle (0.18cm,0.06cm);
        },
      ]
        \addplot+[fill=err-pass!60,   draw=err-pass!70!black,   mark=none]
          coordinates {(16.8,zs-CO46)(13.9,zs-GLM5)(12.1,ag-MM25)(14.7,ag-GLM5)(20.1,ag-CO46)};
        \addplot+[fill=mode-agentic!70, draw=mode-agentic!70!black, mark=none]
          coordinates {(53.7,zs-CO46)(26.9,zs-GLM5)(62.1,ag-MM25)(45.1,ag-GLM5)(77.9,ag-CO46)};
        \addplot+[fill=err-cmiss,  draw=err-cmiss!70!black,  mark=none]
          coordinates {(0.0,zs-CO46)(0.2,zs-GLM5)(3.5,ag-MM25)(0.3,ag-GLM5)(0.3,ag-CO46)};
        \addplot+[fill=err-nmkind, draw=err-nmkind!70!black, mark=none]
          coordinates {(0.0,zs-CO46)(0.0,zs-GLM5)(1.0,ag-MM25)(0.9,ag-GLM5)(1.1,ag-CO46)};
        \addplot+[fill=err-comp,   draw=err-comp!70!black,   mark=none]
          coordinates {(29.5,zs-CO46)(58.9,zs-GLM5)(21.3,ag-MM25)(39.0,ag-GLM5)(0.6,ag-CO46)};
        \addplot+[fill=err-tmo,    draw=err-tmo!70!black,    mark=none]
          coordinates {(0.0,zs-CO46)(0.1,zs-GLM5)(0.0,ag-MM25)(0.0,ag-GLM5)(0.0,ag-CO46)};
        \legend{Pass, NMExpr, CMiss, NMKind, CompFail, Timeout}
      \end{axis}
    \end{tikzpicture}
    \vspace{-15pt}
    \captionof{figure}{
      IAF error-type breakdown for representative models.
    }
    \label{fig:iaf-err-main}
  \end{minipage}
  \hfill
  \begin{minipage}{0.65\linewidth}
    \centering
    \begin{tikzpicture}[every node/.style={font=\scriptsize}]
      \def\bh{0.28}\def\sp{0.33}\def\sc{9}
      \def\labL{-3.15}
      \pgfmathsetmacro{\ytop}{0.10}
      \pgfmathsetmacro{\ybot}{-5*\sp+\sp-\bh-0.18}
      \foreach \v in {-0.3,-0.2,-0.1,0,0.1,0.2,0.3}{
        \pgfmathsetmacro{\x}{\v*\sc}
        \draw[gray!30, thin] (\x, \ytop) -- (\x, \ybot);
      }
      \draw[black] (-3.0,\ytop) rectangle (3.0,\ybot);
      \draw[gray!30, thick] (0, \ytop) -- (0, \ybot);
      \foreach \v in {-0.3,-0.2,-0.1,0,0.1,0.2,0.3}{
        \pgfmathsetmacro{\x}{\v*\sc}
        \draw[gray!60] (\x,\ybot)--(\x,\ybot-0.10);
        \node[anchor=north, font=\tiny] at (\x,\ybot-0.11) {\v};
      }
      \foreach \tag/\d/\col [count=\i from 1] in {%
        {References a figure}/-0.161/mode-agentic,
        {Contains informal notation}/-0.159/mode-agentic,
        {Presents a concrete instance}/-0.136/mode-agentic,
        {Uses a metavariable}/+0.081/mode-zeroshot,
        {Short item ($<$25 tokens)}/+0.125/mode-zeroshot%
      }{
        \pgfmathsetmacro{\y}{-\i*\sp}
        \pgfmathsetmacro{\x}{\d*\sc}
        \fill[\col, fill opacity=0.6] (0,\y) rectangle (\x,\y+\bh);
        \draw[\col!70!black] (0,\y) rectangle (\x,\y+\bh);
        \node[anchor=east, font=\scriptsize] at (\labL,\y+\bh/2) {\tag};
        \ifdim \x pt < 0pt
          \pgfmathsetmacro{\xout}{\x-0.10}
          \node[anchor=east, font=\tiny, text=red!30!black] at (\xout,\y+\bh/2) {$\d$};
        \else
          \pgfmathsetmacro{\xout}{\x+0.10}
          \node[anchor=west, font=\tiny, text=black] at (\xout,\y+\bh/2) {$\d$};
        \fi
      }
    \end{tikzpicture}
    \vspace{-15pt}
    \captionof{figure}{
      Pass-rate difference (items \emph{with} tag minus \emph{without}) for selected feature tags on IAF evaluation results with CO46.
    }
    \label{fig:passdiff-surface-main}
  \end{minipage}
  \vspace{-10px}
\end{figure}

\noindent\textbf{RQ6: What causes formalization failures?}
Figure~\ref{fig:iaf-err-main} breaks down error types.
Even in agentic mode, only a few of the strongest models nearly eliminate compilation failures; weaker models still struggle with compilation.
Among compiled-but-failed submissions, the overwhelming category is NMExpr (non-matching expression): the generated term compiles and has the right syntactic shape, but is not semantically equivalent.
Missing constants (CMiss) and kind mismatches (NMKind) occur very rarely, indicating that models generally follow the structural template.

To understand what makes an item semantically hard, we annotate items along many feature axes and measure the per-tag pass-rate difference (with selected features shown in Figure~\ref{fig:passdiff-surface-main}).
Notably, items that reference a figure or use informal notation are the hardest to formalize: their intended semantics must be inferred from heterogeneous presentation.
Conversely, short items and those that introduce a named object are easier, consistent with models being most reliable on self-contained, concisely stated definitions.
A full per-dimension breakdown is given in Appendix~\ref{sec:appendix-tagging-analysis}.

\vspace{-5px}
\section{Related Works}
\label{sec:related-works}
\vspace{-5px}

\textbf{Autoformalization and Theorem-Proving Benchmarks.}
Existing Lean-based benchmarks target isolated mathematical statements: competition-math~\cite{zheng2022minif2f,tsoukalas2024putnambench,liu2023fimo}, textbook and domain-specific suites~\cite{azerbayev2023proofnet,jiang2025fate,liu2025combibench,xu2025leancat,yang2025formalml}, and repo-scale proof-search benchmarks~\cite{yang2023leandojo,hu2024minictx}.
Most rely on Mathlib as prebuilt infrastructure and treat libraries as fixed context.
Autoformalization datasets and methods~\cite{wu2022autoformalization,liu2025atlas,wang2025aria,lu2024process,zhang2025drift,guo2025autoformalizer,Zhang2024consistent,min2026divide,yang2024formal,li2024survey,weng2025autoformalization} similarly focus on single-statement or definition-level translation.
Concurrent theory-level efforts~\cite{min2026theorylevel}, such as M2F~\cite{wang2026m2f}, Atlas~\cite{rammal2026formalizing}, LeanMarathon~\cite{zhang2026leanmarathon}, and Gauss~\cite{mathinc2025gauss}, formalize mathematical literature at repository scale.
\ours differs by mechanizing an \emph{entire} CS textbook as a stand-alone Lean~4 repository covering heterogeneous declaration kinds completely independent from Mathlib.

\textbf{Formalizing Computer Science.}
Prior human-written CS formalizations span verified compilers~\cite{leroy2009formal,zhao2013formal}, OS kernels~\cite{klein2009sel4,gu2016certikos}, distributed systems~\cite{hawblitzel2015ironfleet,wilcox2015verdi}, and the DeepSpec expedition~\cite{appel2017deepspec,appel2014program}.
Machine-checked textbooks~\cite{pierce2010software,nipkow2014concrete,chlipala2013certified} show theory-scale CS formalization is feasible but requires years of expert effort.
Logic-specific mechanizations~\cite{from2022secav,from2020teaching,herbelin2024analysis,aydemir2005mechanized} are closely related to \ours's content.
\ours repositions this tradition as an ML-evaluable task.

\textbf{Neurosymbolic and Agentic Approaches.}
\ours is motivated by neurosymbolic programming~\cite{li2023scallop,huang2021scallop,li2024neurosymbolic,manhaeve2018deepproblog,yang2020neurasp}, in which neural agents produce structured symbolic artifacts that can be formally verified.
NL-to-formal-specification work for planning and probabilistic programs~\cite{zhang2024proc2pddl,huang2025limit,bingham2019pyro,carpenter2017stan,deraedt2007problog} provides evidence that LLMs can translate informal descriptions into verifiable representations.
Code-oriented agentic frameworks~\cite{jimenez2024swe,yang2024swe,wang2025openhands} establish the tool-equipped-agent design pattern.
LLM-based provers~\cite{xin2024deepseek,xin2024deepseekv15,ren2025deepseek,lin2025goedel,wang2025kimina,jiang2023draft,wang2023lego,thakur2023copra,lample2022hypertree} supply natural baselines for our TP tracks.

\textbf{Evaluation Metrics for Autoformalization.}
Standard metrics use type-checking (pass@$k$) or logical-equivalence checks.
BEq+~\cite{poiroux2025reliable} and related approaches~\cite{li2024autoformalize,zhang2025autoformalization,moore2025evaluating,lu2024formalalign} address statement-level semantic consistency but cannot handle inductive types, structures, or constructor-bearing definitions.
Recent audits~\cite{meek2026numerical,ilin2026sorries} further show that kernel acceptance overstates formalization quality.
Our \emph{definitional equivalence checker} fills this gap.

\vspace{-5px}
\section{Limitations and Conclusion}
\label{sec:conclusion}
\vspace{-5px}

We present \ours, a Lean~4 benchmark derived from \emph{Logic for Computer Science} for evaluating LLM and agent capabilities on auto-formalization and theorem proving, independent of any existing Lean library.
Our evaluations show that current models struggle to solve the benchmark, with performance degrading sharply as difficulty increases.

\textbf{Limitations.}
A few textbook theorems resisted both human and LLM-agent proof attempts; at the current level of deep embedding, we cannot fully capture the textbook's complexity, and future work may adopt mature libraries to reach more advanced results.
\ours is also scoped to a single textbook; extending it across multiple CS textbooks and domains would broaden source diversity and strengthen the generalizability of our semi-automated pipeline and experimental findings.
Our DefEq checker is also conservative and incomplete; reported scores are therefore reference-relative and conservatively estimate open-ended auto-formalization success.

\section*{Licensing, Permission, and Release}
\label{sec:licensing}
\ours is derived from Jean H.~Gallier's \emph{Logic for Computer Science: Foundations of Automatic Theorem Proving}; we extract the excerpts from the author-hosted PDF of the Dover second edition~\cite{gallier2015logic}, \copyright~1986, 2015 J.~Gallier.
Prof.~Gallier granted us written permission to reproduce and, where necessary for formalization, lightly adapt the book's statements, and to distribute them free of charge for research and educational use.
The excerpts are item statements only, without proofs or connecting exposition, and amount to approximately 11.8\% of the book by character count; the Lean artifact is newly written formalization derived from the book's technical content.
The benchmark data is released under CC BY 4.0 except for the excerpts, whose copyright remains with the author under the permission notice, with no sublicensing rights; the code and Lean artifact are released under MIT.
The release includes \texttt{DATASHEET.md} and \texttt{LICENSE-DATA}; the dataset is hosted on Kaggle at \lcskaggle, and the code and Lean artifact on GitHub at \lcsgit.

\begin{ack}
Research reported in this publication was supported by an Amazon Research Award, Fall 2025, and by the 2026 Amazon Nova AI Challenge: Trusted Software Agents.
The views and conclusions contained herein are those of the authors and should not be interpreted as necessarily representing the official policies of the supporting sponsors.
The authors declare no competing interests.
\end{ack}

\bibliographystyle{unsrtnat}
\bibliography{references}

\begin{thebibliography}{86}
\providecommand{\natexlab}[1]{#1}
\providecommand{\url}[1]{\texttt{#1}}
\expandafter\ifx\csname urlstyle\endcsname\relax
  \providecommand{\doi}[1]{doi: #1}\else
  \providecommand{\doi}{doi: \begingroup \urlstyle{rm}\Url}\fi

\bibitem[Wu et~al.(2022)Wu, Jiang, Li, Rabe, Staats, Jamnik, and
  Szegedy]{wu2022autoformalization}
Yuhuai Wu, Albert~Q. Jiang, Wenda Li, Markus~N. Rabe, Charles Staats, Mateja
  Jamnik, and Christian Szegedy.
\newblock Autoformalization with large language models.
\newblock In \emph{Advances in Neural Information Processing Systems
  (NeurIPS)}, 2022.

\bibitem[Yang et~al.(2024{\natexlab{a}})Yang, Poesia, He, Li, Lauter,
  Chaudhuri, and Song]{yang2024formal}
Kaiyu Yang, Gabriel Poesia, Jingxuan He, Wenda Li, Kristin Lauter, Swarat
  Chaudhuri, and Dawn Song.
\newblock Formal mathematical reasoning: A new frontier in {AI}.
\newblock \emph{arXiv preprint}, 2024{\natexlab{a}}.

\bibitem[Li et~al.(2024{\natexlab{a}})Li, Sun, Murphy, Su, Li, Zhang, Yang, and
  Si]{li2024survey}
Zhaoyu Li, Jialiang Sun, Logan Murphy, Qidong Su, Zenan Li, Xian Zhang, Kaiyu
  Yang, and Xujie Si.
\newblock A survey on deep learning for theorem proving.
\newblock \emph{arXiv preprint}, 2024{\natexlab{a}}.

\bibitem[Weng et~al.(2025)Weng, Du, Li, Lu, Sun, Liu, and
  Zhang]{weng2025autoformalization}
Ke~Weng, Lun Du, Sirui Li, Wangyue Lu, Haozhe Sun, Hengyu Liu, and Tiancheng
  Zhang.
\newblock Autoformalization in the era of large language models: A survey.
\newblock \emph{arXiv preprint}, 2025.

\bibitem[mathlib Community(2020)]{mathlib2020}
The mathlib Community.
\newblock The lean mathematical library.
\newblock In \emph{Proceedings of the 9th ACM SIGPLAN International Conference
  on Certified Programs and Proofs}, CPP 2020, pages 367--381, New York, NY,
  USA, 2020. Association for Computing Machinery.
\newblock ISBN 9781450370974.
\newblock \doi{10.1145/3372885.3373824}.
\newblock URL \url{https://doi.org/10.1145/3372885.3373824}.

\bibitem[Zheng et~al.(2022)Zheng, Han, and Polu]{zheng2022minif2f}
Kunhao Zheng, Jesse~Michael Han, and Stanislas Polu.
\newblock {MiniF2F}: a cross-system benchmark for formal {O}lympiad-level
  mathematics.
\newblock In \emph{International Conference on Learning Representations
  (ICLR)}, 2022.

\bibitem[Azerbayev et~al.(2023)Azerbayev, Piotrowski, Schoelkopf, Ayers, Radev,
  and Avigad]{azerbayev2023proofnet}
Zhangir Azerbayev, Bartosz Piotrowski, Hailey Schoelkopf, Edward~W. Ayers,
  Dragomir Radev, and Jeremy Avigad.
\newblock {ProofNet}: Autoformalizing and formally proving undergraduate-level
  mathematics.
\newblock \emph{arXiv preprint}, 2023.

\bibitem[Liu et~al.(2023)Liu, Shen, Xin, Liu, Yuan, Wang, Ju, Zheng, Yin, Li,
  Zhang, and Liu]{liu2023fimo}
Chengwu Liu, Jianhao Shen, Huajian Xin, Zhengying Liu, Ye~Yuan, Haiming Wang,
  Wei Ju, Chuanyang Zheng, Yichun Yin, Lin Li, Ming Zhang, and Qun Liu.
\newblock {FIMO}: A challenge formal dataset for automated theorem proving.
\newblock \emph{arXiv preprint}, 2023.

\bibitem[Jiang et~al.(2025)Jiang, He, Wang, Gao, Hu, Wang, Guan, Wu, Dai, Xiao,
  and Dong]{jiang2025fate}
Jiedong Jiang, Wanyi He, Yuefeng Wang, Guoxiong Gao, Yongle Hu, Jingting Wang,
  Nailin Guan, Peihao Wu, Chunbo Dai, Liang Xiao, and Bin Dong.
\newblock {FATE}: A formal benchmark series for frontier algebra of multiple
  difficulty levels.
\newblock \emph{arXiv preprint}, 2025.

\bibitem[Liu et~al.(2025{\natexlab{a}})Liu, Lin, Bayer, Dillies, Jiang, Liang,
  Soletskyi, Wang, Xie, Xiong, Yang, Zhang, Zhi, Li, and
  Liu]{liu2025combibench}
Junqi Liu, Xiaohan Lin, Jonas Bayer, Ya{\"e}l Dillies, Weijie Jiang, Xiaodan
  Liang, Roman Soletskyi, Haiming Wang, Yunzhou Xie, Beibei Xiong, Zhengfeng
  Yang, Jujian Zhang, Lihong Zhi, Jia Li, and Zhengying Liu.
\newblock {CombiBench}: Benchmarking {LLM} capability for combinatorial
  mathematics.
\newblock \emph{arXiv preprint}, 2025{\natexlab{a}}.

\bibitem[Xu et~al.(2025)Xu, Dai, Fu, Jiang, Nie, Wang, Yang, and
  Zhang]{xu2025leancat}
Rongge Xu, Hui Dai, Yiming Fu, Jiedong Jiang, Tianjiao Nie, Junkai Wang,
  Holiverse Yang, and Zhi-Hao Zhang.
\newblock {LeanCat}: A benchmark suite for formal category theory in {L}ean
  (part {I}: 1-categories).
\newblock \emph{arXiv preprint}, 2025.

\bibitem[Tsoukalas et~al.(2024)Tsoukalas, Lee, Jennings, Xin, Ding, Jennings,
  Thakur, and Chaudhuri]{tsoukalas2024putnambench}
George Tsoukalas, Jasper Lee, John Jennings, Jimmy Xin, Michelle Ding, Michael
  Jennings, Amitayush Thakur, and Swarat Chaudhuri.
\newblock {PUTNAMBENCH}: evaluating neural theorem-provers on the putnam
  mathematical competition.
\newblock In \emph{Proceedings of the 38th International Conference on Neural
  Information Processing Systems}, NIPS '24, Red Hook, NY, USA, 2024. Curran
  Associates Inc.
\newblock ISBN 9798331314385.

\bibitem[Yang et~al.(2023)Yang, Swope, Gu, Chalamala, Song, Yu, Godil, Prenger,
  and Anandkumar]{yang2023leandojo}
Kaiyu Yang, Aidan~M. Swope, Alex Gu, Rahul Chalamala, Peiyang Song, Shixing Yu,
  Saad Godil, Ryan Prenger, and Anima Anandkumar.
\newblock {LeanDojo}: Theorem proving with retrieval-augmented language models.
\newblock In \emph{Advances in Neural Information Processing Systems (NeurIPS)
  Datasets and Benchmarks Track}, 2023.

\bibitem[Hu et~al.(2024)Hu, Zhu, and Welleck]{hu2024minictx}
Jiewen Hu, Thomas Zhu, and Sean Welleck.
\newblock {miniCTX}: Neural theorem proving with (long-)contexts.
\newblock \emph{arXiv preprint}, 2024.

\bibitem[Gallier(2015)]{gallier2015logic}
Jean Gallier.
\newblock \emph{Logic for Computer Science: Foundations of Automatic Theorem
  Proving}.
\newblock Dover Publications, second edition, 2015.
\newblock A corrected version of the original Wiley edition (pp. 511, 1986).

\bibitem[{Anthropic}(2025)]{claudeopus4}
{Anthropic}.
\newblock {C}laude opus 4 system card.
\newblock Technical report, Anthropic, 2025.
\newblock URL \url{https://www.anthropic.com/claude-opus-4-system-card}.

\bibitem[{OpenAI}(2025{\natexlab{a}})]{gpt5}
{OpenAI}.
\newblock {GPT-5} system card.
\newblock Technical report, OpenAI, 2025{\natexlab{a}}.
\newblock URL \url{https://openai.com/research/gpt-5-system-card}.

\bibitem[Paruchuri and {Datalab}(2024)]{marker2024}
Vikas Paruchuri and {Datalab}.
\newblock Marker: Convert pdf to markdown, 2024.
\newblock URL \url{https://github.com/datalab-to/marker}.

\bibitem[{Google DeepMind}(2026)]{gemini3}
{Google DeepMind}.
\newblock Gemini 3.1 pro -- model card.
\newblock \url{https://deepmind.google/models/model-cards/gemini-3-1-pro/}, feb
  2026.

\bibitem[{DeepSeek-AI} et~al.(2025){DeepSeek-AI}, Guo, Yang, Zhang, Song, Wang,
  Zhu, Xu, Zhang, Ma, Bi, Zhang, Yu, Wu, et~al.]{deepseekr1}
{DeepSeek-AI}, Daya Guo, Dejian Yang, Haowei Zhang, Junxiao Song, Peiyi Wang,
  Qihao Zhu, Runxin Xu, Ruoyu Zhang, Shirong Ma, Xiao Bi, Xiaokang Zhang,
  Xingkai Yu, Yu~Wu, et~al.
\newblock {DeepSeek-R1}: Incentivizing reasoning capability in {LLMs} via
  reinforcement learning, 2025.

\bibitem[{DeepSeek-AI} et~al.(2024){DeepSeek-AI}, Liu, Feng, Xue, Wang, Wu, Lu,
  Zhao, Deng, Zhang, Ruan, Dai, Guo, Yang, Chen, et~al.]{deepseekv3}
{DeepSeek-AI}, Aixin Liu, Bei Feng, Bing Xue, Bingxuan Wang, Bochao Wu, Chengda
  Lu, Chenggang Zhao, Chengqi Deng, Chenyu Zhang, Chong Ruan, Damai Dai, Daya
  Guo, Dejian Yang, Deli Chen, et~al.
\newblock {DeepSeek-V3} technical report, 2024.

\bibitem[{Kimi Team} et~al.(2025){Kimi Team}, Bai, Bao, Chen, Chen,
  et~al.]{kimik2}
{Kimi Team}, Yifan Bai, Yiping Bao, Cheng Chen, Guanduo Chen, et~al.
\newblock {Kimi K2}: Open agentic intelligence, 2025.

\bibitem[Rastogi et~al.(2025)Rastogi, Yang, Jiang, Liu, Sablayrolles,
  et~al.]{devstral}
Abhinav Rastogi, Adam Yang, Albert~Q. Jiang, Alexander~H. Liu, Alexandre
  Sablayrolles, et~al.
\newblock Devstral: Fine-tuning language models for coding agent applications,
  2025.

\bibitem[{Meta AI}(2025)]{llama4}
{Meta AI}.
\newblock The {Llama 4} herd: The beginning of a new era of natively multimodal
  {AI} innovation.
\newblock \url{https://ai.meta.com/blog/llama-4-multimodal-intelligence/},
  2025.
\newblock Blog post, April 2025.

\bibitem[Yang et~al.(2025{\natexlab{a}})Yang, Li, Yang, Zhang, Hui, Zheng, Yu,
  Gao, Huang, et~al.]{qwen3}
An~Yang, Anfeng Li, Baosong Yang, Beichen Zhang, Binyuan Hui, Bo~Zheng, Bowen
  Yu, Chang Gao, Chengen Huang, et~al.
\newblock {Qwen3} technical report, 2025{\natexlab{a}}.

\bibitem[Hui et~al.(2024)Hui, Yang, Cui, Yang, Liu, Zhang, Liu, Zhang, Yu, Lu,
  Dang, Fan, Zhang, Yang, Men, Huang, Zheng, Miao, Quan, Feng, Ren, Ren, Zhou,
  and Lin]{qwen25coder}
Binyuan Hui, Jian Yang, Zeyu Cui, Jiaxi Yang, Dayiheng Liu, Lei Zhang, Tianyu
  Liu, Jiajun Zhang, Bowen Yu, Keming Lu, Kai Dang, Yang Fan, Yichang Zhang,
  An~Yang, Rui Men, Fei Huang, Bo~Zheng, Yibo Miao, Shanghaoran Quan, Yunlong
  Feng, Xingzhang Ren, Xuancheng Ren, Jingren Zhou, and Junyang Lin.
\newblock {Qwen2.5-Coder} technical report, 2024.

\bibitem[{MiniMax} et~al.(2025){MiniMax}, Chen, Li, Gong, et~al.]{minimaxm1}
{MiniMax}, Aili Chen, Aonian Li, Bangwei Gong, et~al.
\newblock {MiniMax-M1}: Scaling test-time compute efficiently with lightning
  attention, 2025.

\bibitem[{Team GLM} et~al.(2024){Team GLM}, Zeng, Xu, Wang, Zhang, Yin, Zhang,
  Rojas, Feng, Zhao, Lai, Yu, Wang, Sun, Zhang, Cheng, Gui, Tang, Zhang, Sun,
  Li, Zhao, Wu, Zhong, Liu, et~al.]{chatglm4}
{Team GLM}, Aohan Zeng, Bin Xu, Bowen Wang, Chenhui Zhang, Da~Yin, Dan Zhang,
  Diego Rojas, Guanyu Feng, Hanlin Zhao, Hanyu Lai, Hao Yu, Hongning Wang,
  Jiadai Sun, Jiajie Zhang, Jiale Cheng, Jiayi Gui, Jie Tang, Jing Zhang,
  Jingyu Sun, Juanzi Li, Lei Zhao, Lindong Wu, Lucen Zhong, Mingdao Liu, et~al.
\newblock {ChatGLM}: A family of large language models from {GLM-130B} to
  {GLM-4} all tools, 2024.

\bibitem[{OpenAI}(2025{\natexlab{b}})]{gptoss}
{OpenAI}.
\newblock gpt-oss-120b \& gpt-oss-20b model card, 2025{\natexlab{b}}.

\bibitem[Yang et~al.(2024{\natexlab{b}})Yang, Jimenez, Wettig, Lieret, Yao,
  Narasimhan, and Press]{yang2024swe}
John Yang, Carlos~E. Jimenez, Alexander Wettig, Kilian Lieret, Shunyu Yao,
  Karthik Narasimhan, and Ofir Press.
\newblock {SWE-agent}: Agent-computer interfaces enable automated software
  engineering.
\newblock In \emph{Advances in Neural Information Processing Systems
  (NeurIPS)}, 2024{\natexlab{b}}.

\bibitem[Yang et~al.(2025{\natexlab{b}})Yang, Zhang, Cao, Zhou, Li, Guo, Yao,
  Chen, Li, and Ma]{yang2025formalml}
Xiao-Wen Yang, Zihao Zhang, Jianuo Cao, Zhi Zhou, Zenan Li, Lan-Zhe Guo, Yuan
  Yao, Taolue Chen, Yu-Feng Li, and Xiaoxing Ma.
\newblock Formalml: A benchmark for evaluating formal subgoal completion in
  machine learning theory, 2025{\natexlab{b}}.
\newblock URL \url{https://arxiv.org/abs/2510.02335}.

\bibitem[Liu et~al.(2025{\natexlab{b}})Liu, Bao, Zhang, Liu, Chen, Liu, Jiao,
  and Luo]{liu2025atlas}
Xiaoyang Liu, Kangjie Bao, Jiashuo Zhang, Yunqi Liu, Yu~Chen, Yuntian Liu, Yang
  Jiao, and Tao Luo.
\newblock {ATLAS}: Autoformalizing theorems through lifting, augmentation, and
  synthesis of data.
\newblock \emph{arXiv preprint}, 2025{\natexlab{b}}.

\bibitem[Wang et~al.(2025{\natexlab{a}})Wang, Xie, Wang, Gao, Yu, and
  Dong]{wang2025aria}
Hanyu Wang, Ruohan Xie, Yutong Wang, Guoxiong Gao, Xintao Yu, and Bin Dong.
\newblock {Aria}: An agent for retrieval and iterative auto-formalization via
  dependency graph.
\newblock \emph{arXiv preprint}, 2025{\natexlab{a}}.

\bibitem[Lu et~al.(2024{\natexlab{a}})Lu, Wan, Liu, Huang, Xiong, Liu, Shen,
  Jin, Zhang, Wang, Yang, Tang, and Guo]{lu2024process}
Jianqiao Lu, Yingjia Wan, Zhengying Liu, Yinya Huang, Jing Xiong, Chengwu Liu,
  Jianhao Shen, Hui Jin, Jipeng Zhang, Haiming Wang, Zhicheng Yang, Jing Tang,
  and Zhijiang Guo.
\newblock Process-driven autoformalization in {L}ean 4.
\newblock \emph{arXiv preprint}, 2024{\natexlab{a}}.

\bibitem[Zhang et~al.(2025{\natexlab{a}})Zhang, Borchert, Gritta, and
  Lampouras]{zhang2025drift}
Meiru Zhang, Philipp Borchert, Milan Gritta, and Gerasimos Lampouras.
\newblock {DRIFT}: Decompose, retrieve, illustrate, then formalize theorems.
\newblock \emph{arXiv preprint}, 2025{\natexlab{a}}.

\bibitem[Guo et~al.(2025)Guo, Wang, Zhang, Kong, Huang, Xi, Wang, Wang, Cai,
  Zhang, and Ye]{guo2025autoformalizer}
Qi~Guo, Jianing Wang, Jianfei Zhang, Deyang Kong, Xiangzhou Huang, Xiangyu Xi,
  Wei Wang, Jingang Wang, Xunliang Cai, Shikun Zhang, and Wei Ye.
\newblock Autoformalizer with tool feedback.
\newblock \emph{arXiv preprint}, 2025.

\bibitem[Zhang et~al.(2024{\natexlab{a}})Zhang, Quan, and
  Freitas]{Zhang2024consistent}
Lan Zhang, Xin Quan, and Andre Freitas.
\newblock Consistent autoformalization for constructing mathematical libraries.
\newblock In \emph{Proceedings of the 2024 Conference on Empirical Methods in
  Natural Language Processing}, page 4020–4033. Association for Computational
  Linguistics, 2024{\natexlab{a}}.
\newblock \doi{10.18653/v1/2024.emnlp-main.233}.
\newblock URL \url{http://dx.doi.org/10.18653/v1/2024.emnlp-main.233}.

\bibitem[Min et~al.(2026{\natexlab{a}})Min, Gao, Sy, Li, Si, and
  Bastani]{min2026divide}
Marcus~J. Min, Yeqi Gao, Wilson Sy, Zhaoyu Li, Xujie Si, and Osbert Bastani.
\newblock Divide and abstract: Autoformalization via decomposition and
  abstraction learning.
\newblock In \emph{The Fourteenth International Conference on Learning
  Representations}, 2026{\natexlab{a}}.
\newblock URL \url{https://openreview.net/forum?id=NjgaeXNit3}.

\bibitem[Min et~al.(2026{\natexlab{b}})Min, He, Li, Yi, Malik, Gupta, Si, and
  Bastani]{min2026theorylevel}
Marcus~J. Min, Mike He, Zhaoyu Li, Zixuan Yi, Sharad Malik, Aarti Gupta, Xujie
  Si, and Osbert Bastani.
\newblock Position: Theory-level autoformalization, from isolated statements to
  unified formal knowledge bases.
\newblock In \emph{Proceedings of the 43rd International Conference on Machine
  Learning}, volume 306 of \emph{Proceedings of Machine Learning Research},
  pages 171486--171501. PMLR, 2026{\natexlab{b}}.
\newblock URL \url{https://proceedings.mlr.press/v306/min26h.html}.

\bibitem[Wang et~al.(2026)Wang, Ma, Ming, Zhang, Yuan, and Wen]{wang2026m2f}
Zichen Wang, Wanli Ma, Zhenyu Ming, Gong Zhang, Kun Yuan, and Zaiwen Wen.
\newblock {M2F}: Automated formalization of mathematical literature at scale.
\newblock \emph{arXiv preprint arXiv:2602.17016}, 2026.

\bibitem[Rammal et~al.(2026)Rammal, Patel, Gloeckle, Hayat, Kempe, Munos,
  Arnal, and Cabannes]{rammal2026formalizing}
Ahmad Rammal, Niket Patel, Fabian Gloeckle, Amaury Hayat, Julia Kempe, Remi
  Munos, Charles Arnal, and Vivien Cabannes.
\newblock Formalizing mathematics at scale.
\newblock \emph{arXiv preprint arXiv:2605.29955}, 2026.

\bibitem[Zhang et~al.(2026)Zhang, Sun, Suzuki, Lee, and
  Liu]{zhang2026leanmarathon}
Yuanhe Zhang, Yuekai Sun, Taiji Suzuki, Jason~D. Lee, and Fanghui Liu.
\newblock {LeanMarathon}: Toward reliable {AI} co-mathematicians through
  long-horizon {Lean} autoformalization.
\newblock \emph{arXiv preprint arXiv:2606.05400}, 2026.

\bibitem[{Math, Inc.}(2025)]{mathinc2025gauss}
{Math, Inc.}
\newblock Introducing {Gauss}, an agent for autoformalization.
\newblock \url{https://www.math.inc/gauss}, 2025.

\bibitem[Leroy(2009)]{leroy2009formal}
Xavier Leroy.
\newblock Formal verification of a realistic compiler.
\newblock \emph{Communications of the ACM}, 52\penalty0 (7):\penalty0 107--115,
  2009.

\bibitem[Zhao et~al.(2013)Zhao, Nagarakatte, Martin, and
  Zdancewic]{zhao2013formal}
Jianzhou Zhao, Santosh Nagarakatte, Milo M.~K. Martin, and Steve Zdancewic.
\newblock Formal verification of {SSA}-based optimizations for {LLVM}.
\newblock In \emph{Proceedings of the ACM SIGPLAN Conference on Programming
  Language Design and Implementation (PLDI)}, 2013.

\bibitem[Klein et~al.(2009)Klein, Elphinstone, Heiser, Andronick, Cock, Derrin,
  Elkaduwe, Engelhardt, Kolanski, Norrish, Sewell, Tuch, and
  Winwood]{klein2009sel4}
Gerwin Klein, Kevin Elphinstone, Gernot Heiser, June Andronick, David Cock,
  Philip Derrin, Dhammika Elkaduwe, Kai Engelhardt, Rafal Kolanski, Michael
  Norrish, Thomas Sewell, Harvey Tuch, and Simon Winwood.
\newblock {seL4}: formal verification of an {OS} kernel.
\newblock In \emph{Proceedings of the ACM SIGOPS Symposium on Operating Systems
  Principles (SOSP)}, pages 207--220, 2009.

\bibitem[Gu et~al.(2016)Gu, Shao, Chen, Wu, Kim, Sj{\"o}berg, and
  Costanzo]{gu2016certikos}
Ronghui Gu, Zhong Shao, Hao Chen, Xiongnan~(Newman) Wu, Jieung Kim, Vilhelm
  Sj{\"o}berg, and David Costanzo.
\newblock {CertiKOS}: An extensible architecture for building certified
  concurrent {OS} kernels.
\newblock In \emph{USENIX Symposium on Operating Systems Design and
  Implementation (OSDI)}, 2016.

\bibitem[Hawblitzel et~al.(2015)Hawblitzel, Howell, Kapritsos, Lorch, Parno,
  Roberts, Setty, and Zill]{hawblitzel2015ironfleet}
Chris Hawblitzel, Jon Howell, Manos Kapritsos, Jacob~R. Lorch, Bryan Parno,
  Michael~L. Roberts, Srinath Setty, and Brian Zill.
\newblock {IronFleet}: Proving practical distributed systems correct.
\newblock In \emph{Proceedings of the ACM Symposium on Operating Systems
  Principles (SOSP)}, 2015.

\bibitem[Wilcox et~al.(2015)Wilcox, Woos, Panchekha, Tatlock, Wang, Ernst, and
  Anderson]{wilcox2015verdi}
James~R. Wilcox, Doug Woos, Pavel Panchekha, Zachary Tatlock, Xi~Wang,
  Michael~D. Ernst, and Thomas Anderson.
\newblock {Verdi}: A framework for implementing and formally verifying
  distributed systems.
\newblock In \emph{Proceedings of the ACM SIGPLAN Conference on Programming
  Language Design and Implementation (PLDI)}, 2015.

\bibitem[Appel et~al.(2017)Appel, Beringer, Chlipala, Morrisett, Pierce,
  et~al.]{appel2017deepspec}
Andrew~W. Appel, Lennart Beringer, Adam Chlipala, Greg Morrisett, Benjamin~C.
  Pierce, et~al.
\newblock The {D}eep{S}pec project: {T}he science of deep specification.
\newblock NSF Expedition in Computing, 2017.
\newblock \url{https://deepspec.org/}.

\bibitem[Appel(2014)]{appel2014program}
Andrew~W. Appel.
\newblock \emph{Program Logics for Certified Compilers}.
\newblock Cambridge University Press, 2014.

\bibitem[Pierce et~al.(2010)Pierce, Appel, Chlipala, Weirich, Casinghino,
  de~Amorim, Beringer, Yorgey, et~al.]{pierce2010software}
Benjamin~C. Pierce, Andrew~W. Appel, Adam Chlipala, Stephanie Weirich, Chris
  Casinghino, Arthur~Azevedo de~Amorim, Lennart Beringer, Brent Yorgey, et~al.
\newblock Software foundations.
\newblock Electronic textbook series, 2010.
\newblock \url{https://softwarefoundations.cis.upenn.edu/}.

\bibitem[Nipkow and Klein(2014)]{nipkow2014concrete}
Tobias Nipkow and Gerwin Klein.
\newblock \emph{Concrete Semantics with {I}sabelle/{HOL}}.
\newblock Springer, 2014.

\bibitem[Chlipala(2013)]{chlipala2013certified}
Adam Chlipala.
\newblock \emph{Certified Programming with Dependent Types}.
\newblock MIT Press, 2013.

\bibitem[From et~al.(2022)From, Jacobsen, and Villadsen]{from2022secav}
Asta~Halkj{\ae}r From, Frederik~Krogsdal Jacobsen, and J{\o}rgen Villadsen.
\newblock {SeCaV}: A sequent calculus verifier in {I}sabelle/{HOL}.
\newblock \emph{arXiv preprint}, 2022.

\bibitem[From et~al.(2020)From, Jensen, Schlichtkrull, and
  Villadsen]{from2020teaching}
Asta~Halkj{\ae}r From, Alexander~Birch Jensen, Anders Schlichtkrull, and
  J{\o}rgen Villadsen.
\newblock Teaching a formalized logical calculus.
\newblock \emph{arXiv preprint}, 2020.

\bibitem[Herbelin and Ilik(2024)]{herbelin2024analysis}
Hugo Herbelin and Danko Ilik.
\newblock An analysis of the constructive content of {H}enkin's proof of
  {G}{\"o}del's completeness theorem.
\newblock \emph{arXiv preprint}, 2024.

\bibitem[Aydemir et~al.(2005)Aydemir, Bohannon, Fairbairn, Foster, Pierce,
  Sewell, Vytiniotis, Washburn, Weirich, and Zdancewic]{aydemir2005mechanized}
Brian~E. Aydemir, Aaron Bohannon, Matthew Fairbairn, J.~Nathan Foster,
  Benjamin~C. Pierce, Peter Sewell, Dimitrios Vytiniotis, Geoffrey Washburn,
  Stephanie Weirich, and Steve Zdancewic.
\newblock Mechanized metatheory for the masses: The {POPL}mark challenge.
\newblock In \emph{International Conference on Theorem Proving in Higher Order
  Logics (TPHOLs)}, 2005.

\bibitem[Li et~al.(2023)Li, Huang, and Naik]{li2023scallop}
Ziyang Li, Jiani Huang, and Mayur Naik.
\newblock {S}callop: A language for neurosymbolic programming.
\newblock \emph{Proceedings of the ACM on Programming Languages (PLDI)}, 7,
  2023.
\newblock \doi{10.1145/3591280}.

\bibitem[Huang et~al.(2021)Huang, Li, Chen, Samel, Naik, Song, and
  Si]{huang2021scallop}
Jiani Huang, Ziyang Li, Binghong Chen, Karan Samel, Mayur Naik, Le~Song, and
  Xujie Si.
\newblock {S}callop: From probabilistic deductive databases to scalable
  differentiable reasoning.
\newblock In \emph{Advances in Neural Information Processing Systems
  (NeurIPS)}, 2021.

\bibitem[Li et~al.(2024{\natexlab{b}})Li, Huang, Liu, and
  Naik]{li2024neurosymbolic}
Ziyang Li, Jiani Huang, Jason Liu, and Mayur Naik.
\newblock Neurosymbolic programming in {S}callop: Principles and practice.
\newblock \emph{Foundations and Trends in Programming Languages},
  2024{\natexlab{b}}.
\newblock \doi{10.1561/2500000059}.

\bibitem[Manhaeve et~al.(2018)Manhaeve, Dumancic, Kimmig, Demeester, and
  De~Raedt]{manhaeve2018deepproblog}
Robin Manhaeve, Sebastijan Dumancic, Angelika Kimmig, Thomas Demeester, and Luc
  De~Raedt.
\newblock {D}eep{P}rob{L}og: Neural probabilistic logic programming.
\newblock In \emph{Advances in Neural Information Processing Systems
  (NeurIPS)}, 2018.

\bibitem[Yang et~al.(2020)Yang, Ishay, and Lee]{yang2020neurasp}
Zhun Yang, Adam Ishay, and Joohyung Lee.
\newblock {N}eur{ASP}: Embracing neural networks into answer set programming.
\newblock In \emph{International Joint Conference on Artificial Intelligence
  (IJCAI)}, 2020.

\bibitem[Zhang et~al.(2024{\natexlab{b}})Zhang, Zhang, Hou, Wang, Gu, Clark,
  Callison-Burch, and Tandon]{zhang2024proc2pddl}
Tianyi Zhang, Li~Zhang, Zhaoyi Hou, Ziyu Wang, Yuling Gu, Peter Clark, Chris
  Callison-Burch, and Niket Tandon.
\newblock {PROC2PDDL}: Open-domain planning representations from texts.
\newblock In \emph{Workshop on Natural Language Reasoning and Structured
  Explanations (NLRSE) at NAACL}, 2024{\natexlab{b}}.

\bibitem[Huang and Zhang(2025)]{huang2025limit}
Cassie Huang and Li~Zhang.
\newblock On the limit of language models as planning formalizers.
\newblock In \emph{Annual Meeting of the Association for Computational
  Linguistics (ACL)}, 2025.

\bibitem[Bingham et~al.(2019)Bingham, Chen, Jankowiak, Obermeyer, Pradhan,
  Karaletsos, Singh, Szerlip, Horsfall, and Goodman]{bingham2019pyro}
Eli Bingham, Jonathan~P. Chen, Martin Jankowiak, Fritz Obermeyer, Neeraj
  Pradhan, Theofanis Karaletsos, Rohit Singh, Paul Szerlip, Paul Horsfall, and
  Noah~D. Goodman.
\newblock {P}yro: Deep universal probabilistic programming.
\newblock \emph{Journal of Machine Learning Research}, 2019.

\bibitem[Carpenter et~al.(2017)Carpenter, Gelman, Hoffman, Lee, Goodrich,
  Betancourt, Brubaker, Guo, Li, and Riddell]{carpenter2017stan}
Bob Carpenter, Andrew Gelman, Matthew~D. Hoffman, Daniel Lee, Ben Goodrich,
  Michael Betancourt, Marcus~A. Brubaker, Jiqiang Guo, Peter Li, and Allen
  Riddell.
\newblock {S}tan: A probabilistic programming language.
\newblock \emph{Journal of Statistical Software}, 76\penalty0 (1), 2017.

\bibitem[De~Raedt et~al.(2007)De~Raedt, Kimmig, and
  Toivonen]{deraedt2007problog}
Luc De~Raedt, Angelika Kimmig, and Hannu Toivonen.
\newblock {P}rob{L}og: A probabilistic {P}rolog and its application in link
  discovery.
\newblock In \emph{International Joint Conference on Artificial Intelligence
  (IJCAI)}, 2007.

\bibitem[Jimenez et~al.(2024)Jimenez, Yang, Wettig, Yao, Pei, Press, and
  Narasimhan]{jimenez2024swe}
Carlos~E Jimenez, John Yang, Alexander Wettig, Shunyu Yao, Kexin Pei, Ofir
  Press, and Karthik~R Narasimhan.
\newblock {SWE}-bench: Can language models resolve real-world github issues?
\newblock In \emph{The Twelfth International Conference on Learning
  Representations}, 2024.
\newblock URL \url{https://openreview.net/forum?id=VTF8yNQM66}.

\bibitem[Wang et~al.(2025{\natexlab{b}})Wang, Li, Song, Xu, Tang, Zhuge, Pan,
  Song, Li, Singh, Tran, Li, Ma, Zheng, Qian, Shao, Muennighoff, Zhang, Hui,
  Lin, Brennan, Peng, Ji, and Neubig]{wang2025openhands}
Xingyao Wang, Boxuan Li, Yufan Song, Frank~F. Xu, Xiangru Tang, Mingchen Zhuge,
  Jiayi Pan, Yueqi Song, Bowen Li, Jaskirat Singh, Hoang~H. Tran, Fuqiang Li,
  Ren Ma, Mingzhang Zheng, Bill Qian, Yanjun Shao, Niklas Muennighoff, Yizhe
  Zhang, Binyuan Hui, Junyang Lin, Robert Brennan, Hao Peng, Heng Ji, and
  Graham Neubig.
\newblock {OpenHands}: An open platform for {AI} software developers as
  generalist agents.
\newblock In \emph{International Conference on Learning Representations
  (ICLR)}, 2025{\natexlab{b}}.

\bibitem[Xin et~al.(2024{\natexlab{a}})Xin, Guo, Shao, Ren, Zhu, Liu, Ruan, Li,
  and Liang]{xin2024deepseek}
Huajian Xin, Daya Guo, Zhihong Shao, Zhizhou Ren, Qihao Zhu, Bo~Liu, Chong
  Ruan, Wenda Li, and Xiaodan Liang.
\newblock {DeepSeek-Prover}: Advancing theorem proving in {LLMs} through
  large-scale synthetic data.
\newblock \emph{arXiv preprint}, 2024{\natexlab{a}}.

\bibitem[Xin et~al.(2024{\natexlab{b}})Xin, Ren, Song, Shao, Zhao, Wang, Liu,
  Zhang, Lu, Du, Gao, Zhu, Yang, Gou, Wu, Luo, and Ruan]{xin2024deepseekv15}
Huajian Xin, Z.Z. Ren, Junxiao Song, Zhihong Shao, Wanjia Zhao, Haocheng Wang,
  Bo~Liu, Liyue Zhang, Xuan Lu, Qiushi Du, Wenjun Gao, Qihao Zhu, Dejian Yang,
  Zhibin Gou, Z.F. Wu, Fuli Luo, and Chong Ruan.
\newblock {DeepSeek-Prover-V1.5}: Harnessing proof assistant feedback for
  reinforcement learning and monte-carlo tree search.
\newblock \emph{arXiv preprint}, 2024{\natexlab{b}}.

\bibitem[Ren et~al.(2025)Ren, Shao, Song, Xin, Wang, Zhao, Zhang, Fu, Zhu,
  Yang, Wu, Gou, Ma, Tang, Liu, Gao, Guo, and Ruan]{ren2025deepseek}
Z.Z. Ren, Zhihong Shao, Junxiao Song, Huajian Xin, Haocheng Wang, Wanjia Zhao,
  Liyue Zhang, Zhe Fu, Qihao Zhu, Dejian Yang, Z.F. Wu, Zhibin Gou, Shirong Ma,
  Hongxuan Tang, Yuxuan Liu, Wenjun Gao, Daya Guo, and Chong Ruan.
\newblock {DeepSeek-Prover-V2}: Advancing formal mathematical reasoning via
  reinforcement learning for subgoal decomposition.
\newblock \emph{arXiv preprint}, 2025.

\bibitem[Lin et~al.(2025)Lin, Tang, Lyu, Wu, Lin, Yang, Li, Xia, Chen, Arora,
  and Jin]{lin2025goedel}
Yong Lin, Shange Tang, Bohan Lyu, Jiayun Wu, Hongzhou Lin, Kaiyu Yang, Jia Li,
  Mengzhou Xia, Danqi Chen, Sanjeev Arora, and Chi Jin.
\newblock {Goedel-Prover}: A frontier model for open-source automated theorem
  proving.
\newblock \emph{arXiv preprint}, 2025.

\bibitem[Wang et~al.(2025{\natexlab{c}})Wang, Unsal, Lin, Baksys, Liu, Santos,
  Sung, Vinyes, Ying, Zhu, Lu, de~Saxc{\'e}, Bailey, Song, Xiao, Zhang, Zhang,
  Pu, Zhu, et~al.]{wang2025kimina}
Haiming Wang, Mert Unsal, Xiaohan Lin, Mantas Baksys, Junqi Liu, Marco~Dos
  Santos, Flood Sung, Marina Vinyes, Zhenzhe Ying, Zekai Zhu, Jianqiao Lu,
  Hugues de~Saxc{\'e}, Bolton Bailey, Chendong Song, Chenjun Xiao, Dehao Zhang,
  Ebony Zhang, Frederick Pu, Han Zhu, et~al.
\newblock {Kimina-Prover} preview: Towards large formal reasoning models with
  reinforcement learning.
\newblock \emph{arXiv preprint}, 2025{\natexlab{c}}.

\bibitem[Jiang et~al.(2023)Jiang, Welleck, Zhou, Li, Liu, Jamnik, Lacroix, Wu,
  and Lample]{jiang2023draft}
Albert~Q. Jiang, Sean Welleck, Jin~Peng Zhou, Wenda Li, Jiacheng Liu, Mateja
  Jamnik, Timoth{\'e}e Lacroix, Yuhuai Wu, and Guillaume Lample.
\newblock Draft, sketch, and prove: Guiding formal theorem provers with
  informal proofs.
\newblock In \emph{International Conference on Learning Representations
  (ICLR)}, 2023.

\bibitem[Wang et~al.(2023)Wang, Xin, Zheng, Li, Liu, Cao, Huang, Xiong, Shi,
  Xie, Yin, Li, Liao, and Liang]{wang2023lego}
Haiming Wang, Huajian Xin, Chuanyang Zheng, Lin Li, Zhengying Liu, Qingxing
  Cao, Yinya Huang, Jing Xiong, Han Shi, Enze Xie, Jian Yin, Zhenguo Li, Heng
  Liao, and Xiaodan Liang.
\newblock {LEGO-Prover}: Neural theorem proving with growing libraries.
\newblock \emph{arXiv preprint}, 2023.

\bibitem[Thakur et~al.(2023)Thakur, Tsoukalas, Wen, Xin, and
  Chaudhuri]{thakur2023copra}
Amitayush Thakur, George Tsoukalas, Yeming Wen, Jimmy Xin, and Swarat
  Chaudhuri.
\newblock An in-context learning agent for formal theorem-proving.
\newblock \emph{arXiv preprint}, 2023.

\bibitem[Lample et~al.(2022)Lample, Lachaux, Lavril, Martinet, Hayat, Ebner,
  Rodriguez, and Lacroix]{lample2022hypertree}
Guillaume Lample, Marie-Anne Lachaux, Thibaut Lavril, Xavier Martinet, Amaury
  Hayat, Gabriel Ebner, Aur{\'e}lien Rodriguez, and Timoth{\'e}e Lacroix.
\newblock {HyperTree} proof search for neural theorem proving.
\newblock \emph{arXiv preprint}, 2022.

\bibitem[Poiroux et~al.(2025)Poiroux, Weiss, Kun{\v{c}}ak, and
  Bosselut]{poiroux2025reliable}
Auguste Poiroux, Gail Weiss, Viktor Kun{\v{c}}ak, and Antoine Bosselut.
\newblock Reliable evaluation and benchmarks for statement autoformalization.
\newblock In \emph{Conference on Empirical Methods in Natural Language
  Processing (EMNLP)}, 2025.

\bibitem[Li et~al.(2024{\natexlab{c}})Li, Wu, Li, Wei, Zhang, Yang, and
  Ma]{li2024autoformalize}
Zenan Li, Yifan Wu, Zhaoyu Li, Xinming Wei, Xian Zhang, Fan Yang, and Xiaoxing
  Ma.
\newblock Autoformalize mathematical statements by symbolic equivalence and
  semantic consistency.
\newblock In \emph{Advances in Neural Information Processing Systems
  (NeurIPS)}, 2024{\natexlab{c}}.

\bibitem[Zhang et~al.(2025{\natexlab{b}})Zhang, Valentino, and
  Freitas]{zhang2025autoformalization}
Lan Zhang, Marco Valentino, and Andre Freitas.
\newblock Autoformalization in the wild: Assessing {LLMs} on real-world
  mathematical definitions.
\newblock \emph{arXiv preprint}, 2025{\natexlab{b}}.

\bibitem[Moore and Shah(2025)]{moore2025evaluating}
Hayden Moore and Asfahan Shah.
\newblock Evaluating autoformalization robustness via semantically similar
  paraphrasing.
\newblock \emph{arXiv preprint}, 2025.

\bibitem[Lu et~al.(2024{\natexlab{b}})Lu, Wan, Huang, Xiong, Liu, and
  Guo]{lu2024formalalign}
Jianqiao Lu, Yingjia Wan, Yinya Huang, Jing Xiong, Zhengying Liu, and Zhijiang
  Guo.
\newblock {FormalAlign}: Automated alignment evaluation for autoformalization.
\newblock \emph{arXiv preprint}, 2024{\natexlab{b}}.

\bibitem[Meek et~al.(2026)Meek, Ge, Xiang, Chess, and Ilin]{meek2026numerical}
Theodore Meek, Siyuan Ge, Di~Qiu Xiang, Simon Chess, and Vasily Ilin.
\newblock Formalizing numerical analysis: An agent pipeline and quality audit
  beyond kernel acceptance.
\newblock \emph{arXiv preprint arXiv:2606.14000}, 2026.

\bibitem[Ilin and Nugent(2026)]{ilin2026sorries}
Vasily Ilin and Brian Nugent.
\newblock Sorries are not the hard part: An expert-review case study of a
  semi-autonomous formalization.
\newblock \emph{arXiv preprint arXiv:2606.13925}, 2026.

\end{thebibliography}

\appendix

\section{Details of the Automated Formalization Pipeline}
\label{sec:appendix-pipeline}

This appendix gives a detailed account of the four-stage automated pipeline summarized in Section~\ref{sec:methodology}.

\subsection{Pre-Processing}

The textbook PDF is converted to Markdown using Marker~\cite{marker2024}, producing one file per subsection under a strict formatting guide so that every numbered item carries a stable label.
Graphics such as inference rules, proof trees, and directed acyclic graphs are redrawn by hand as ASCII diagrams embedded in the Markdown, giving the downstream agent parseable structured content.
Each textbook item is then wrapped in a typed XML tag: \texttt{<theorem>}, \texttt{<lemma>}, \texttt{<corollary>}, \texttt{<definition>}, \texttt{<example>}, \texttt{<proof>}, and \texttt{<problem>}.
Many untagged statements carry formal weight---scope restrictions, standing assumptions, and informal lemmas stated in prose---and these are also annotated manually.
The result is a set of per-subsection Markdown files in which every item is a discrete, labeled unit with a stable ID of the form \texttt{chap3::definition-3.2.1}.

\subsection{Formalization Planning}

Planning is divided into two sub-stages corresponding to the two intermediate representations.

\noindent\textbf{L1 annotation (concept graph).}
A \emph{formalize-annotator} subagent reads each section and extracts, sentence by sentence, every keyword, sub-clause, and concept.
Each extracted node must be a concrete noun phrase (e.g., ``propositional formula'', ``left parenthesis count'').
Edges encode textbook-level relationships: \textsc{is-a}, \textsc{has-a}, \textsc{relies-on}, and \textsc{defined-by}.
A sentence-coverage validator requires every sentence to contribute at least one node, and every concept node must be realizable from within the textbook itself with no delegation to external libraries.

\noindent\textbf{L2 annotation (formal signature graph).}
A \emph{formalize-specifier} subagent then annotates each L1 node with its Lean realization: \texttt{lean\_kind} ($\in$ \{\texttt{inductive}, \texttt{structure}, \texttt{def}, \texttt{abbrev}, \texttt{notation}, \texttt{theorem}, \texttt{lemma}, \texttt{instance}\}), constructor names, field names, companion definitions, and an \emph{embedding level}:
\emph{deep} (syntax DSLs as inductive types),
\emph{shallow} (semantic properties over deep syntax using Lean's \texttt{Prop} layer), or
\emph{mixed} (meta-theorems connecting the two layers, such as soundness, completeness, and termination results).
The embedding level determines whether a concept must be a first-class inductive type or may live purely in \texttt{Prop}.
A structural-completeness validator checks that every textbook sentence maps to at least one L2 node and that \textsc{has-a} decompositions respect the dependency order of the L1 graph.
This stage is also where implicit standing assumptions must be resolved: for instance, §3.4.2's connective restriction requires a separate \texttt{PropFormNorm} type rather than re-using \texttt{PropForm}.

\subsection{Lean Implementation}

A \emph{formalize-implementer} subagent translates each module in three rounds, mirroring the dependency order of the L2 graph.

\noindent\textbf{Round 1 (L2 $\to$ L3): Structures and inductive types.}
All \texttt{inductive} types and \texttt{structure}s are defined first.
Each declaration receives an \texttt{@[lcsitem id]} attribute encoding the stable textbook item ID, and \texttt{@[lcsmain]} marks the primary authored declaration to distinguish it from companion definitions and elaboration byproducts.
After this round \texttt{lake build} is run and a review script verifies that the declared types match the L2 signature graph.

\noindent\textbf{Round 2 (L3 $\to$ L4a): Definitions, functions, and examples.}
Building on the type scaffolding, the agent defines functions, abbreviations, instances, and examples, covering both computational content (evaluation functions, normal-form transformers, search procedures) and semantic predicates (validity, satisfiability, tautology).
A review script checks that the definitions preserve textbook semantics.

\noindent\textbf{Round 3 (L4a $\to$ L4): Theorem and lemma statements.}
The agent writes theorem and lemma \emph{statements} from the L2 graph, leaving proof bodies as \texttt{sorry} placeholders.
An implementation validator flags vacuous theorem statements, missing inductive cases, identity-function stubs, and dead \texttt{@[lcsitem]} declarations.
The goal of this round is theory-scale coherence, which requires all statements to type-check, all dependencies to compile, and the overall structure to faithfully mirror the textbook, without requiring complete formal proofs.

\subsection{Formal Correctness Review}

The formal correctness review focuses on reviewing the faithfulness and mathematical correctness of the Lean statement by filling the \texttt{sorry} placeholders or compiling a counter-example.
The central design principle is that sorry-count is \emph{not} a quality metric: a sorry-preserved-but-faithful statement is progress; a proved-but-unfaithful statement requires revert.

\noindent\textbf{Proof search loop.}
For each \texttt{sorry}-bearing theorem, an automated proving agent operates in an iterative LSP-driven loop.
At each step, it reads the current proof state from LSP diagnostics and attempts tactics in a fixed priority order: terminal tactics (\texttt{rfl}, \texttt{decide}, \texttt{omega}), followed by \texttt{simp} and \texttt{ring}, with \texttt{grind} and \texttt{aesop} used as fallbacks.
The agent reports \texttt{PROVED} when it successfully produces a proof that compiles without \texttt{sorry}, \texttt{PARTIAL} when it discharges some but not all goals in a larger proof, and \texttt{FALSE} when it finds evidence that the theorem is not provable under the current formalization.
Successful fills are committed atomically in an isolated git worktree to be reviewed.

\noindent\textbf{Counter-example search.}
When proof search stalls (e.g. multiple agents report \texttt{FALSE}), the pipeline attempts to prove the \emph{negation} of the statement: for $\forall x,\ P(x)$, it constructs $x$ with $\neg P(x)$.
A successful counter-example immediately reclassifies the item as incorrect and routes it to faithfulness review to investigate the root cause rather than further proof attempts.

\noindent\textbf{Triage, decomposition, and review gate.}
A high-level orchestrator manages the loop through an issue-tracking system (600+ YAML issue files).
Issues are triaged by severity (faithfulness failures first, then blocking dependencies, then sorry density).
Stuck tasks are decomposed into sub-goals with hard caps ($\leq$3 sorries, $\leq$2 files per sub-issue).
Before any worktree is merged, a mandatory multi-phase \emph{mathematical review} is triggered.
The review compares the current Lean declaration against (1) the original textbook text, (2) the L2 formal signature graph, (3) the initial auto-formalized implementation at a pinned commit, and (4) the current state.
Any spurious divergence at any layer results in a \texttt{REJECT} verdict; if a downstream review later discovers that an upstream \texttt{FAITHFUL} verdict was erroneous, the entire dependent chain is reverted.

\noindent\textbf{Loophole catalog.}
Over the course of the project, 13 classes of faithfulness failure (L1--L13) were identified from real committed-then-reverted work, each with a detection signature and a prescribed fix.
Representative examples include: L1 (existential-over-state-record, replacing a concrete witness with $\exists w$), L4 (trivial existential of a reflexively-true predicate), L8 (path-disconnected Hintikka set), L12 (unwritten invariant: the signature admits $X$, the proof needs $Y \subset X$, and an adversary lives in $X \setminus Y$), and L13 (silent strengthening, the dual of L1--L12, where a hypothesis added to defeat an adversary narrows the textbook claim).
The loophole catalog serves as a mandatory checklist for regression-testing review quality throughout the project, and is continuously expanded as new failure modes are uncovered during formalization and proof repair.

\section{Dataset Details}
\label{sec:appendix-dataset-details}

\subsection{Evaluation Protocol}
\label{sec:appendix-eval-protocol}

\begin{figure}[t]
\centering

\tikzset{
  filebox/.style={
    draw=codebox-frame, fill=codebox-bg,
    rounded corners=3pt,
    font=\ttfamily\scriptsize,
    inner sep=5pt, minimum height=18pt,
    text centered,
  },
  modelbox/.style={
    draw=codebox-frame!80, fill=codebox-frame!20,
    rounded corners=3pt,
    font=\bfseries\scriptsize,
    inner sep=5pt, minimum height=18pt,
    text centered,
  },
  checkerbox/.style={
    draw=codebox-frame!60, fill=codebox-frame!10,
    rounded corners=3pt,
    font=\itshape\scriptsize,
    inner sep=5pt, minimum height=18pt,
    text centered,
    dashed,
  },
  optionalbox/.style={
    draw=codebox-frame, fill=codebox-bg,
    rounded corners=3pt,
    font=\ttfamily\scriptsize,
    inner sep=5pt, minimum height=18pt,
    text centered,
    dashed,
  },
  arr/.style={-{Stealth[length=5pt]}, thick, color=codebox-frame!80},
  checkrarr/.style={-{Stealth[length=5pt]}, thick, color=codebox-frame!50, dashed},
  grouplabel/.style={font=\footnotesize\bfseries, anchor=west},
}

\begin{subfigure}{\linewidth}
  \centering
  \begin{tikzpicture}[node distance=6pt]

  \node[grouplabel] (aflabel) {};

  \node[filebox, below=20pt of aflabel.west, anchor=west] (quote) {Quote.md};
  \node[filebox, below=6pt of quote] (ctx)   {Context.lean};
  \node[filebox, below=6pt of ctx]   (start) {Starter.lean.in};

  \node[modelbox, right=28pt of ctx] (model) {Model};

  \draw[arr] (quote.east)  -- (model.west);
  \draw[arr] (ctx.east)    -- (model.west);
  \draw[arr] (start.east)  -- (model.west);

  \node[filebox, right=28pt of model] (sub) {Submission.lean};
  \draw[arr] (model.east) -- (sub.west);

  \node[checkerbox, align=center, right=20pt of sub] (gt) {GroundTruth.lean \\ {\normalfont\textit{Against DefEq Checker}}};

  \draw[checkrarr] (sub.east)  -- (gt.west);

  \node[font=\scriptsize, right=4pt of gt, text=gray] {\cmark / \xmark};

  \end{tikzpicture}
  \caption{AF tracks: the model receives \texttt{Quote.md}, \texttt{Context.lean}, and \texttt{Starter.lean.in}, and produces \texttt{Submission.lean}, which is checked against \texttt{GroundTruth.lean} by the DefEq Checker.}
\end{subfigure}

\begin{subfigure}{\linewidth}
\centering
\begin{tikzpicture}[node distance=6pt]

\node[grouplabel] (tplabel) {};

\node[optionalbox, below=20pt of tplabel.west, anchor=west] (quote) {Quote.md};
\node[filebox, below=6pt of quote] (ctx)   {Context.lean};
\node[filebox, align=center, below=6pt of ctx] (start)
    {Starter.lean.in \\ {\normalfont\itshape (theorems \texttt{sorry}-stubbed)}};

\node[modelbox, right=28pt of ctx] (model) {Model};

\draw[arr, densely dashed] (quote.east) -- (model.west);
\draw[arr]                 (ctx.east)   -- (model.west);
\draw[arr]                 (start.east) -- (model.west);

\node[filebox, align=center, right=28pt of model] (sub)
    {Submission.lean \\ {\normalfont\itshape (no \texttt{sorry})}};
\draw[arr] (model.east) -- (sub.west);
q
\node[checkerbox, align=center, right=20pt of sub] (chk)
    {Lean type-checker \\ {+ cheat scan}};
\draw[checkrarr] (sub.east) -- (chk.west);
\node[font=\scriptsize, right=4pt of chk, text=gray] {\cmark / \xmark};

\end{tikzpicture}
\caption{TP tracks: the model receives \texttt{Context.lean}, \texttt{Starter.lean.in} (theorem bodies replaced by \texttt{sorry}), and, for ITP only, \texttt{Quote.md}; it produces \texttt{Submission.lean},
accepted iff all target declarations are present, the file builds, no \texttt{sorry} remains, and no cheat patterns (new \texttt{axiom}, \texttt{opaque}, \texttt{@[implemented\_by]}, \ldots) appear.}
\end{subfigure}

\caption{Evaluation protocol diagrams.}
\label{fig:eval-protocol}
\end{figure}
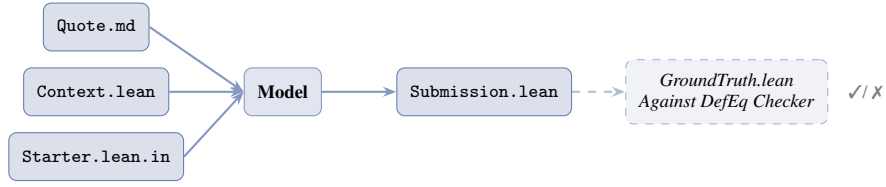
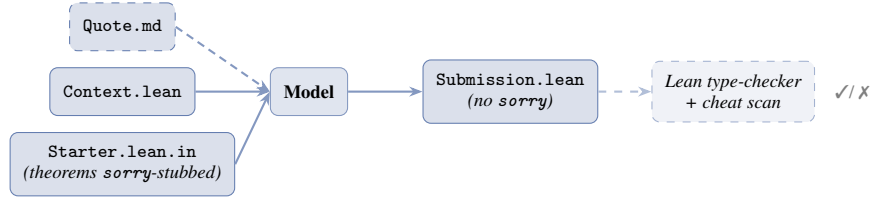

Figure~\ref{fig:eval-protocol} illustrates the evaluation protocol for the AF and TP tracks.

\definecolor{err-pass}  {RGB}{ 95,155, 95}
\definecolor{err-nmexpr}{RGB}{178, 90, 85}
\definecolor{err-cmiss} {RGB}{216,152,148}
\definecolor{err-nmkind}{RGB}{120, 60, 78}
\definecolor{err-comp}  {RGB}{160,160,160}
\definecolor{err-tmo}   {RGB}{220,220,220}

\pgfplotsset{
  iaferrbar/.style={
    xmin=0, xmax=100,
    xtick distance=10,
    xmajorgrids, grid style={dashed, gray!30},
    xlabel={Item share (\%)},
    xlabel style={font=\scriptsize},
    x tick label style={font=\scriptsize},
    y tick label style={font=\scriptsize},
    tick align=outside,
    bar width=8pt,
    legend style={
      at={(0.5,1.02)}, anchor=south,
      draw=black!30, fill=white,
      font=\scriptsize, inner sep=2pt, row sep=-1pt,
      legend columns=-1, column sep=0.6ex,
    },
    legend cell align=left,
    legend image code/.code={
      \draw[#1] (0cm,-0.06cm) rectangle (0.20cm,0.06cm);
    },
  },
}

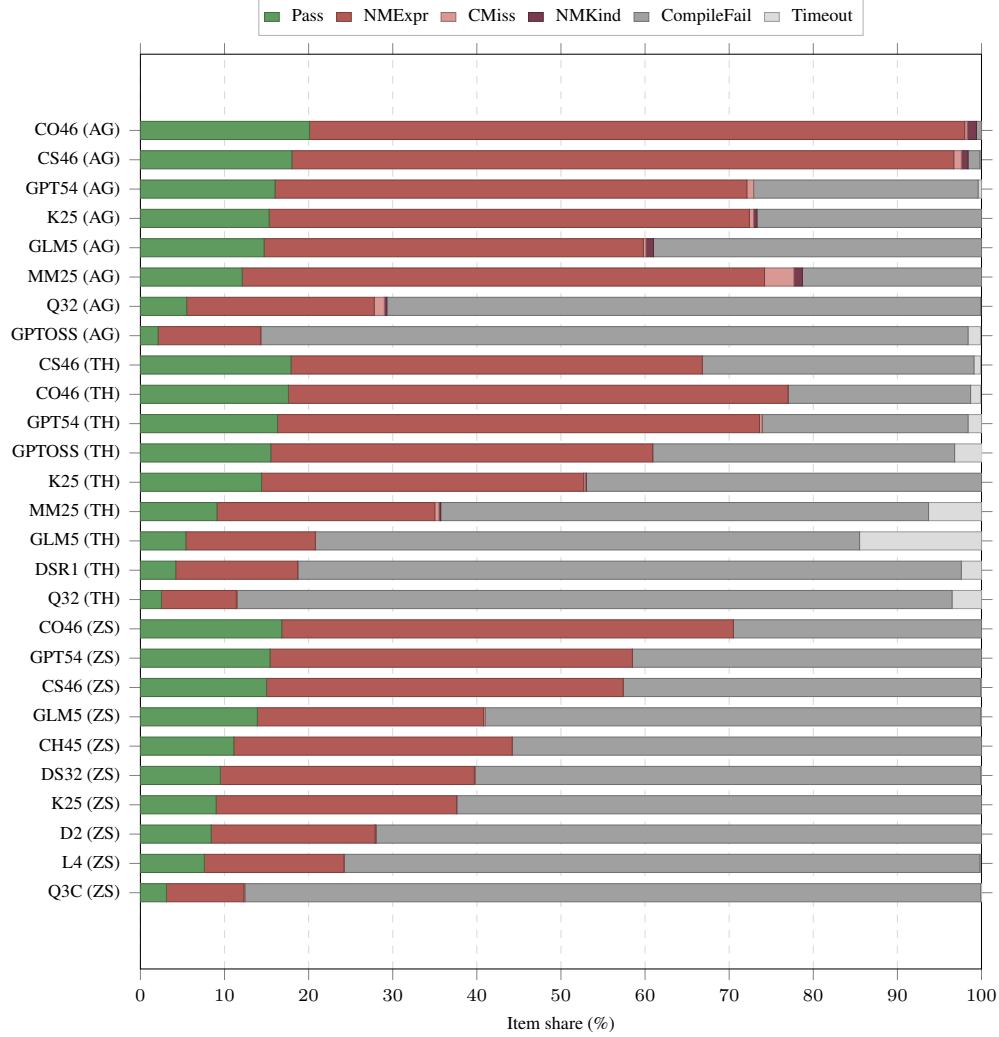
\begin{figure}[t]
\centering
\resizebox{0.95\linewidth}{!}{
\begin{tikzpicture}
\begin{axis}[
  xbar stacked, iaferrbar,
  width=13cm, height=17.6cm,
  bar width=7pt,
  symbolic y coords={
    zs-Q3C, zs-L4, zs-D2, zs-KK25, zs-DS32, zs-GLM52, zs-CH45, zs-GLM5, zs-CS46, zs-GPT54, zs-CO46,
    th-DSR1, th-Q32, th-MM25, th-KK25, th-GLM5, th-GPTOSS, th-KK27, th-GPT54, th-CO46, th-GLM52, th-CS46, th-GP31,
    ag-GPTOSS, ag-Q32, ag-MM25, ag-KK25, ag-GLM5, ag-GPT54, ag-CS46, ag-CO46, ag-GLM52, ag-GP31, ag-KK27
  },
  ytick={zs-Q3C, zs-L4, zs-D2, zs-KK25, zs-DS32, zs-GLM52, zs-CH45, zs-GLM5, zs-CS46, zs-GPT54, zs-CO46,
         th-DSR1, th-Q32, th-MM25, th-KK25, th-GLM5, th-GPTOSS, th-KK27, th-GPT54, th-CO46, th-GLM52, th-CS46, th-GP31,
         ag-GPTOSS, ag-Q32, ag-MM25, ag-KK25, ag-GLM5, ag-GPT54, ag-CS46, ag-CO46, ag-GLM52, ag-GP31, ag-KK27},
  yticklabels={
    {Q3C (ZS)}, {L4 (ZS)}, {D2 (ZS)}, {KK25 (ZS)}, {DS32 (ZS)}, {GLM52 (ZS)}, {CH45 (ZS)}, {GLM5 (ZS)}, {CS46 (ZS)}, {GPT54 (ZS)}, {CO46 (ZS)},
    {DSR1 (TH)}, {Q32 (TH)}, {MM25 (TH)}, {KK25 (TH)}, {GLM5 (TH)}, {GPTOSS (TH)}, {KK27 (TH)}, {GPT54 (TH)}, {CO46 (TH)}, {GLM52 (TH)}, {CS46 (TH)}, {GP31 (TH)},
    {GPTOSS (AG)}, {Q32 (AG)}, {MM25 (AG)}, {KK25 (AG)}, {GLM5 (AG)}, {GPT54 (AG)}, {CS46 (AG)}, {CO46 (AG)}, {GLM52 (AG)}, {GP31 (AG)}, {KK27 (AG)}},
]
  \addplot+[fill=err-pass,   draw=err-pass!70!black,   mark=none]
    coordinates {(3.1,zs-Q3C) (7.6,zs-L4) (8.4,zs-D2) (9.0,zs-KK25) (9.5,zs-DS32) (11.1,zs-CH45) (13.9,zs-GLM5) (15.0,zs-CS46) (15.4,zs-GPT54) (16.8,zs-CO46) (6.8,th-Q32) (4.2,th-DSR1) (14.1,th-GLM5) (8.1,th-MM25) (10.7,th-KK25) (15.5,th-GPTOSS) (16.3,th-GPT54) (17.6,th-CO46) (17.9,th-CS46) (2.1,ag-GPTOSS) (5.5,ag-Q32) (12.1,ag-MM25) (15.5,ag-GLM5) (14.3,ag-KK25) (16.0,ag-GPT54) (18.0,ag-CS46) (20.1,ag-CO46) (10.3,zs-GLM52) (19.7,th-GP31) (17.6,th-GLM52) (16.1,th-KK27) (20.3,ag-GP31) (20.2,ag-GLM52) (20.3,ag-KK27)};
  \addplot+[fill=err-nmexpr, draw=err-nmexpr!70!black, mark=none]
    coordinates {(9.2,zs-Q3C) (16.6,zs-L4) (19.5,zs-D2) (28.6,zs-KK25) (30.2,zs-DS32) (33.1,zs-CH45) (26.9,zs-GLM5) (42.4,zs-CS46) (43.1,zs-GPT54) (53.7,zs-CO46) (13.4,th-Q32) (14.5,th-DSR1) (26.1,th-GLM5) (14.1,th-MM25) (26.8,th-KK25) (45.4,th-GPTOSS) (57.3,th-GPT54) (59.4,th-CO46) (48.9,th-CS46) (12.2,ag-GPTOSS) (22.3,ag-Q32) (62.1,ag-MM25) (63.4,ag-GLM5) (71.8,ag-KK25) (56.1,ag-GPT54) (78.7,ag-CS46) (77.9,ag-CO46) (31.5,zs-GLM52) (65.3,th-GP31) (52.3,th-GLM52) (44.9,th-KK27) (77.9,ag-GP31) (78.5,ag-GLM52) (72.6,ag-KK27)};
  \addplot+[fill=err-cmiss,  draw=err-cmiss!70!black,  mark=none]
    coordinates {(0.1,zs-Q3C) (0.0,zs-L4) (0.0,zs-D2) (0.0,zs-KK25) (0.1,zs-DS32) (0.0,zs-CH45) (0.2,zs-GLM5) (0.0,zs-CS46) (0.0,zs-GPT54) (0.0,zs-CO46) (0.0,th-Q32) (0.0,th-DSR1) (0.0,th-GLM5) (0.8,th-MM25) (0.1,th-KK25) (0.0,th-GPTOSS) (0.3,th-GPT54) (0.0,th-CO46) (0.0,th-CS46) (0.0,ag-GPTOSS) (1.2,ag-Q32) (3.5,ag-MM25) (1.7,ag-GLM5) (0.6,ag-KK25) (0.8,ag-GPT54) (0.9,ag-CS46) (0.3,ag-CO46) (0.0,zs-GLM52) (0.5,th-GP31) (0.2,th-GLM52) (0.2,th-KK27) (0.2,ag-GP31) (0.2,ag-GLM52) (0.2,ag-KK27)};
  \addplot+[fill=err-nmkind, draw=err-nmkind!70!black, mark=none]
    coordinates {(0.0,zs-Q3C) (0.0,zs-L4) (0.1,zs-D2) (0.0,zs-KK25) (0.0,zs-DS32) (0.0,zs-CH45) (0.0,zs-GLM5) (0.0,zs-CS46) (0.0,zs-GPT54) (0.0,zs-CO46) (0.0,th-Q32) (0.0,th-DSR1) (0.0,th-GLM5) (0.0,th-MM25) (0.0,th-KK25) (0.0,th-GPTOSS) (0.0,th-GPT54) (0.0,th-CO46) (0.0,th-CS46) (0.0,ag-GPTOSS) (0.3,ag-Q32) (1.0,ag-MM25) (1.0,ag-GLM5) (0.4,ag-KK25) (0.0,ag-GPT54) (0.8,ag-CS46) (1.1,ag-CO46) (0.0,zs-GLM52) (0.0,th-GP31) (0.1,th-GLM52) (0.0,th-KK27) (0.9,ag-GP31) (0.2,ag-GLM52) (0.2,ag-KK27)};
  \addplot+[fill=err-comp,   draw=err-comp!70!black,   mark=none]
    coordinates {(87.5,zs-Q3C) (75.6,zs-L4) (72.0,zs-D2) (62.4,zs-KK25) (60.1,zs-DS32) (55.8,zs-CH45) (58.9,zs-GLM5) (42.5,zs-CS46) (41.5,zs-GPT54) (29.5,zs-CO46) (79.2,th-Q32) (78.9,th-DSR1) (59.8,th-GLM5) (25.4,th-MM25) (62.4,th-KK25) (35.9,th-GPTOSS) (24.5,th-GPT54) (21.7,th-CO46) (32.3,th-CS46) (84.1,ag-GPTOSS) (70.6,ag-Q32) (21.3,ag-MM25) (17.5,ag-GLM5) (12.9,ag-KK25) (26.7,ag-GPT54) (1.4,ag-CS46) (0.6,ag-CO46) (58.2,zs-GLM52) (14.5,th-GP31) (26.3,th-GLM52) (29.2,th-KK27) (0.7,ag-GP31) (0.9,ag-GLM52) (6.1,ag-KK27)};
  \addplot+[fill=err-tmo,    draw=err-tmo!70!black,    mark=none]
    coordinates {(0.2,zs-Q3C) (0.1,zs-L4) (0.1,zs-D2) (0.0,zs-KK25) (0.1,zs-DS32) (0.0,zs-CH45) (0.1,zs-GLM5) (0.1,zs-CS46) (0.0,zs-GPT54) (0.0,zs-CO46) (0.6,th-Q32) (2.4,th-DSR1) (0.0,th-GLM5) (51.7,th-MM25) (0.0,th-KK25) (3.3,th-GPTOSS) (1.6,th-GPT54) (1.2,th-CO46) (0.8,th-CS46) (1.5,ag-GPTOSS) (0.0,ag-Q32) (0.0,ag-MM25) (0.8,ag-GLM5) (0.0,ag-KK25) (0.4,ag-GPT54) (0.1,ag-CS46) (0.0,ag-CO46) (0.0,zs-GLM52) (0.0,th-GP31) (3.5,th-GLM52) (9.7,th-KK27) (0.0,ag-GP31) (0.0,ag-GLM52) (0.6,ag-KK27)};
  \legend{Pass, NMExpr, CMiss, NMKind, CompileFail, Timeout}
\end{axis}
\end{tikzpicture}
}
\caption{IAF error-type breakdown per (model, mode) cell.
  Each bar is a 100\% stacked composition: \textbf{Pass};
  \textbf{NMExpr} / \textbf{CMiss} / \textbf{NMKind} (compiled but
  checker-failed, resolved by precedence NMExpr~$>$~NMKind~$>$~CMiss
  so the three are mutually exclusive); \textbf{CompileFail}
  (submission did not elaborate); \textbf{Timeout} (no submission
  produced within the time/token budget).}
\label{fig:iaf-err}
\end{figure}

\begin{table}[t]
\centering
\scriptsize
\setlength{\tabcolsep}{2.5pt}
\renewcommand{\arraystretch}{1.08}

\newcommand{\modebar}[1]{
\rowcolor{gray!15}
\multicolumn{16}{c}{\textbf{#1}}\\
}

\resizebox{\linewidth}{!}{
\begin{tabular}{@{}lrrrrr@{\hspace{4pt}}ccccc@{\hspace{4pt}}ccccc@{}}
\toprule
\multirow{2}{*}{\textbf{Model}} &
\multicolumn{5}{c}{\textbf{IAF}} &
\multicolumn{5}{c}{\textbf{SSAF}} &
\multicolumn{5}{c}{\textbf{IAF-D}} \\
\cmidrule(lr){2-6} \cmidrule(lr){7-11} \cmidrule(lr){12-16}
& \multicolumn{1}{c}{\textbf{Pass}}
& \multicolumn{1}{c}{\textbf{Compile}}
& \multicolumn{1}{c}{\textbf{CMiss}}
& \multicolumn{1}{c}{\textbf{NMExpr}}
& \multicolumn{1}{c}{\textbf{NMKind}}
& \multicolumn{1}{c}{\textbf{Pass}}
& \multicolumn{1}{c}{\textbf{Compile}}
& \multicolumn{1}{c}{\textbf{CMiss}}
& \multicolumn{1}{c}{\textbf{NMExpr}}
& \multicolumn{1}{c}{\textbf{NMKind}}
& \multicolumn{1}{c}{\textbf{Pass}}
& \multicolumn{1}{c}{\textbf{Compile}}
& \multicolumn{1}{c}{\textbf{CMiss}}
& \multicolumn{1}{c}{\textbf{NMExpr}}
& \multicolumn{1}{c}{\textbf{NMKind}} \\
\midrule

\modebar{Zero-shot}
D2    & $8.4_{\pm 0.37}$  & $27.9_{\pm 1.24}$ & $2.7_{\pm 0.41}$  & $19.4_{\pm 0.97}$ & $0.1_{\pm 0.10}$ & $1.7$ & $7.8$  & $0.9$  & $6.0$  & $0.0$ & $7.6$  & $18.0$ & $2.1$ & $10.1$ & $0.0$ \\
DS32  & $9.5_{\pm 0.61}$  & $39.8_{\pm 0.81}$ & $6.3_{\pm 0.10}$  & $29.9_{\pm 0.54}$ & $0.2_{\pm 0.10}$ & $1.7$ & $18.1$ & $5.2$  & $16.4$ & $0.0$ & $7.3$  & $33.0$ & $4.0$ & $25.4$ & $0.0$ \\
GLM5  & $13.9_{\pm 0.62}$ & $41.0_{\pm 0.71}$ & $4.6_{\pm 0.31}$  & $26.8_{\pm 1.18}$ & $0.0_{\pm 0.00}$ & $0.9$ & $10.3$ & $1.7$  & $9.5$  & $0.0$ & $10.7$ & $36.4$ & $3.1$ & $25.4$ & $0.3$ \\
GLM52 & $10.3_{\pm 0.10}$ & $41.8_{\pm 0.82}$ & $3.6_{\pm 0.10}$ & $31.5_{\pm 0.77}$ & $0.0_{\pm 0.00}$ & -- & -- & -- & -- & -- & -- & -- & -- & -- & -- \\
GPT54 & $15.4_{\pm 0.41}$ & $58.5_{\pm 1.35}$ & $7.6_{\pm 0.77}$  & $42.5_{\pm 1.07}$ & $0.0_{\pm 0.00}$ & -- & -- & -- & -- & -- & -- & -- & -- & -- & -- \\
CH45  & $11.1_{\pm 0.27}$ & $44.2_{\pm 1.64}$ & $6.7_{\pm 0.64}$  & $32.7_{\pm 1.79}$ & $0.0_{\pm 0.00}$ & $2.6$ & $19.8$ & $2.6$  & $17.2$ & $0.0$ & $9.8$  & $36.4$ & $4.9$ & $26.3$ & $0.0$ \\
KK25   & $9.0_{\pm 0.67}$  & $37.6_{\pm 0.77}$ & $5.0_{\pm 0.71}$  & $27.7_{\pm 0.37}$ & $0.0_{\pm 0.00}$ & $2.6$ & $13.8$ & $2.6$  & $11.2$ & $0.0$ & $7.0$  & $33.0$ & $3.7$ & $26.0$ & $0.3$ \\
L4    & $7.6_{\pm 0.18}$  & $24.3_{\pm 0.62}$ & $2.1_{\pm 0.31}$  & $16.6_{\pm 0.74}$ & $0.0_{\pm 0.00}$ & $0.9$ & $12.9$ & $0.9$  & $12.1$ & $0.0$ & $6.4$  & $21.7$ & $1.5$ & $15.0$ & $0.0$ \\
CO46  & $16.8_{\pm 0.47}$ & $70.5_{\pm 0.87}$ & $12.8_{\pm 0.64}$ & $52.9_{\pm 1.33}$ & $0.9_{\pm 0.18}$ & $2.6$ & $50.0$ & $16.4$ & $47.4$ & $1.7$ & $14.1$ & $64.5$ & $8.9$ & $48.6$ & $0.9$ \\
Q3C   & $3.1_{\pm 0.31}$  & $12.3_{\pm 0.74}$ & $2.3_{\pm 0.27}$  & $9.0_{\pm 0.57}$  & $0.0_{\pm 0.00}$ & $0.0$ & $0.0$  & $0.0$  & $0.0$  & $0.0$ & $3.4$  & $8.3$  & $0.6$ & $4.6$  & $0.3$ \\
CS46  & $15.0_{\pm 0.31}$ & $57.4_{\pm 0.44}$ & $8.9_{\pm 0.53}$  & $41.2_{\pm 0.51}$ & $0.0_{\pm 0.00}$ & $\underline{5.2}$ & $40.5$ & $9.5$ & $35.3$ & $0.0$ & $12.8$ & $55.4$ & $7.6$ & $41.9$ & $0.0$ \\

\midrule
\modebar{Thinking}
DSR1   & $4.2_{\pm 0.10}$  & $18.7_{\pm 0.18}$ & $2.9_{\pm 0.10}$  & $14.0_{\pm 0.10}$ & $0.0_{\pm 0.00}$ & $0.9$ & $4.3$  & $0.9$  & $3.4$  & $0.0$ & $2.8$  & $8.9$  & $2.1$  & $6.1$  & $0.0$ \\
GLM5   & $14.1_{\pm 0.35}$ & $40.2_{\pm 0.51}$ & $3.3_{\pm 0.44}$ & $26.1_{\pm 0.62}$ & $0.0_{\pm 0.00}$ & $0.9$ & $9.5$  & $3.4$  & $8.6$  & $0.0$ & $4.3$  & $18.7$ & $2.4$  & $14.4$ & $0.0$ \\
GLM52 & $17.6_{\pm 0.37}$ & $70.2_{\pm 1.15}$ & $9.6_{\pm 0.54}$ & $52.3_{\pm 1.41}$ & $0.5_{\pm 0.10}$ & -- & -- & -- & -- & -- & -- & -- & -- & -- & -- \\
GP31 & $19.7_{\pm 0.44}$ & $85.5_{\pm 0.20}$ & $13.1_{\pm 0.88}$ & $65.3_{\pm 0.57}$ & $0.1_{\pm 0.10}$ & -- & -- & -- & -- & -- & -- & -- & -- & -- & -- \\
GPT54  & $16.3_{\pm 0.57}$ & $73.9_{\pm 0.97}$ & $12.5_{\pm 0.92}$ & $55.9_{\pm 1.43}$ & $0.2_{\pm 0.10}$ & -- & -- & -- & -- & -- & -- & -- & -- & -- & -- \\
GPTOSS & $15.5_{\pm 0.44}$ & $60.9_{\pm 0.61}$ & $11.4_{\pm 1.33}$ & $44.8_{\pm 0.10}$ & $0.0_{\pm 0.00}$ & -- & -- & -- & -- & -- & -- & -- & -- & -- & -- \\
KK25    & $10.7_{\pm 0.35}$ & $37.6_{\pm 2.84}$ & $4.6_{\pm 1.06}$ & $26.8_{\pm 2.74}$ & $0.2_{\pm 0.10}$ & $\mathbf{6.9}$ & $31.9$ & $6.0$ & $25.0$ & $0.0$ & $11.0$ & $45.6$ & $6.7$ & $33.6$ & $0.0$ \\
KK27 & $16.1_{\pm 0.27}$ & $61.2_{\pm 1.45}$ & $5.9_{\pm 0.41}$ & $44.9_{\pm 1.14}$ & $0.1_{\pm 0.10}$ & -- & -- & -- & -- & -- & -- & -- & -- & -- & -- \\
MM25   & $8.1_{\pm 0.20}$ & $22.9_{\pm 0.47}$ & $2.8_{\pm 0.18}$ & $14.1_{\pm 0.61}$ & $0.2_{\pm 0.10}$ & $1.7$ & $5.2$  & $0.9$  & $2.6$  & $0.0$ & $10.1$ & $40.4$ & $7.0$ & $28.7$ & $0.0$ \\
CO46   & $17.6_{\pm 0.27}$ & $77.1_{\pm 1.70}$ & $14.1_{\pm 0.47}$ & $58.3_{\pm 1.58}$ & $1.0_{\pm 0.27}$ & $4.3$ & $56.9$ & $14.7$ & $52.6$ & $1.7$ & $\mathbf{16.2}$ & $\mathbf{76.5}$ & $13.8$ & $59.3$ & $1.2$ \\
Q32    & $6.8_{\pm 0.10}$ & $20.2_{\pm 0.81}$ & $2.1_{\pm 0.18}$ & $13.4_{\pm 0.71}$ & $0.1_{\pm 0.10}$ & $1.7$ & $6.0$  & $0.9$  & $4.3$  & $0.9$ & $5.2$  & $13.5$ & $1.5$  & $8.0$  & $0.0$ \\
CS46   & $17.9_{\pm 0.82}$ & $66.9_{\pm 0.97}$ & $10.1_{\pm 1.27}$ & $48.5_{\pm 1.12}$ & $0.3_{\pm 0.18}$ & $2.6$ & $42.2$ & $11.2$ & $39.7$ & $0.0$ & $\underline{15.6}$ & $\underline{67.0}$ & $9.8$ & $50.2$ & $0.3$ \\

\midrule
\modebar{Agentic}
GLM5   & $15.5_{\pm 0.20}$ & $81.7_{\pm 0.71}$ & $21.1_{\pm 0.88}$ & $63.4_{\pm 0.71}$ & $3.3_{\pm 0.80}$ & $4.3$ & $\mathbf{96.6}$ & $46.6$ & $91.4$ & $9.5$ & -- & -- & -- & -- & -- \\
GLM52 & $\underline{20.2}_{\pm 0.53}$ & $99.1_{\pm 0.18}$ & $17.2_{\pm 0.37}$ & $78.5_{\pm 0.71}$ & $0.3_{\pm 0.18}$ & -- & -- & -- & -- & -- & -- & -- & -- & -- & -- \\
GP31 & $\mathbf{20.3}_{\pm 0.27}$ & $\underline{99.3}_{\pm 0.27}$ & $17.0_{\pm 0.10}$ & $77.9_{\pm 0.37}$ & $2.4_{\pm 0.47}$ & -- & -- & -- & -- & -- & -- & -- & -- & -- & -- \\
GPT54  & $16.1_{\pm 0.37}$ & $73.4_{\pm 0.64}$ & $13.0_{\pm 0.44}$ & $55.0_{\pm 0.47}$ & $0.1_{\pm 0.10}$ & -- & -- & -- & -- & -- & -- & -- & -- & -- & -- \\
GPTOSS & $2.1_{\pm 0.31}$  & $14.4_{\pm 0.61}$ & $3.9_{\pm 0.62}$  & $12.2_{\pm 0.31}$ & $0.5_{\pm 0.10}$ & -- & -- & -- & -- & -- & -- & -- & -- & -- & -- \\
KK25    & $14.3_{\pm 1.25}$ & $87.1_{\pm 0.44}$ & $22.4_{\pm 0.44}$ & $71.8_{\pm 1.15}$ & $2.7_{\pm 0.54}$ & $3.4$ & $\underline{92.2}$ & $41.4$ & $88.8$ & $6.9$ & -- & -- & -- & -- & -- \\
KK27 & $\mathbf{20.3}_{\pm 0.54}$ & $93.3_{\pm 0.35}$ & $15.2_{\pm 0.37}$ & $72.6_{\pm 0.20}$ & $0.5_{\pm 0.27}$ & -- & -- & -- & -- & -- & -- & -- & -- & -- & -- \\
MM25   & $12.1_{\pm 0.10}$ & $78.7_{\pm 1.12}$ & $25.7_{\pm 0.77}$ & $61.3_{\pm 0.71}$ & $3.3_{\pm 0.62}$ & $2.6$ & $85.3$ & $44.8$ & $81.0$ & $7.8$ & -- & -- & -- & -- & -- \\
CO46   & $20.1_{\pm 0.67}$ & $\mathbf{99.4}_{\pm 0.47}$ & $21.1_{\pm 0.18}$ & $76.1_{\pm 0.61}$ & $4.1_{\pm 0.54}$ & $\underline{5.2}$ & $\mathbf{96.6}$ & $38.8$ & $91.4$ & $3.4$ & -- & -- & -- & -- & -- \\
Q32
& \multicolumn{1}{c}{$5.5$}
& \multicolumn{1}{c}{$29.4$}
& \multicolumn{1}{c}{$5.2$}
& \multicolumn{1}{c}{$22.0$}
& \multicolumn{1}{c}{$0.3$}
& $0.0$ & $0.9$ & $0.9$ & $0.9$ & $0.0$
& -- & -- & -- & -- & -- \\
CS46   & $18.0_{\pm 0.53}$ & $98.5_{\pm 0.18}$ & $23.4_{\pm 0.37}$ & $77.5_{\pm 0.67}$ & $3.9_{\pm 0.20}$ & $3.4$ & $\mathbf{96.6}$ & $41.4$ & $93.1$ & $6.9$ & -- & -- & -- & -- & -- \\

\bottomrule
\end{tabular}
}
\caption{Per-cell performance across three AF tracks. \textbf{Pass} = strict binary pass rate (\%) — submission passes Algorithm~\ref{alg:check}. \textbf{Compile} = fraction of items whose submission elaborates as valid Lean (header success, regardless of constant matching). \textbf{CMiss} / \textbf{NMExpr} / \textbf{NMKind} report the percentage of items whose grader output contains at least one \texttt{constmissing}, \texttt{notmatchexpr}, or \texttt{notmatchconstkind} error. An item can contribute to multiple error columns if it triggers several types. Rows are grouped by inference mode using shaded separator bars. Dashes indicate no curated run for that cell. IAF values are mean $\pm$ SEM across three independent runs; SSAF and IAF-D are single runs reported without variance.}
\label{table:af-eval-passrate}
\end{table}

\subsubsection{Model I/O}
\label{sec:appendix-eval-io}

For non-agentic modes (zero-shot and thinking), the prompt is the concatenation of a mode-specific template with fenced blocks for \texttt{Context.lean}, \texttt{Starter.lean.in}, and (when present) \texttt{Quote.md}.
The response is parsed by taking the last fenced \texttt{lean4}/\texttt{lean} block.
If no fenced block is recovered, the model response to the item is recorded with \texttt{status=missing}.

For agentic mode, the model writes its submission directly to \texttt{Submission.lean} over successive turns; the initial \texttt{Submission.lean} is seeded from the starter with target bodies initialized to \texttt{sorry}.
When the loop exits, the final on-disk \texttt{Submission.lean} is taken as $\hat{y}$.

\subsubsection{Modes and Agent Tools}
\label{sec:appendix-eval-modes}

Thinking variants enable the provider-native extended-reasoning channel.
For models with a binary on/off toggle, we enable thinking; for models that expose graded effort levels, we set \texttt{reasoning\_effort=high}.
For non-Anthropic models, reasoning tokens and visible-answer tokens share a single budget, which we set to \texttt{max\_tokens=32,768}.
Opus and Sonnet expose the thinking budget and the visible-output budget as separate caps, which we both set to \texttt{16,384}.

The agentic scaffold is a lightweight SWE-agent-style loop based on mini-swe-agent~\cite{yang2024swe}.
In place of the single \texttt{bash} tool of the original scaffold, we expose four custom tools tailored to Lean editing:

\begin{itemize}[leftmargin=*, topsep=2pt, itemsep=1pt, parsep=0pt]
\item \texttt{read\_file(which)} --- reads one of \texttt{context},
\texttt{quote}, or \texttt{submission}.
\item \texttt{edit\_submission(old\_str, new\_str)} --- string-replacement
edit on \texttt{Submission.lean} that must match \texttt{old\_str}
exactly once.
\item \texttt{lake\_build()} --- runs \texttt{lake env lean} on the
current \texttt{Submission.lean} and returns the compiler output.
\item \texttt{done(reason?)} --- exits the loop.
\end{itemize}

\subsubsection{Scoring}
\label{sec:appendix-eval-scoring}

\textbf{AF tracks.}
The DefEq checker emits a per-target triple of flags: CMiss (constant missing: expected constant not declared), NMkind (Not matching kind: kind mismatch such as \texttt{def} vs. \texttt{theorem}), and NMExpr (Not matching expression: body or statement not definitionally equivalent).
An item is recorded as \texttt{status=pass} iff it passes Algorithm~\ref{alg:check}.
When the submission fails to elaborate, the item is assigned CompileFail (compilation failure).
A more detailed error type breakdown of the IAF track evaluation is shown in Figure~\ref{fig:iaf-err}.
A more comprehensive table showing the evaluation results of all three AF tracks is in Table~\ref{table:af-eval-passrate}.

\textbf{TP tracks.}
A TP submission is accepted as \texttt{pass} iff all four of the following hold:
\begin{enumerate}[leftmargin=*, topsep=2pt, itemsep=1pt, parsep=0pt]

\item \emph{Declarations present.} Every target in $T_y$ appears in $\hat{y}$ with a type signature whitespace-equivalent to its reference signature.

\item \emph{Build succeeds.} \texttt{lake env lean} on $\hat{y}$ returns exit code zero; warnings are permitted.

\item \emph{No \texttt{sorry}.} After stripping comments and string literals, no \texttt{sorry} keyword remains.

\item \emph{No cheats.} The stripped source contains none of: \texttt{axiom <name>}, \texttt{opaque <name>}, \texttt{unsafe} declarations, \texttt{@[extern]}, \texttt{@[implemented\_by]}, or direct uses of \texttt{sorryAx}.

\end{enumerate}
Any condition failing yields \texttt{status=fail}.

\phantomsection\label{sec:appendix-fairness}
\textbf{Benchmark fairness.}
Because the \ours artifact was constructed with the assistance of Claude Opus~4.6 and GPT~5.3, we compare the IAF solve sets of the four Claude and GPT models evaluated in both thinking and agentic modes (CO46, CS46, GPT54, GPTOSS) with those of six models from other families (GP31, GLM52, KK27, GLM5, KK25, MM25).
An item counts as solved by a model if any of its three one-shot thinking or three agentic runs passes our DefEq checker, and as solved by a family if any of its models solves it.
Table~\ref{tab:fairness-family} reports family-level coverage over the 327 IAF items, and Table~\ref{tab:fairness-model} reports per-model solves and unique solves, i.e., items solved by exactly one of the ten models.
Of the 101 items solved by Claude or GPT models, only 7 are solved by no other family, while 19 items are solved only by other families.
In agentic mode, GP31, KK27, and GLM52 also exceed every Claude and GPT model in average pass rate, as shown in Figure~\ref{fig:iaf-passrate}.
The small family-exclusive tails indicate that item difficulty is governed by the formalization content, not by model-specific familiarity with the artifact.

\begin{table}[h]
\centering
\small
\begin{minipage}[t]{0.46\linewidth}
\centering
\begin{tabular}{lrr}
\toprule
Region & Items & \% \\
\midrule
Solved by Claude/GPT & 101 & 30.9 \\
Solved by other families & 113 & 34.6 \\
Solved by both & 94 & 28.7 \\
Only by Claude/GPT & 7 & 2.1 \\
Only by other families & 19 & 5.8 \\
Solved by none & 207 & 63.3 \\
\bottomrule
\end{tabular}
\captionof{table}{Family-level IAF coverage.}
\label{tab:fairness-family}
\end{minipage}
\hfill
\begin{minipage}[t]{0.50\linewidth}
\centering
\begin{tabular}{lrrr}
\toprule
Model & Solved & \% & Unique \\
\midrule
GLM52 & 90 & 27.5 & 2 \\
KK27 & 88 & 26.9 & 3 \\
CO46 & 87 & 26.6 & 4 \\
GP31 & 86 & 26.3 & 4 \\
CS46 & 81 & 24.8 & 1 \\
KK25 & 75 & 22.9 & 2 \\
GPT54 & 73 & 22.3 & 0 \\
GLM5 & 71 & 21.7 & 2 \\
GPTOSS & 66 & 20.2 & 0 \\
MM25 & 66 & 20.2 & 1 \\
\bottomrule
\end{tabular}
\captionof{table}{Per-model IAF solves.}
\label{tab:fairness-model}
\end{minipage}
\end{table}

\phantomsection\label{sec:appendix-passk}
\textbf{Pass@$k$.}
Table~\ref{tab:passk} reports IAF pass@3, the fraction of the 327 items solved in at least one of three runs, for every model and mode; average pass rates are shown in Figure~\ref{fig:iaf-passrate}.

\begin{table}[h]
\centering
\small
\begin{tabular}{lr@{\hspace{2.5em}}lr@{\hspace{2.5em}}lr}
\toprule
\multicolumn{2}{c}{Zero-shot} & \multicolumn{2}{c}{Thinking} & \multicolumn{2}{c}{Agentic} \\
\cmidrule(r{2.5em}){1-2}\cmidrule(r{2.5em}){3-4}\cmidrule{5-6}
Model & Pass@3 & Model & Pass@3 & Model & Pass@3 \\
\midrule
CO46 & 19.0 & GLM52 & 23.9 & KK27 & 25.7 \\
GPT54 & 18.0 & GP31 & 23.5 & GLM52 & 25.1 \\
CS46 & 18.0 & CO46 & 22.6 & GP31 & 23.9 \\
GLM5 & 16.8 & CS46 & 21.7 & CO46 & 23.5 \\
GLM52 & 15.9 & KK27 & 20.5 & KK25 & 21.1 \\
KK25 & 14.1 & GPTOSS & 19.9 & CS46 & 20.8 \\
CH45 & 13.8 & GPT54 & 19.0 & GPT54 & 20.8 \\
DS32 & 13.5 & GLM5 & 17.4 & GLM5 & 19.3 \\
D2 & 11.9 & KK25 & 15.6 & MM25 & 17.4 \\
L4 & 9.5 & MM25 & 12.8 & GPTOSS & 5.5 \\
Q3C & 4.3 & Q32 & 9.8 & Q32 & -- \\
 & & DSR1 & 8.0 & & \\
\bottomrule
\end{tabular}
\caption{IAF pass@3 (\%) under our DefEq checker. Q32 agentic has a single run.}
\label{tab:passk}
\end{table}

\subsection{Data Engine}
\label{sec:appendix-data-engine}

We develop a deterministic, artifact-grounded data engine that performs systematic dependency analysis and Lean code modification to automatically generate all evaluation tracks from the \ours Lean artifact.
This process is AI-free and fully reproducible: whatever is stabilized in the artifact is faithfully used to derive the evaluation datasets without any model intervention.

For AF targets, we specify which Lean declarations to include as targets, and the engine fetches all transitive dependencies, removes unwanted proof bodies, and composes them into a compact context $C$ while generating the starter template $T_y$ and reference formalization $y^*$.
IAF-D is generated in the same way, with the addition of distractor declarations injected beyond the minimal dependency set; specifically, for each instance, we add distractors amounting to half the number of real context dependencies.

For TP targets, the engine replaces proof bodies with \texttt{sorry} to produce $C$.
A TP data point is generated only when the entire dependency tree is well-formed and fully proved; unproved declarations in the current artifact are therefore excluded from the TP tracks.

By adjusting the engine's configuration, it is straightforward to generate further scaled-up variants such as CAF (chapter-level auto-formalization) or SSTP (subsection-level theorem-proving).
Through our evaluation, however, we find these variants too difficult for current models and leave them for future work.

\section{Equivalence Checker}
\label{sec:appendix-defeq-checker}

To evaluate the faithfulness of each submission, we developed \texttt{Check.lean}, a program
that verifies every constant declared in \texttt{GroundTruth.lean} appears in
\texttt{Submission.lean} in a semantically equivalent form, and that both files share the
same imports. Equivalence is defined differently depending on the \emph{kind} of constant,
as summarised in Table~\ref{tab:equiv}.

\begin{table}[h]
\centering
\small
\begin{tabular}{lll}
\toprule
\textbf{Kind} & \textbf{What is compared} & \textbf{Equivalence criterion} \\
\midrule
Definition   & Body expression    & Definitionally equal (\textsc{isDefEq}) or congruence closure (\textsc{grind}) \\
Theorem      & Statement type     & Definitionally equal (\textsc{isDefEq}) or congruence closure (\textsc{grind}) \\
Inductive    & Constructor names  & Same set of constructors, order-independent (structural check) \\
Constructor  & Constructor type   & Definitionally equal (\textsc{isDefEq} only, no \textsc{grind}) \\
\bottomrule
\end{tabular}
\caption{Equivalence criteria used by Equivalence Checker for each constant kind.}
\label{tab:equiv}
\end{table}

\noindent\textbf{Definition}: Two definitions are considered equivalent if the Lean
expressions corresponding to their bodies are equal after full reduction, so different
but computationally equivalent implementations are treated as equal.
\vspace{4pt}

\noindent
\begin{minipage}[t]{0.48\textwidth}
\begin{tcolorbox}[codebox, title=Reference]
\begin{lstlisting}
def double (n : Nat) : Nat :=
  n + n
\end{lstlisting}
\end{tcolorbox}
\end{minipage}
\hfill
\begin{minipage}[t]{0.48\textwidth}
\begin{tcolorbox}[codebox, title=Submission (\protect\cmark\ accepted)]
\begin{lstlisting}
def double (n : Nat) : Nat :=
  2 * n
\end{lstlisting}
\end{tcolorbox}
\end{minipage}

\vspace{8pt}

\noindent\textbf{Theorem}: For theorems (which are effectively prop-typed non computable definitions), comparing types alone is too loose: since under proposition extensionality all true propositions are provably equal, meaning any two
true theorems would be considered equivalent. We therefore require the statement types to
be definitionally equal (\textsc{isDefEq}) or equal under congruence closure
(\textsc{grind}), which captures the notion that two theorems are talking about the same idea but just phrased slightly differently.
\vspace{4pt}

\noindent
\begin{minipage}[t]{0.48\textwidth}
\begin{tcolorbox}[codebox, title=Reference]
\begin{lstlisting}
theorem add_zero (n : Nat)
    : n + 0 = n := by simp
\end{lstlisting}
\end{tcolorbox}
\end{minipage}
\hfill
\begin{minipage}[t]{0.48\textwidth}
\begin{tcolorbox}[codebox, title=Submission (\protect\cmark\ accepted)]
\begin{lstlisting}
theorem add_zero (n : Nat) :
  n + 0 = n := Nat.add_zero n
\end{lstlisting}
\end{tcolorbox}
\end{minipage}

\vspace{8pt}

\noindent\textbf{Inductive}: Two inductive definitions are equivalent if they share the
same set of constructor names (up to permutation). Each constructor is then checked
independently: the constructor from the submission must have the same type as the
corresponding constructor in the reference (by \textsc{isDefEq}).
\vspace{4pt}

\noindent
\begin{minipage}[t]{0.48\textwidth}
\begin{tcolorbox}[codebox, title=Reference]
\begin{lstlisting}
inductive Shape where
  | circle (r : Float)
  | rect   (w h : Float)
\end{lstlisting}
\end{tcolorbox}
\end{minipage}
\hfill
\begin{minipage}[t]{0.48\textwidth}
\begin{tcolorbox}[codebox, title=Submission (\protect\cmark\ accepted)]
\begin{lstlisting}
inductive Shape where
  | rect   (w h : Float)
  | circle (r : Float)
\end{lstlisting}
\end{tcolorbox}
\end{minipage}

\noindent\textbf{Binder renaming resilience.} Moreover, by using Lean's definitional equality instead of structural equality of the underlying expressions, our checker is resilient to renaming of locally bound variable names, since changing binder names preserves definitional equality.

\noindent
\begin{minipage}[t]{0.48\textwidth}
\begin{tcolorbox}[codebox, title=Reference]
\begin{lstlisting}
def comp (f g : Nat -> Nat)
  (x : Nat) : Nat := f (g x)
\end{lstlisting}
\end{tcolorbox}
\end{minipage}
\hfill
\begin{minipage}[t]{0.48\textwidth}
\begin{tcolorbox}[codebox, title=Submission (\protect\cmark\ accepted)]
\begin{lstlisting}
def comp (f g : Nat -> Nat)
  (n : Nat) : Nat := f (g n)
\end{lstlisting}
\end{tcolorbox}
\end{minipage}

\noindent\textbf{Sandboxing.} The constants of the Reference and Submission are elaborated in their respective environments, and each constant's body is fully reduced in its own environment before comparison using \texttt{reduceFully}, which runs Lean's kernel reducer at full transparency.
This means all definitions are unfolded, including typeclass projections and instance fields that would otherwise be opaque at lower transparency levels.
As a result, any semantic pollution introduced in the Submission is inlined away during reduction, exposing the underlying kernel primitives rather than the Submission's local definitions.
The comparison itself is then run in the Reference environment elaborated only up to (but not including) $c$, ensuring that the Submission's local definitions are not in scope and that $c$ itself is not in scope during its own check.

The algorithm for the \texttt{Check.lean} program is summarized below:
\begin{algorithm}[H]
\caption{\texttt{Check}(\textit{Reference}, \textit{Submission})}
\label{alg:check}
\begin{algorithmic}[1]

\State Elaborate both files; abort on errors. Flag $\mathit{passed} \leftarrow \textbf{false}$ if imports differ.

\For{each constant $c$ in \textit{Reference} (excluding auto-generated recursors)}
    \If{$c \notin \textit{Submission}$ \textbf{or} $\mathrm{kind}(c)$ differs} record failure; \textbf{continue} \EndIf
    \State Let $e_1, e_2$ be the fully reduced relevant expressions of $c$ in each file
    \State \textbf{Definition / Theorem:} $\textsc{Equiv}(e_1, e_2, \text{grind}=\textbf{true})$ \Comment{body / statement type}
    \State \textbf{Inductive:} success iff constructors of $c$ are the same set in both files
    \State \textbf{Constructor:} $\textsc{Equiv}(e_1, e_2, \text{grind}=\textbf{false})$ \Comment{constructor type}
\EndFor

\State \Return $\mathit{passed}\ \wedge$ all checks succeeded

\Statex
\Function{Equiv}{$e_1,\, e_2,\, \text{grind}$}
    \If{$\textsc{isDefEq}(e_1, e_2)$} \Return \textsc{Success} \EndIf
    \If{$\text{grind} \wedge \textsc{grind}(e_1 = e_2)$ succeeds} \Return \textsc{Success} \EndIf
    \State \Return \textsc{Mismatch}
\EndFunction

\end{algorithmic}
\end{algorithm}

\phantomsection\label{sec:appendix-beq}
\textbf{Comparison with BEq+.}
We also evaluate the theorem-kind declarations of every IAF run with BEq+~\cite{poiroux2025reliable} (Tables~\ref{tab:beq-union} and~\ref{tab:beq-confusion}).
Admitting BEq+ as an additional acceptance route raises the average pass rate by only 0.4 points, and the two checkers agree on 90.9\% of the 19,072 declarations to which BEq+ applies.
Our checker is thus comparable with BEq+ on propositions, while also covering the definitions, inductive types, and examples that BEq+ does not.

\begin{table}[h]
\centering
\footnotesize
\begin{tabular}{lrr@{\hspace{2em}}lrr@{\hspace{2em}}lrr}
\toprule
\multicolumn{3}{c}{Zero-shot} & \multicolumn{3}{c}{Thinking} & \multicolumn{3}{c}{Agentic} \\
\cmidrule(r{2em}){1-3}\cmidrule(r{2em}){4-6}\cmidrule{7-9}
Model & Ours & Ours$\,\cup\,$BEq+ & Model & Ours & Ours$\,\cup\,$BEq+ & Model & Ours & Ours$\,\cup\,$BEq+ \\
\midrule
CO46 & 16.8 & 16.9 & GP31 & 19.7 & 20.2 & KK27 & 20.3 & 21.1 \\
GPT54 & 15.4 & 15.7 & CS46 & 17.9 & 18.3 & GP31 & 20.3 & 20.8 \\
CS46 & 15.0 & 15.0 & CO46 & 17.6 & 18.3 & GLM52 & 20.2 & 21.0 \\
GLM5 & 13.9 & 14.5 & GLM52 & 17.6 & 18.8 & CO46 & 20.1 & 20.5 \\
CH45 & 11.1 & 11.8 & GPT54 & 16.3 & 16.6 & CS46 & 18.0 & 18.1 \\
DS32 & 9.5 & 9.5 & KK27 & 16.1 & 16.4 & GPT54 & 16.1 & 16.3 \\
KK25 & 9.0 & 9.3 & GPTOSS & 15.5 & 15.9 & GLM5 & 15.5 & 16.3 \\
D2 & 8.4 & 8.5 & GLM5 & 14.1 & 14.8 & KK25 & 14.3 & 14.9 \\
\bottomrule
\end{tabular}
\caption{Average IAF pass rate (\%) under our DefEq checker alone (Ours) and with BEq+ admitted as an additional acceptance route (Ours~$\cup$~BEq+).}
\label{tab:beq-union}
\end{table}

\begin{table}[h]
\centering
\small
\begin{tabular}{lrrr}
\toprule
 & BEq+ pass & BEq+ fail & Total \\
\midrule
Ours pass & 10,769 (56.5\%) & 1,036 (5.4\%) & 11,805 (61.9\%) \\
Ours fail & 708 (3.7\%) & 6,559 (34.4\%) & 7,267 (38.1\%) \\
\midrule
Total & 11,477 (60.2\%) & 7,595 (39.8\%) & 19,072 \\
\bottomrule
\end{tabular}
\caption{Declaration-level confusion matrix of our DefEq checker against BEq+ over the declaration comparisons to which BEq+ applies.}
\label{tab:beq-confusion}
\end{table}

\section{Feature Tagging and Analysis}
\label{sec:appendix-tagging-analysis}

\subsection{Tagging schema}
\label{ssec:tag-schema}

Each benchmark item is annotated with tags drawn from five orthogonal
categories, capturing different dimensions along which
auto-formalization difficulty may vary.
\emph{Surface features} (Table~\ref{tab:nl-surface-features}) describe
the natural-language presentation of the textbook item, such as whether
it uses schematic metavariables, ellipsis notation, enumerated
clauses, or refers to a figure—properties that affect how readily the
prose transcribes into Lean syntax.
\emph{Subject matter} (Table~\ref{tab:subject-matter}) categorizes the
mathematical content of the item along the syntax--semantics axis,
distinguishing claims about derivations and formulas (\emph{syntactic})
from claims about models and truth (\emph{semantic}), with further
tags for meta-theorems, object-level claims, and algorithmic content.
\emph{Logical shape} (Table~\ref{tab:logical-shape}) records the
high-level logical form of the statement, including its dominant
quantifier (\emph{universal}, \emph{existential},
\emph{mixed-quantifier}, \emph{quantifier-free}), connective structure
(\emph{conjunctive}, \emph{disjunctive}, \emph{conditional},
\emph{iff}, \emph{negative-claim}), and whether it is \emph{multi-part}
or asserts an \emph{equality-claim}.
\emph{Vocabulary} (Table~\ref{tab:vocabulary}) tags the mathematical
machinery the item relies on, ranging from \emph{set-theoretic} and
\emph{higher-order} constructions to \emph{binders-substitution},
\emph{numeric-arithmetic}, \emph{string-level}, and \emph{infinitary}
reasoning.
Finally, \emph{definition shape} (Table~\ref{tab:definition-shape})
applies to definitional items and identifies their structural form:
whether the item introduces a \emph{type-definition},
\emph{inductive-definition}, \emph{function-definition},
\emph{predicate-definition}, or \emph{recursive-definition}, and
whether it is \emph{parametric}, \emph{notational}, uses
\emph{mutual-recursion}, or \emph{defines-multiple} concepts at once.
Tags within each category are non-exclusive, since a single item may
exhibit several features simultaneously (e.g., a universally quantified
meta-theorem that bridges syntax and semantics).

Tag assignments are produced by the pre-processing stage of the
LCS-Bench pipeline: a model proposes tags
from a fixed controlled vocabulary, every assignment is reviewed by
human experts, and disagreements are reconciled before the item is
admitted to the released artifact. The five tables below report the
absolute count and percentage of items in $N{=}554$ tagged items
carrying each tag.

\begin{table}[t]                          
    \centering                                                                             
    \small
    \begin{tabular}{@{}l p{0.50\linewidth} r r@{}}                                                         
      \toprule                                                                                             
      \textbf{Tag} & \textbf{Description} & \textbf{Items} & \textbf{\%} \\                                
      \midrule                                                                                             
      \texttt{uses-metavariable}  & Uses schematic letters (e.g.\ $A$, $B$ for arbitrary formulas; $M$, $N$
   for models). & 432 & 78.0 \\                                                                            
      \texttt{uses-ellipsis}      & Uses ``\ldots'' or \verb|\dots| for continuation, as in                
  $a_1,\dots,a_n$. & 165 & 29.8 \\                                                                         
      \texttt{enumerated-clauses} & Body lists items as (i)(ii)(iii) or (a)(b)(c). & 134 & 24.2 \\
      \texttt{name-introducing}   & Uses ``we call $X$ a $Y$'', ``$X$ is said to be $Y$'', ``we say''. & 95
   & 17.1 \\                                                                                               
      \texttt{concrete-instance}  & Claim about a specific object or example rather than a schema. & 88 &  
  15.9 \\                                                                                                  
      \texttt{informal-notation}  & Visibly informal or typographic notation that will not transcribe
  verbatim. & 40 & 7.2 \\                                                                                  
      \texttt{references-figure}  & Refers to a figure, diagram, or tree picture. & 34 & 6.1 \\
      \bottomrule                                                                                          
    \end{tabular}      
    \caption{Natural-language surface features tagged on benchmark items ($N{=}554$). Tags are             
  non-exclusive, so column sums exceed 100\%.}                     
    \label{tab:nl-surface-features}
  \end{table}

 \begin{table}[t]                                                                                         
    \centering                                                                                             
    \small                                                                                                 
    \begin{tabular}{@{}l p{0.50\linewidth} r r@{}}
      \toprule                                                                                             
      \textbf{Tag} & \textbf{Description} & \textbf{Items} & \textbf{\%} \\                              
      \midrule                                                                                             
      \texttt{syntactic}                & Talks about derivations, formulas, or strings only. & 464 & 83.8
  \\                                                                                                       
      \texttt{meta-theorem}             & Claim about the proof system itself (soundness, completeness,
  cut-elimination, \ldots). & 219 & 39.5 \\                                                                
      \texttt{semantic}                 & Talks about models, satisfaction, or truth. & 184 & 33.2 \\
      \texttt{bridges-syntax-semantics} & Relates syntactic $\vdash$ to semantic $\vDash$. & 132 & 23.8 \\ 
      \texttt{object-level}             & Claim about a derivation, sequent, or formula inside a system. & 
  80 & 14.4 \\                                                                                             
      \texttt{algorithm}                & Describes a procedure or algorithm. & 56 & 10.1 \\               
              
      \bottomrule                                                                                          
    \end{tabular}
    \caption{Subject-matter tags on benchmark items ($N{=}554$). Tags are non-exclusive, so column sums    
  exceed 100\%.}                                                                                           
    \label{tab:subject-matter}
  \end{table}   

\begin{table}[t]                  
    \centering                     
    \small   
    \begin{tabular}{@{}l
  p{0.50\linewidth} r r@{}}         
      \toprule
      \textbf{Tag} &                
  \textbf{Description} &          
  \textbf{Items} & \textbf{\%} \\
      \midrule
      \texttt{universal}        &   
  Top-level shape is $\forall       
  x.\,\varphi(x)$ (one or more      
  universally quantified variables, 
  no leading existential). & 159 &
  28.7 \\
      \texttt{conditional}      &
  Statement is an implication
  $\varphi \rightarrow \psi$ at the
  top level. & 82 & 14.8 \\
      \texttt{multi-part}       &
  Prose enumerates several          
  sub-claims that must all hold
  (``\emph{(i) \ldots (ii)          
  \ldots}''). & 66 & 11.9 \\      
      \texttt{conjunctive}      &
  Top-level conjunction $\varphi    
  \wedge \psi$ (after stripping
  outer quantifiers). & 46 & 8.3 \\ 
      \texttt{existential}      & 
  Top-level shape is $\exists       
  x.\,\varphi(x)$, requiring witness
   construction. & 39 & 7.0 \\      
      \texttt{iff}              & 
  Biconditional $\varphi            
  \leftrightarrow \psi$, typically a
   characterisation lemma. & 37 &   
  6.7 \\                          
      \texttt{equality-claim}   &
  Asserts equality of two terms,    
  sets, or derivations. & 25 & 4.5
  \\                                
      \texttt{mixed-quantifier} & 
  Alternating quantifier prefix such
   as $\forall\exists$ or
  $\exists\forall$. & 23 & 4.2 \\   
      \texttt{quantifier-free}  & 
  Propositional or open-formula     
  claim with no quantifier in scope.
   & 20 & 3.6 \\                    
      \texttt{negative-claim}   & 
  Asserts a negation $\neg\varphi$  
  or non-existence. & 10 & 1.8 \\
      \texttt{disjunctive}      &   
  Top-level disjunction $\varphi    
  \vee \psi$ (case-split shape). & 6
   & 1.1 \\                         
      \bottomrule                 
    \end{tabular}                   
    \caption{Logical-shape tags on
  benchmark items ($N{=}554$). Tags 
      classify the top-level      
  connective / quantifier pattern of
   the                            
      statement to be formalized;   
  they are non-exclusive, so column 
  sums
      exceed 100\%. Two long-tail   
  tags                              
      (\texttt{existence-and-uniquen
  ess}, \texttt{tfae}) with $n{=}1$ 
      are omitted.}               
    \label{tab:logical-shape}       
  \end{table}                     

 \begin{table}[t]
    \centering                      
    \small                        
    \begin{tabular}{@{}l
  p{0.50\linewidth} r r@{}}
      \toprule
      \textbf{Tag} &
  \textbf{Description} &            
  \textbf{Items} & \textbf{\%} \\
      \midrule                      
      \texttt{binders-substitution}
  & Reasons about bound variables,  
  free-variable sets,
  $\alpha$-renaming, or             
  capture-avoiding substitution. &
  207 & 37.4 \\
      \texttt{set-theoretic}
  & Manipulates sets, relations,    
  functions-as-graphs, or
  cardinality at the meta-level. &  
  158 & 28.5 \\                   
      \texttt{string-level}
  & Operates on raw symbol strings, 
  lists of tokens, or concatenations
   (e.g.\ string parses, length     
  lemmas). & 48 & 8.7 \\          
      \texttt{numeric-arithmetic}
  & Uses $\mathbb{N}$ /             
  $\mathbb{Z}$, induction on
  naturals, counting, or arithmetic 
  inequalities. & 40 & 7.2 \\     
      \texttt{infinitary}
  & Mentions infinite sets,         
  sequences, K\H{o}nig's lemma, or
  compactness — i.e.\ requires      
  non-finitary reasoning. & 33 & 6.0
   \\
      \texttt{higher-order}
  & Quantifies over predicates,     
  formulas, or other higher-order
  objects. & 25 & 4.5 \\            
      \texttt{equational-chain}   
  & Proof or statement is a chain of
   equalities/equivalences ($a = b =
   c = \ldots$). & 11 & 2.0 \\      
      \bottomrule                 
    \end{tabular}
    \caption{Vocabulary tags on     
  benchmark items ($N{=}554$). Tags
      classify the kinds of         
  mathematical objects the item     
  manipulates;
      they are non-exclusive, so    
  column sums exceed 100\%.}      
    \label{tab:vocabulary}
  \end{table}

  \begin{table}[t]                  
    \centering                    
    \small
    \begin{tabular}{@{}l
  p{0.50\linewidth} r r@{}}         
      \toprule
      \textbf{Tag} &                
  \textbf{Description} &          
  \textbf{Items} & \textbf{\%} \\
      \midrule
      \texttt{parametric}
  & Definition is parameterised by a
   type, signature, or earlier
  structure (e.g.\ a proof system   
  parameterised by its rules). & 99
  & 17.9 \\
      \texttt{defines-multiple}
  & Single textbook item introduces 
  several mutually related
  declarations (e.g.\ a syntactic   
  class plus its constructors). & 62
   & 11.2 \\
      \texttt{predicate-definition}
  & Introduces a new                
  \texttt{Prop}-valued relation or
  judgement (e.g.\ ``$\Gamma \vdash 
  \varphi$''). & 60 & 10.8 \\     
      \texttt{function-definition}
  & Introduces a new total function 
  over already-defined data. & 53 &
  9.6 \\                            
      \texttt{type-definition}    
  & Introduces a new type,          
  structure, or class (non-recursive
   carrier). & 40 & 7.2 \\          
      \texttt{inductive-definition}
  & Introduces an inductive type or 
  inductively-defined predicate with
   explicit constructors. & 31 & 5.6
   \\                             
      \texttt{recursive-definition}
  & Definition is by structural     
  recursion on one of its arguments.
   & 25 & 4.5 \\                    
      \texttt{notational}         
  & Item is essentially syntactic   
  sugar / a notation declaration
  with no new computational content.
   & 11 & 2.0 \\                  
      \texttt{mutual-recursion}
  & Two or more declarations defined
   in a single mutual block, each
  referring to the other. & 5 & 0.9 
  \\                              
      \bottomrule
    \end{tabular}
    \caption{Definition-shape tags
  on benchmark items ($N{=}554$).   
      Tags fire only on items whose
  role is to introduce a new        
      declaration; they are       
  non-exclusive, so column sums     
  exceed 100\%.}                  
    \label{tab:definition-shape}
  \end{table}

\subsection{Tag distributions}
\label{ssec:tag-distributions}

 \begin{figure}[t]                                                        
  \centering                                                                         
  \resizebox{\linewidth}{!}{%
  \begin{tikzpicture}
                                                                                     
  \begin{scope}[xshift=0cm]
    \def\bw{0.30}\def\sep{0.40}\def\sc{0.0060}                                       
    \foreach \k/\n [count=\i from 0] in {0/414, 1/7, 2/54, 3/53, 4/19, 5/6, 6/1}{    
      \pgfmathsetmacro{\x}{\i*\sep}                                                  
      \pgfmathsetmacro{\h}{\n*\sc}                                                   
      \fill[blue!60] (\x, 0) rectangle (\x+\bw, \h);                                 
      \node[font=\tiny, below] at (\x+\bw/2, 0) {\k};                                
      \node[font=\tiny, above] at (\x+\bw/2, \h) {\n};
    }                                                                                
    \draw[->] (-0.1, 0) -- (3.0, 0);                        
    \draw[->] (0, -0.05) -- (0, 3.0) node[above, font=\tiny] {\#items};              
    \node[font=\small, align=center] at (1.3, -0.7) {(a) definition\_shape};         
  \end{scope}
                                                                                     
  \begin{scope}[xshift=3.7cm]
    \def\bw{0.30}\def\sep{0.40}\def\sc{0.0080}                                       
    \foreach \k/\n [count=\i from 0] in {0/312, 1/75, 2/96, 3/43, 4/21, 5/7}{
      \pgfmathsetmacro{\x}{\i*\sep}                                                  
      \pgfmathsetmacro{\h}{\n*\sc}                                                   
      \fill[red!60] (\x, 0) rectangle (\x+\bw, \h);                                  
      \node[font=\tiny, below] at (\x+\bw/2, 0) {\k};                                
      \node[font=\tiny, above] at (\x+\bw/2, \h) {\n};      
    }                                                                                
    \draw[->] (-0.1, 0) -- (2.6, 0);                        
    \draw[->] (0, -0.05) -- (0, 3.0);                                                
    \node[font=\small, align=center] at (1.1, -0.7) {(b) logical\_shape};            
  \end{scope}
                                                                                     
  \begin{scope}[xshift=7.1cm]
    \def\bw{0.30}\def\sep{0.40}\def\sc{0.0112}                                       
    \foreach \k/\n [count=\i from 0] in {0/17, 1/176, 2/223, 3/54, 4/69, 5/15}{
      \pgfmathsetmacro{\x}{\i*\sep}                                                  
      \pgfmathsetmacro{\h}{\n*\sc}                                                   
      \fill[green!55!black] (\x, 0) rectangle (\x+\bw, \h);                          
      \node[font=\tiny, below] at (\x+\bw/2, 0) {\k};                                
      \node[font=\tiny, above] at (\x+\bw/2, \h) {\n};      
    }                                                                                
    \draw[->] (-0.1, 0) -- (2.6, 0);                        
    \draw[->] (0, -0.05) -- (0, 3.0);                                                
    \node[font=\small, align=center] at (1.1, -0.7) {(c) subject\_matter};           
  \end{scope}
                                                                                     
  \begin{scope}[xshift=10.5cm]
    \def\bw{0.30}\def\sep{0.40}\def\sc{0.0127}
    \foreach \k/\n [count=\i from 0] in {0/0, 1/62, 2/154, 3/197, 4/124, 5/17}{      
      \pgfmathsetmacro{\x}{\i*\sep}                                                  
      \pgfmathsetmacro{\h}{\n*\sc}                                                   
      \fill[orange!75] (\x, 0) rectangle (\x+\bw, \h);                               
      \node[font=\tiny, below] at (\x+\bw/2, 0) {\k};       
      \node[font=\tiny, above] at (\x+\bw/2, \h) {\n};                               
    }                                                       
    \draw[->] (-0.1, 0) -- (2.6, 0);                                                 
    \draw[->] (0, -0.05) -- (0, 3.0);                                                
    \node[font=\small, align=center] at (1.1, -0.7) {(d) surface\_features};
  \end{scope}                                                                        
                                                            
  \begin{scope}[xshift=13.9cm]                              
    \def\bw{0.30}\def\sep{0.40}\def\sc{0.0096}
    \foreach \k/\n [count=\i from 0] in {0/174, 1/259, 2/101, 3/19, 4/1}{            
      \pgfmathsetmacro{\x}{\i*\sep}                                                  
      \pgfmathsetmacro{\h}{\n*\sc}                                                   
      \fill[purple!65] (\x, 0) rectangle (\x+\bw, \h);                               
      \node[font=\tiny, below] at (\x+\bw/2, 0) {\k};                                
      \node[font=\tiny, above] at (\x+\bw/2, \h) {\n};
    }                                                                                
    \draw[->] (-0.1, 0) -- (2.2, 0);                        
    \draw[->] (0, -0.05) -- (0, 3.0);                                                
    \node[font=\small, align=center] at (0.9, -0.7) {(e) vocabulary};                
  \end{scope}
                                                                                     
  \end{tikzpicture}%
  }                                                                                  
  \caption{Per-item tag cardinality across the five tagging dimensions ($N{=}554$
  items). Each panel shows how many tags an item carries within the named dimension; 
  bar heights are item counts (also printed above each bar). Cardinality $0$ means
  the item has no tag in that dimension. Counts: \texttt{definition\_shape} max $414$
   (zero-tag), \texttt{logical\_shape} max $312$, \texttt{subject\_matter} max $223$,
   \texttt{surface\_features} max $197$, \texttt{vocabulary} max $259$. Y-scales are
  panel-specific so the shape of each distribution is visible.}
  \label{fig:tag-cardinality}
  \end{figure}                                                                       
  

Figure~\ref{fig:tag-cardinality} shows the distribution of the number
of tags assigned per item, summed across all five categories. The
median item carries roughly a dozen tags and the distribution is
unimodal; no item is left untagged on every axis, and only a handful
of items concentrate more than $20$ tags. Two regularities are worth
noting. First, surface-feature tags fire densely (almost every item
has a non-trivial length tag, and most have at least one of
\texttt{uses-metavariable}, \texttt{enumerated-clauses}, or
\texttt{uses-ellipsis}); this is a property of the textbook's
discursive style rather than of our scheme. Second,
\emph{definition-shape} tags fire only on the
$\sim$$30\%$ of items whose role is to introduce a new declaration,
which explains the long lower tail in the per-axis cardinality.
Together, the figure shows that the schema partitions the benchmark
finely enough to support the per-tag pass-rate analysis that follows
without leaving large untagged subsets.

\subsection{Per-tag pass-rate analysis}
\label{ssec:tag-passrate}

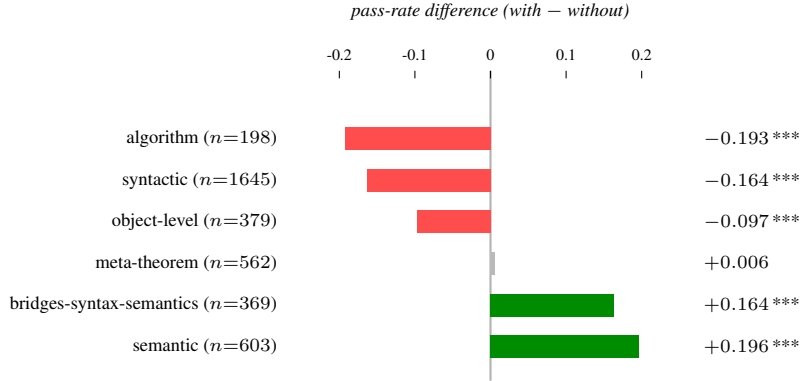
\begin{figure}[t]                                                                  
  \centering                                                                         
  \begin{tikzpicture}[every node/.style={font=\scriptsize}]                          
    \def\bh{0.30}\def\sp{0.55}\def\sc{10}                                            
    \def\labL{-2.7}\def\labR{2.7}                                                    
    \draw[gray!60, thick] (0, 0.50) -- (0, -3.6);                                    
    \foreach \v in {-0.2,-0.1,0,0.1,0.2}{                                            
      \pgfmathsetmacro{\x}{\v*\sc}                          
      \draw (\x,0.40)--(\x,0.50);                                                    
      \node[anchor=south, font=\tiny] at (\x,0.52) {\v};                             
    }                                                                                
    \node[anchor=south, font=\scriptsize\itshape] at (0,1.05)                        
          {pass-rate difference (with $-$ without)};                                 
    \foreach \tag/\d/\n/\s/\col [count=\i from 1] in {%
      algorithm/-0.193/198/{***}/red!70,                                             
      syntactic/-0.164/1645/{***}/red!70,                                            
      object-level/-0.097/379/{***}/red!70,                                          
      meta-theorem/+0.006/562/{}/gray!55,                                            
      bridges-syntax-semantics/+0.164/369/{***}/green!55!black,                      
      semantic/+0.196/603/{***}/green!55!black%
    }{                                                                               
      \pgfmathsetmacro{\y}{-\i*\sp}                                                  
      \pgfmathsetmacro{\x}{\d*\sc}                                                   
      \fill[\col] (0,\y) rectangle (\x,\y+\bh);                                      
      \node[anchor=east] at (\labL,\y+\bh/2) {\tag~($n{=}\n$)};
      \node[anchor=west] at (\labR,\y+\bh/2) {$\d$\,\s};                             
    }                                                       
  \end{tikzpicture}                                                                  
  \caption{Per-tag pass-rate difference for \textbf{subject\_matter}, aggregated over
   six runs of Opus 4.6 on track1\_v3 ($N{=}1950$ item--run trials). Bar = rate(items
   with tag) $-$ rate(items without tag). Stars: * $p{<}.05$, ** $p{<}.01$, ***
  $p{<}.001$ (two-proportion $z$-test). Tags with $n_{\text{with}}{<}20$ or          
  $n_{\text{without}}{<}20$ omitted.}                       
  \label{fig:diff-subject-matter}
  \end{figure}

\begin{figure}[t]
  \centering                                                                         
  \begin{tikzpicture}[every node/.style={font=\scriptsize}] 
    \def\bh{0.28}\def\sp{0.45}\def\sc{10}
    \def\labL{-2.7}\def\labR{2.7}                                                    
    \draw[gray!60, thick] (0, 0.50) -- (0, -5.3);
    \foreach \v in {-0.2,-0.1,0,0.1,0.2}{                                            
      \pgfmathsetmacro{\x}{\v*\sc}                          
      \draw (\x,0.40)--(\x,0.50);                                                    
      \node[anchor=south, font=\tiny] at (\x,0.52) {\v};    
    }                                                                                
    \node[anchor=south, font=\scriptsize\itshape] at (0,1.05)
          {pass-rate difference (with $-$ without)};                                 
    \foreach \tag/\d/\n/\s/\col [count=\i from 1] in {%
      disjunctive/-0.176/30/{*}/red!70,                                              
      quantifier-free/-0.162/102/{***}/red!70,              
      existential/-0.120/198/{***}/red!70,                                           
      multi-part/-0.052/250/{*}/red!70,                                              
      equality-claim/-0.040/118/{}/gray!55,                                          
      mixed-quantifier/-0.040/125/{}/gray!55,                                        
      conditional/+0.031/389/{}/gray!55,                                             
      universal/+0.060/752/{***}/green!55!black,                                     
      conjunctive/+0.072/245/{**}/green!55!black,                                    
      iff/+0.082/132/{*}/green!55!black,                                             
      negative-claim/+0.127/54/{*}/green!55!black%
    }{                                                                               
      \pgfmathsetmacro{\y}{-\i*\sp}                                                  
      \pgfmathsetmacro{\x}{\d*\sc}                                                   
      \fill[\col] (0,\y) rectangle (\x,\y+\bh);                                      
      \node[anchor=east] at (\labL,\y+\bh/2) {\tag~($n{=}\n$)};
      \node[anchor=west] at (\labR,\y+\bh/2) {$\d$\,\s};                             
    }                                                                                
  \end{tikzpicture}                                                                  
  \caption{Per-tag pass-rate difference for \textbf{logical\_shape}. Same conventions
   as Fig.~\ref{fig:diff-subject-matter}.}                                           
  \label{fig:diff-logical-shape}
  \end{figure}
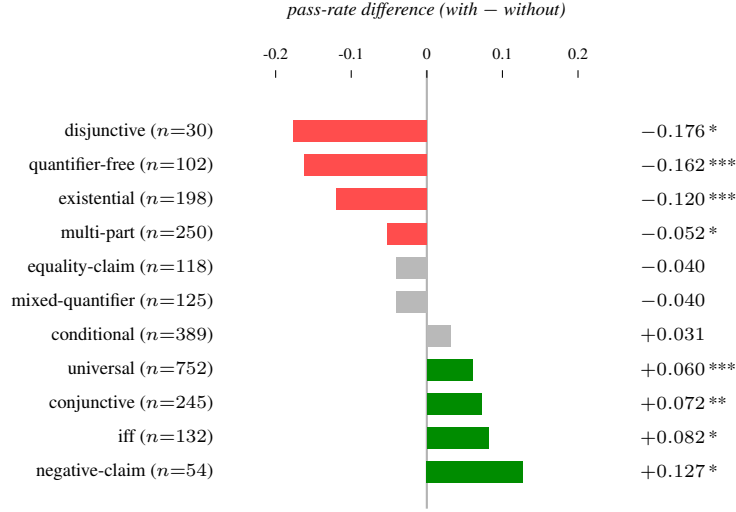

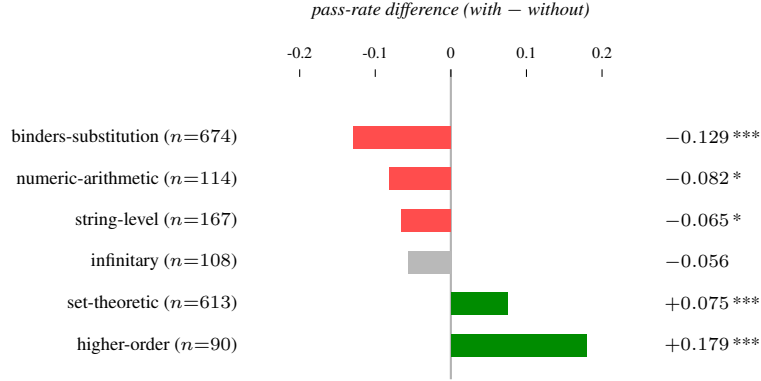
\begin{figure}[t]
  \centering
  \begin{tikzpicture}[every node/.style={font=\scriptsize}]                          
    \def\bh{0.30}\def\sp{0.55}\def\sc{10}
    \def\labL{-2.7}\def\labR{2.7}                                                    
    \draw[gray!60, thick] (0, 0.50) -- (0, -3.6);                                    
    \foreach \v in {-0.2,-0.1,0,0.1,0.2}{
      \pgfmathsetmacro{\x}{\v*\sc}                                                   
      \draw (\x,0.40)--(\x,0.50);                           
      \node[anchor=south, font=\tiny] at (\x,0.52) {\v};                             
    }                                                       
    \node[anchor=south, font=\scriptsize\itshape] at (0,1.05)                        
          {pass-rate difference (with $-$ without)};                                 
    \foreach \tag/\d/\n/\s/\col [count=\i from 1] in {%
      binders-substitution/-0.129/674/{***}/red!70,                                  
      numeric-arithmetic/-0.082/114/{*}/red!70,                                      
      string-level/-0.065/167/{*}/red!70,                                            
      infinitary/-0.056/108/{}/gray!55,                                              
      set-theoretic/+0.075/613/{***}/green!55!black,                                 
      higher-order/+0.179/90/{***}/green!55!black%
    }{                                                                               
      \pgfmathsetmacro{\y}{-\i*\sp}                                                  
      \pgfmathsetmacro{\x}{\d*\sc}                                                   
      \fill[\col] (0,\y) rectangle (\x,\y+\bh);
      \node[anchor=east] at (\labL,\y+\bh/2) {\tag~($n{=}\n$)};                      
      \node[anchor=west] at (\labR,\y+\bh/2) {$\d$\,\s};    
    }                                                                                
  \end{tikzpicture}                                         
  \caption{Per-tag pass-rate difference for \textbf{vocabulary}. Same conventions as 
  Fig.~\ref{fig:diff-subject-matter}.}                                               
  \label{fig:diff-vocabulary}
  \end{figure}

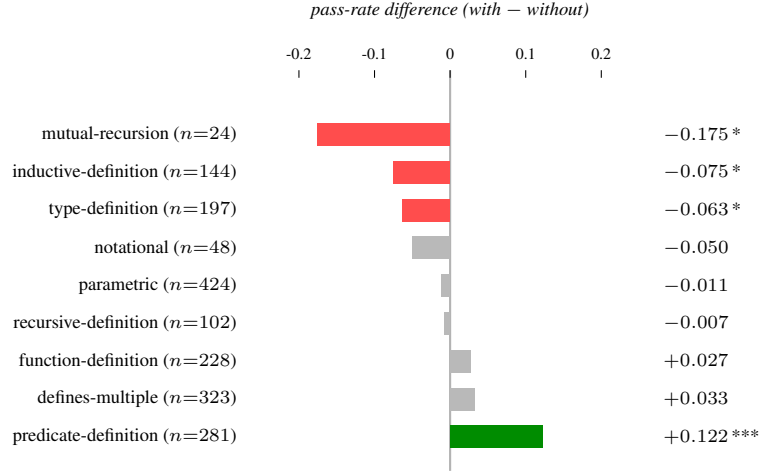
\begin{figure}[t]
  \centering      
  
  \begin{tikzpicture}[every node/.style={font=\scriptsize}] 
    \def\bh{0.30}\def\sp{0.50}\def\sc{10}
    \def\labL{-2.7}\def\labR{2.7}                                                    
    \draw[gray!60, thick] (0, 0.50) -- (0, -4.8);
    \foreach \v in {-0.2,-0.1,0,0.1,0.2}{                                            
      \pgfmathsetmacro{\x}{\v*\sc}                          
      \draw (\x,0.40)--(\x,0.50);                                                    
      \node[anchor=south, font=\tiny] at (\x,0.52) {\v};    
    }                                                                                
    \node[anchor=south, font=\scriptsize\itshape] at (0,1.05)
          {pass-rate difference (with $-$ without)};                                 
    \foreach \tag/\d/\n/\s/\col [count=\i from 1] in {%
      mutual-recursion/-0.175/24/{*}/red!70,                                         
      inductive-definition/-0.075/144/{*}/red!70,           
      type-definition/-0.063/197/{*}/red!70,                                         
      notational/-0.050/48/{}/gray!55,                                               
      parametric/-0.011/424/{}/gray!55,                                              
      recursive-definition/-0.007/102/{}/gray!55,                                    
      function-definition/+0.027/228/{}/gray!55,                                     
      defines-multiple/+0.033/323/{}/gray!55,                                        
      predicate-definition/+0.122/281/{***}/green!55!black%
    }{                                                                               
      \pgfmathsetmacro{\y}{-\i*\sp}                                                  
      \pgfmathsetmacro{\x}{\d*\sc}                                                   
      \fill[\col] (0,\y) rectangle (\x,\y+\bh);             
      \node[anchor=east] at (\labL,\y+\bh/2) {\tag~($n{=}\n$)};                      
      \node[anchor=west] at (\labR,\y+\bh/2) {$\d$\,\s};
    }                                                                                
  \end{tikzpicture}                                         
  \caption{Per-tag pass-rate difference for \textbf{definition\_shape}. Same         
  conventions as Fig.~\ref{fig:diff-subject-matter}.}                                
  \label{fig:diff-definition-shape}
  \end{figure}

\paragraph{Setup.}
For each tag $t$, we compare the pass rate of items carrying $t$
against the pass rate of items not carrying $t$ on Track 1 (IAF). To
maximise statistical power, we aggregate the six curated
\texttt{claude-opus-4-6} runs (three zero-shot and three thinking-mode)
and treat every (item, run) pair as an independent trial, yielding
$N{=}1,950$ trials over $325$ items. We then compute
$r_{\text{with}}(t)$, $r_{\text{without}}(t)$, and the difference
$\Delta(t) = r_{\text{with}}(t) - r_{\text{without}}(t)$ in
percentage points. Significance is assessed with a two-proportion
$z$-test; bars in
Figures~\ref{fig:diff-subject-matter}--\ref{fig:diff-definition-shape} are
annotated with $^{*}$, $^{**}$, $^{***}$ for $p < 0.05$, $0.01$, and
$0.001$ respectively. We restrict to tags with
$n_{\text{with}} \geq 20$, dropping a handful of singleton tags
(\texttt{tfae}, \texttt{existence-and-uniqueness},
\texttt{equational-chain}). Wilson 95\,\% confidence intervals on the
per-tag rates are reported in
\texttt{tag\_analysis\_out/opus\_aggregate/per\_tag.csv}; we omit them
from the bar plots for legibility. Treating runs as independent
slightly inflates significance under within-item correlation, but the
sign and ordering of effects is robust: re-running the analysis on
each individual run separately reproduces the qualitative ranking on
every axis.
We report Opus~4.6 here because it is among the strongest models in our
suite and has the most curated runs; the per-tag effects are
qualitatively preserved across the four largest models we evaluated
(Sonnet~4.6, Kimi-K2.5, Qwen3-Coder), as confirmed by the cross-model
overlay we omit here for space.

\begin{figure}[t]
\centering
\begin{tikzpicture}[every node/.style={font=\scriptsize}]
  \def\bh{0.30}\def\sp{0.55}\def\sc{10}
  \def\labL{-2.7}\def\labR{2.7}
  \draw[gray!60, thick] (0, 0.50) -- (0, -5.85);
  \foreach \v in {-0.2,-0.1,0,0.1,0.2}{
    \pgfmathsetmacro{\x}{\v*\sc}
    \draw (\x,0.40)--(\x,0.50);
    \node[anchor=south, font=\tiny] at (\x,0.52) {\v};
  }
  \node[anchor=south, font=\scriptsize\itshape] at (0,1.05)
        {pass-rate difference (with $-$ without)};
  \foreach \tag/\d/\n/\s/\col [count=\i from 1] in {%
    references-figure/-0.161/131/{***}/red!70,
    informal-notation/-0.159/149/{***}/red!70,
    concrete-instance/-0.136/451/{***}/red!70,
    len-long/-0.109/266/{***}/red!70,
    enumerated-clauses/-0.102/418/{***}/red!70,
    uses-ellipsis/-0.077/484/{***}/red!70,
    len-medium/-0.061/1020/{***}/red!70,
    name-introducing/+0.061/455/{**}/green!55!black,
    uses-metavariable/+0.081/1515/{***}/green!55!black,
    len-short/+0.125/664/{***}/green!55!black%
  }{
    \pgfmathsetmacro{\y}{-\i*\sp}
    \pgfmathsetmacro{\x}{\d*\sc}
    \fill[\col] (0,\y) rectangle (\x,\y+\bh);
    \node[anchor=east] at (\labL,\y+\bh/2) {\tag~($n{=}\n$)};
    \node[anchor=west] at (\labR,\y+\bh/2) {$\d$\,\s};
  }
\end{tikzpicture}
\caption{Per-tag pass-rate difference for \textbf{surface\_features}.
Length buckets (\texttt{len-short}, \texttt{len-medium},
\texttt{len-long}) are included and show a monotone effect: shorter
items pass at $+12.5$\,pp while longer items lose $-10.9$\,pp. Same
conventions as Fig.~\ref{fig:diff-subject-matter}.}
\label{fig:diff-surface-features}
\end{figure}
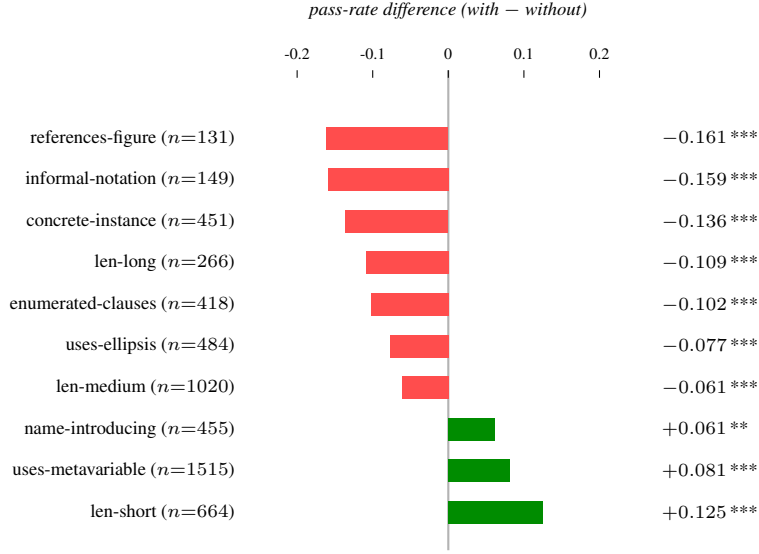

\paragraph{Surface features (Fig.~\ref{fig:diff-surface-features}).}
The surface-feature axis shows the cleanest length effect on the
benchmark: \texttt{len-short} items pass at $+12.5$\,pp,
\texttt{len-medium} at $-6.1$\,pp, and \texttt{len-long} at
$-10.9$\,pp, monotone across all three buckets. Beyond raw length, the
strongest negative effects come from informal or visually-anchored
phrasings: items that \texttt{references-figure} ($-16.1$\,pp),
\texttt{informal-notation} ($-15.9$\,pp), embed a
\texttt{concrete-instance} ($-13.6$\,pp), \texttt{enumerated-clauses}
($-10.2$\,pp), or \texttt{uses-ellipsis} ($-7.7$\,pp). The two
positive non-length tags are \texttt{uses-metavariable} ($+8.1$\,pp)
and \texttt{name-introducing} ($+6.1$\,pp); both indicate that the
prose is already abstract and close to a definition. Read as a whole,
the figure suggests that the auto-formalizer struggles much more with
rendering \emph{semi-formal mathematical English} (figures, ellipses,
case lists, worked examples) into Lean than with translating prose
that is already symbolic. The implication is methodological: a
substantial fraction of benchmark difficulty lives in the
prose-to-symbol translation, not in the underlying mathematics.

\paragraph{Subject matter (Fig.~\ref{fig:diff-subject-matter}).}
This axis shows by far the largest spread of any dimension on the
benchmark, $\approx 40$\,pp from worst to best. \emph{Semantic} items,
which talk about models, satisfaction, and truth, gain $+19.6$\,pp,
and \emph{bridges-syntax-semantics} items (soundness/completeness-style
statements) gain $+16.4$\,pp. By contrast, \emph{algorithm} items,
which prescribe procedural behaviour, lose $-19.3$\,pp;
\emph{syntactic} items, which manipulate proof-system machinery
directly, lose $-16.4$\,pp; and \emph{object-level} arithmetic /
string formalizations lose $-9.7$\,pp. \emph{meta-theorem} items sit
near zero (the dimension nominally distinguishes \emph{about-the-system}
from \emph{inside-the-system} content, and meta-theorems straddle that
boundary). Read together with the surface-feature analysis, the
pattern is striking: models are reliable at the \emph{semantic core}
of logic—well-understood, math-shaped, nameable abstractions—and
unreliable on the \emph{syntactic infrastructure} the textbook spends
most of its space building. Because that infrastructure (sequent
syntax, derivability, substitution lemmas, etc.) is exactly what
theory-scale formalization \emph{has to get right first}, this
single dimension explains a large share of why the absolute pass rate
on Track~1 stays below 20\,\%.

                      
\paragraph{Logical shape (Fig.~\ref{fig:diff-logical-shape}).}
The logical-shape axis isolates the effect of the statement's
top-level connective and quantifier structure, holding subject matter
fixed. Items with single-clause, well-quantified shape pass more
often: \texttt{negative-claim} ($+12.7$\,pp), \texttt{iff}
($+8.2$\,pp), \texttt{conjunctive} ($+7.2$\,pp), \texttt{universal}
($+6.0$\,pp), and \texttt{conditional} ($+3.1$\,pp) all help. Items
that involve case-splitting or witness-construction underperform:
\texttt{disjunctive} ($-17.6$\,pp), \texttt{existential}
($-12.0$\,pp), and \texttt{mixed-quantifier} ($-4.0$\,pp).
\texttt{quantifier-free} items, which in this benchmark almost
exclusively appear inside syntactic lemmas operating on raw formulas,
also lose $-16.2$\,pp. \texttt{multi-part} statements lose
$-5.2$\,pp, consistent with a model that handles a single conditional
fluently but degrades when the prose enumerates several sub-claims.
The pattern is consistent with what one would expect of a translator
that is fluent in standard mathematical phrasing but struggles to
invent witnesses or manage case splits at the type level.

          
\paragraph{Vocabulary (Fig.~\ref{fig:diff-vocabulary}).}
The vocabulary axis shows the single largest negative effect on the
entire benchmark: \texttt{binders-substitution}, $-12.9$\,pp at
$n{=}674$ trials. Any item whose statement requires reasoning about
variable capture, $\alpha$-renaming, or substitution machinery is
formalized substantially less reliably, and because LCS-Bench is
derived from a logic textbook, $37\,\%$ of items are tagged this way.
\texttt{numeric-arithmetic} ($-8.2$\,pp) and \texttt{string-level}
($-6.5$\,pp) reasoning, both of which typically require fixing a
concrete representation, also hurt. The positive end is dominated by
abstract, mathlib-shaped vocabulary: \texttt{higher-order}
($+17.9$\,pp) and \texttt{set-theoretic} ($+7.5$\,pp) items pass much
more often, presumably because abstract operators are easier to map
to Lean's standard library and to predicate definitions than concrete
encodings are. The take-away is concrete: if substitution and
binder-correctness reasoning could be brought up to the pass rate of
the rest of the benchmark, the overall Track~1 pass rate would rise
by an estimated $4$--$5$\,pp—which would be the single largest
absolute gain available from any tag-conditioned intervention.

  
\paragraph{Definition shape (Fig.~\ref{fig:diff-definition-shape}).}
Definition-shape tags fire only on the $\sim$$30\%$ of items whose
role is to introduce a new construct, so the absolute counts are
smaller. \texttt{predicate-definition} items pass at $+12.2$\,pp, the
largest definitional gain on the axis—introducing a new
\texttt{Prop}-valued judgement is mostly a syntactic exercise that
maps cleanly onto Lean inductive predicates. The negatives are all
data-type-shaped: \texttt{mutual-recursion} ($-17.5$\,pp;
$n{=}24$, the worst offender on the axis),
\texttt{inductive-definition} ($-7.5$\,pp), and
\texttt{type-definition} ($-6.3$\,pp). Plain
\texttt{function-definition}, \texttt{recursive-definition}, and
\texttt{parametric} items are near zero. The asymmetry mirrors a
practitioner-known pain point: defining a new \texttt{Prop} is
straightforward, but defining a recursive datatype that must satisfy
Lean's structural-recursion checker—especially within a mutual
block—is a frequent source of hand-tuning even for human formalizers,
and the auto-formalizer reproduces that gap.

\subsection{Reasoning effort vs.\ outcome}
\label{ssec:tag-violin}

\begin{figure}[t]
\centering
\begin{subfigure}[t]{\linewidth}
  \centering
  \includegraphics[width=\linewidth]{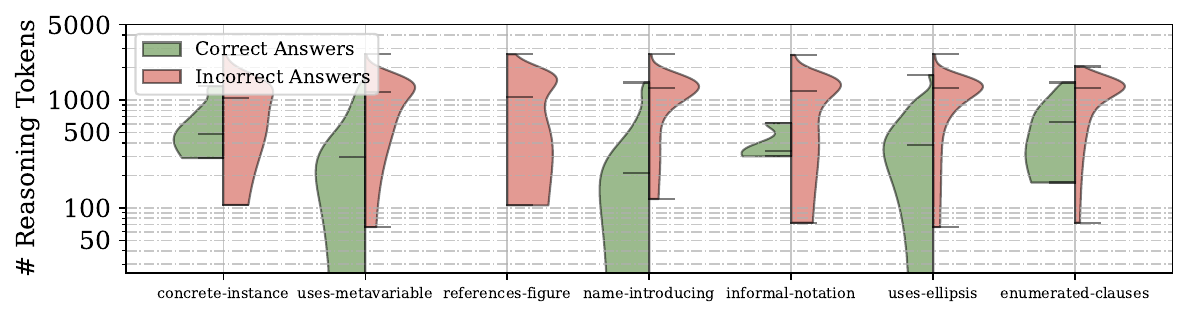}
  \caption{Reasoning tokens by \texttt{surface\_features}.}
  \label{fig:iaf-thinking-reasoning}
\end{subfigure}

\vspace{0.4em}

\begin{subfigure}[t]{\linewidth}
  \centering
  \includegraphics[width=\linewidth]{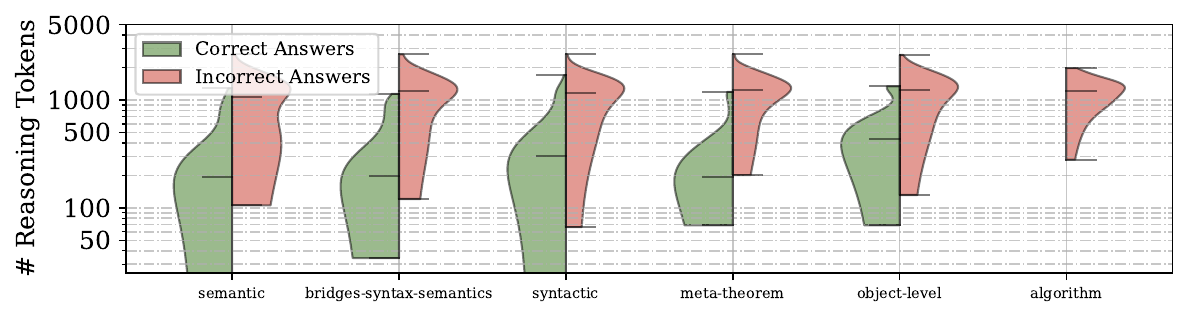}
  \caption{Reasoning tokens by \texttt{subject\_matter}.}
  \label{fig:iaf-agentic-turns}
\end{subfigure}
\caption{IAF reasoning-token distribution per attempt, split by correctness        
  (left half / green: items the model passed; right half / red: items it
  did not), aggregated over three Opus~4.6 thinking-mode runs                        
  ($694$ item--trials with non-zero thinking content). Within each tag, the          
  violin's left half is the distribution over correct attempts and the               
  right half over incorrect attempts; horizontal bars mark the medians.              
  The y-axis is on a log scale.                                                      
  (a) Grouped by \texttt{surface\_features} tag (length buckets                      
  \texttt{len-short}/\texttt{len-medium}/\texttt{len-long} omitted).                 
  (b) Grouped by \texttt{subject\_matter} tag. Tags with fewer than two              
  items on either side are dropped; tags are ordered left-to-right by                
  overall median.}
\label{fig:iaf-violins}
\end{figure}

Figure~\ref{fig:iaf-violins} reports the distribution of reasoning
tokens per attempt, conditioned on outcome (pass vs.\ fail) and
broken down by tag. The data are pooled from the three thinking-mode
Opus~4.6 runs that emit reasoning traces; the metric is approximate
($\#\text{chars}/4$, the standard Anthropic-API conversion) and the
$y$-axis is on a log scale. Two patterns are visible across every
dimension. First, the fail-side median is consistently $4$--$6\times$
higher than the pass-side median: when the model gets an item right
it does so quickly, and when it gets an item wrong it spends an order
of magnitude more reasoning tokens on it. This is a strong signal
that test-time compute is not the binding constraint on the hard
items in LCS-Bench—budgeting is not the issue, the model is failing
\emph{after} sustained deliberation. Second, the spread is wider on
the fail side than on the pass side, which is consistent with the
model exploring multiple unsuccessful strategies before timing out or
emitting a malformed submission. Read together with the per-tag
pass-rate analysis above, the violins suggest that the gains
available from purely scaling reasoning length are limited; the
high-leverage interventions are the structural ones identified in
Sec.~\ref{ssec:tag-passrate}, in particular better handling of
binder/substitution reasoning and of mutual / inductive type
definitions.

\subsection{Take-aways}
\label{ssec:tag-takeaways}

Three patterns recur across the five tagging dimensions.
(i)~\emph{Form matters as much as content.} Surface features alone
(length, ellipses, figure references) move the pass rate by
$10$--$16$\,pp, before any mathematical structure is considered.
(ii)~\emph{Semantic content is easy; syntactic infrastructure is hard.}
Items about models and truth pass at roughly $2$--$3\times$ the rate
of items about derivations and formulas, even though the textbook
spends most of its pages building exactly that syntactic
infrastructure. (iii)~\emph{Within the syntactic side, the dominant
single difficulty is binder/substitution reasoning,} which together
with mutual / inductive type-definitions accounts for the largest
absolute gaps on the benchmark and points to the most actionable
direction for future work on theory-scale auto-formalization.

\end{document}